\RequirePackage{lmodern}
\documentclass[acmtog,nonacm,screen,authorversion]{acmart}

\AtBeginDocument{%
  }

\setcopyright{acmlicensed}
\copyrightyear{2018}
\acmYear{2018}
\acmDOI{XXXXXXX.XXXXXXX}
\acmConference[Conference acronym 'XX]{Make sure to enter the correct
  conference title from your rights confirmation email}{June 03 to 05,
  2018}{Woodstock, NY}
\acmISBN{978-1-4503-XXXX-X/2018/06}
\usepackage{multirow}
\usepackage{pifont}
\usepackage{booktabs}
\usepackage{tabularx}
\usepackage{array}
\usepackage{graphicx}
\usepackage{enumitem}
\usepackage{microtype}
\usepackage{xcolor}
\usepackage{colortbl}
\usepackage{float}
\usepackage{placeins}
\extrafloats{120}

\definecolor{darkgreen}{rgb}{0.0, 0.8, 0.0}
\definecolor{tiergood}{rgb}{0.04,0.42,0.25}
\definecolor{tiermarg}{rgb}{0.62,0.40,0.02}
\definecolor{tierbad}{rgb}{0.70,0.10,0.10}
\newcommand{\good}[1]{\textcolor{tiergood}{\textbf{#1}}}
\newcommand{\marg}[1]{\textcolor{tiermarg}{#1}}
\newcommand{\bad}[1]{\textcolor{tierbad}{#1}}

\acmSubmissionID{1140}

\newcommand{\R}{\mathbb{R}}
\newcommand{\E}{\mathbb{E}}
\newcommand{\I}{\mathbf{I}}

\newcommand{\layers}{\mathcal{L}}
\newcommand{\mesh}{\mathcal{M}}

\newcommand{\kf}{\mathbf{k}}

\newcommand{\noise}{\boldsymbol{\epsilon}}
\newcommand{\loss}{\mathcal{L}}
\newcommand{\cond}{\mathbf{c}}
\newcommand{\layercaption}{\mathbf{t}}

\begin{document}

\title{Bunraku: Turning a Single Illustration into an Editable Live2D Character}

\author{Junhao Chen}
\authornote{Equal Contribution. Work performed during an internship.\quad\textsuperscript{\textdagger}Corresponding Author.}
\email{junhao-c24@mails.tsinghua.edu.cn}
\affiliation{%
  \institution{Tsinghua University}
  \city{Shenzhen}
  \country{China}
}

\author{Jingjia Mao}
\authornotemark[1]
\email{maojj24@mails.tsinghua.edu.cn}
\affiliation{%
  \institution{Tsinghua University}
  \city{Shenzhen}
  \country{China}
}

\author{Dayong Li}
\email{andylida2019@gmail.com}
\affiliation{%
  \institution{Independent}
  \country{Beijing, China}
}

\author{Chenghai Li}
\email{lichenghai99@gmail.com}
\affiliation{%
  \institution{Independent}
  \country{Beijing, China}
}

\author{Saining Zhang}
\email{saining002@e.ntu.edu.sg}
\affiliation{%
  \institution{Nanyang Technological University}
  \country{Singapore}
}

\author{Zhihao Li}
\email{zhihao.li@sparclab.ai}
\affiliation{%
  \institution{SparcAI Inc.}
  \streetaddress{221 W 9th St PMB 141}
  \city{Wilmington}
  \state{DE}
  \postcode{19801}
  \country{USA}
}

\author{Hao Zhao}
\email{zhaohao@air.tsinghua.edu.cn}
\affiliation{%
  \institution{Tsinghua University}
  \city{Beijing}
  \country{China}
}

\author{Yufei Wang}
\authornotemark[2]
\email{yufei.wang@sparclab.ai}
\affiliation{%
  \institution{SparcAI Inc.}
  \streetaddress{221 W 9th St PMB 141}
  \city{Wilmington}
  \state{DE}
  \postcode{19801}
  \country{USA}
}

\author{Ruqi Huang}
\authornotemark[2]
\email{ruqihuang@sz.tsinghua.edu.cn}
\affiliation{%
  \institution{Tsinghua University}
  \city{Shenzhen}
  \country{China}
}

\renewcommand{\shortauthors}{Chen, et al.}

\begin{abstract}
Live2D is the dominant 2D character-animation format for anime characters and virtual avatars, representing each character as a stack of RGBA layers driven by per-layer mesh deformation. Despite its wide use across virtual streaming, mobile games, and interactive characters, authoring a Live2D model still demands weeks of manual layer separation, occlusion completion, mesh placement, and keyframing, and no prior generative method produces such a structured asset end-to-end. We present the first system that, from a single illustration, generates all the structured information a Live2D runtime consumes: ordered RGBA layers, a deformation mesh per layer, and the parameter-driven keypose vertex offsets that make the character move. Stage~1 casts layered decomposition as a layered diffusion process under a Live2D-aware organ-level taxonomy, producing an ordered RGBA stack with hidden-region completion. Stage~2 builds a content-conforming triangle mesh for each layer from its alpha channel alone, and then predicts the keypose displacement field of \emph{all} layers of a character jointly: every vertex of every layer is one token, self-attention spans layer boundaries, and each displacement is factorised into a bounded direction and a log-magnitude. Predicting layers jointly rather than independently is what makes the result a coherent character instead of a set of separately plausible parts, and it is the largest quality gain we measure; scaling the network $112\times$ instead yields none. On $50$ held-out characters, and reported under true generation with no teacher forcing, Stage~2 attains a per-vertex direction cosine of $0.768$ (median $0.828$). Because a layer's mesh is derived from its alpha channel, a clothing layer can be re-textured from a natural-language instruction while the mesh and the predicted animation are reused byte-for-byte. We further contribute \emph{Live2D-Bench}, the first standardized benchmark for the task, and an $8{,}884$-model Live2D corpus with layer and animation supervision, giving the first end-to-end demonstration that turning a single illustration into an editable Live2D model is tractable as a learned task.
\end{abstract}



\begin{teaserfigure}
  \centering
  \makebox[\textwidth][l]{ Project page: \url{https://bunraku-live2d.github.io/}}\\[4pt]
  \includegraphics[width=\textwidth]{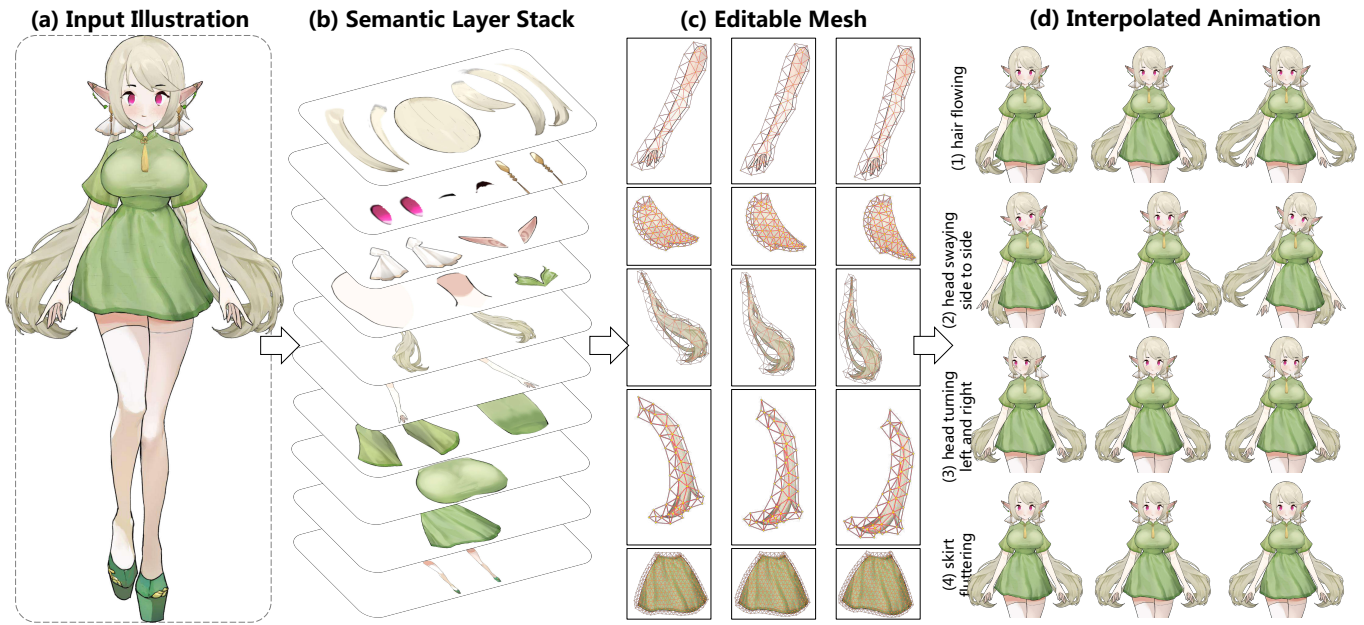}
  \caption{Our method turns a single character illustration into a structured, editable, animatable Live2D model. The output is not a rasterised video clip but a Live2D asset that artists can keep editing and that drops directly into virtual-streamer, game, and interactive-character runtimes.}
  \Description{Teaser figure for our single-illustration to Live2D system.}
\end{teaserfigure}

\maketitle

\section{Introduction}
Live2D ~\cite{live2d2024cubism} is the dominant 2D character-animation format in the anime and VTuber industry. A Live2D model decomposes a character into $30$ to $100$ RGBA layers (hair, face organs, torso, accessories), assigns each layer a 2D deformation mesh, and drives vertex offsets and z-order from named scalar parameters, producing identity-consistent ``3D-controllable, 2D-painted'' animation. Unlike video it disentangles \emph{identity} from \emph{deformation rule}, and unlike 3D rigging it needs no skeleton or skin weights, so hand-painted brushwork survives into an interactive character deployed in VTuber streaming, games and virtual companions. Producing one has nonetheless resisted automation. A senior artist spends four to eight weeks and roughly one thousand US dollars per full-body model, split across three structural rather than pictorial steps: separating semantic layers \emph{with hidden regions painted in}, drawing a deformation mesh on every layer, and authoring keypose vertex offsets along dozens of parameters so all layers move as one body. Each asks the artist to supply geometry and occlusion the flat illustration does not contain, which is what makes the task hard and leaves most illustrations static.

Prior work addresses fragments of this pipeline and stops short of the asset. \emph{Multi-layer decomposition}~\cite{lin2026seethrough, liu2025omnipsd, yin2024qwenimagelayered} produces RGBA stacks but assigns neither mesh nor motion, so the result still has to be rigged by hand. \emph{Autoregressive mesh generation}~\cite{tang2024edgerunner, wang2024llamamesh, siddiqui2024meshgpt, weng2026garmentgpt} serialises meshes into token streams but targets 3D geometry. \emph{Structured keyframe animation}~\cite{Chen_2026_CVPR_LottieGPT} casts vector animation as autoregressive generation but operates on SVG paths, which cannot express a free-form per-vertex warp of raster artwork. Image-to-video methods~\cite{wang2025wan22, yang2024cogvideox, chen2026dancetogether} do produce motion, yet return pixels rather than a drivable object. \emph{No method spans layered RGBA, per-layer 2D mesh and keyframe-driven deformation at once}, so none yields an asset that is re-editable, re-drivable and reusable in a real workflow.

We argue this is a \emph{structured-asset generation} problem whose most overlooked component, the per-layer mesh deformation, has been held back by a modelling assumption rather than by capacity. Predicting each layer independently is the natural formulation and the wrong one: what makes an animated character read as one body is not that every layer deforms plausibly alone, but that all of them deform \emph{consistently with each other}. An iris that is plausible alone yet slides off its eye white destroys the illusion. Our core idea is to predict the layers \emph{together}. We present \textbf{Bunraku}, named after the Japanese puppet theatre in which several operators move one puppet in unison. It turns a single illustration into a drivable Live2D character, and its animation stage is joint over the whole character: every vertex of every layer is one token of one sequence, and self-attention crosses layer boundaries, so a vertex's displacement is conditioned on every other vertex of the character.

\begin{figure*}[t]
  \centering
  \includegraphics[width=\linewidth]{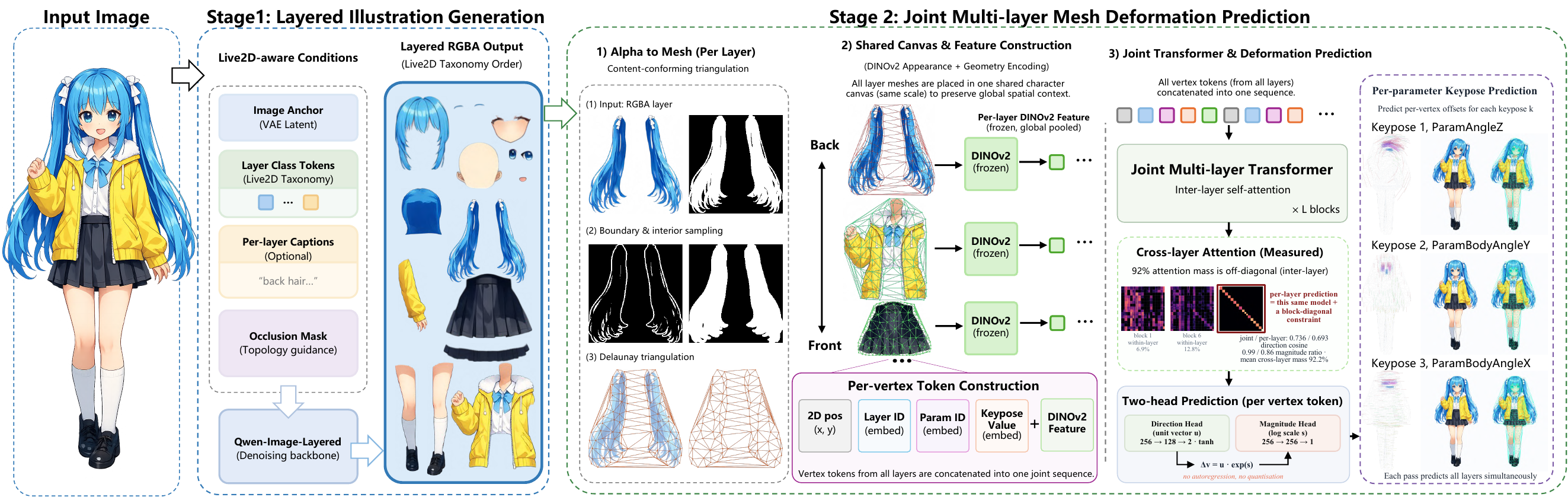}
  \caption{\textbf{Overview of the pipeline.} Left to right: the input illustration; the ordered RGBA
  layer stack with hidden regions completed; the joint animation model, which places every layer's
  vertices in one shared canvas and predicts all layers in a single forward pass; and the drivable rig at
  several absolute parameter values. The dashed strip along the bottom is the editing path of
  \S\ref{subsec:editing}, which re-enters at the layer stack and bypasses the model. Details and the
  measured quantities annotated here are in \S\ref{sec:stage2}.}
  \Description{A five-zone pipeline diagram: input illustration, layer stack with completion outlines, the joint animation model with a shared canvas and token strip, the drivable rig at several parameter values, and an editing path that bypasses the model.}
  \label{fig:intro_pipeline}
\end{figure*}

Bunraku has two stages (Fig.~\ref{fig:intro_pipeline}) and the core idea shapes both. \textbf{Stage~1} is a Live2D-aware layered diffusion model that decomposes the illustration into a taxonomy-compliant RGBA stack with hidden-region completion. Because the layers are later reasoned about jointly, the \emph{set} must be complete and correctly ordered, not merely clean: a missing sleeve is an inter-layer inconsistency no animation model can repair, and we show this is the dominant end-to-end error channel. \textbf{Stage~2} builds a content-conforming triangle mesh per layer from its alpha channel alone and, critically, places every layer's vertices in \emph{one shared character frame} rather than a per-layer crop, so cross-layer geometric proximity exists in the input at all. A single $5.1$\,M-parameter Transformer then regresses per-vertex displacements over the concatenated sequence in one non-autoregressive pass, each factorised into a bounded direction and a log-magnitude because Live2D motion is heavy-tailed.

Every Stage~2 number we report is a \emph{true-generation} number: one forward pass, no teacher forcing. On $50$ held-out characters with zero training overlap, Stage~2 reaches a per-vertex direction cosine of $0.7676$ on average and $0.8278$ at the median, with $31$ of $50$ characters above $0.80$. Joint prediction is what buys this: the per-layer independent formulation scores $0.693$ under an identical protocol and produces exactly the tearing the joint model removes. Scaling from $5.1$\,M to $571.6$\,M parameters does not improve accuracy and a $1.0$\,B run diverges, locating the remaining gap in task ambiguity, decomposition quality and rig-data diversity rather than capacity. The output also behaves like an asset: we extend the vocabulary from $8$ to $24$ parameters, re-texture a layer from a language instruction while reusing every predicted frame byte-for-byte, and run the pipeline on illustrations that were never Live2D models.

Our contributions are summarized as follows:
\begin{enumerate}[leftmargin=1.25em,itemsep=1pt,topsep=2pt]
\item \textbf{Task and dataset.} We formulate \emph{single illustration to editable, mesh-animatable Live2D model} as a structured-asset generation task, identify per-layer mesh animation as its core difficulty, and contribute the largest Live2D corpus to date ($8{,}884$ models) with the derived layer, mesh and keypose records.
\item \textbf{Method.} We propose Bunraku, whose animation stage predicts the displacement field of \emph{all} layers of a character in one pass, with cross-layer self-attention over a shared character frame, a per-layer mesh built from alpha alone, and a direction / log-magnitude factorisation. To our knowledge this is the first system to generate a complete, drivable Live2D rig from one illustration, and we show that coordination, not capacity, makes the animation usable.
\item \textbf{Benchmark.} We propose \emph{Live2D-Bench}, the first standardized benchmark for this task, scoring decomposition, mesh structure and animation on a common random sample of $120$ PSDs. Building it also surfaced a confound future users of the per-vertex metric must control: it is not invariant to mesh density.
\end{enumerate}

\section{Related Work}
\subsection{Raster-pixel and controllable video generation}
\label{subsec:rw_raster}
The first family of character-animation methods returns raster pixels. Image diffusion ~\cite{ho2020denoising, rombach2022stablediffusion} and video diffusion ~\cite{blattmann2023stablevideo, xing2024tooncrafter, wang2025wan22, wan2025wan2, tencent2025hunyuanvideoi2v, xing2024dynamicrafter_eccv, yang2024cogvideox} produce stills or short clips from text or a reference image, and pose-, skeleton- and sketch-driven anime pipelines ~\cite{hu2024animateanyone, wang2024animatex, guo2024animatediff, xu2024magicanimate, zhu2024champ, meng2025anidoc, chen2026dancetogether, Chen_2026_CVPR_hvg3d} add a driving signal alongside an identity image. Multi-layer image generation introduces compositional control by predicting ordered RGBA stacks ~\cite{zhang2024transparent, liu2023text2layer, pu2025art, huang2025psdiffusion, he2026liwi, gao2025layerdecomp, kim2025layeringdiff, li2025layerd, liu2025omnipsd, yin2024qwenimagelayered, lin2026seethrough}, and layered video ~\cite{yang2025layeranimate, niu2024mofavideo, miao2026framessequencestemporallyconsistent} keeps that decomposition alive across time, though its layers are still transported as pixels rather than given a geometric support. A more recent line makes generated video genuinely controllable and even re-editable after the fact, through 3D-enhanced camera and body controls~\cite{cao2025uni3c, chen2026videoworldturningmonocular, chen2026enginenativeeditable3dworld} or by editing sparse 3D point tracks and re-synthesising the frames~\cite{yu2025pointtrackedit}. We take from this family both its conditioning machinery, since our Stage~1 is itself a layered diffusion model, and its correction of a claim we had made too strongly: these methods do make the generation process controllable and the motion re-specifiable, which is real editability. Where we depart is the artefact. What they return is a raster sequence, so a change re-enters the generator, poses that were never synthesised do not exist, and no persistent typed object can be driven at interactive rates from a parameter value never seen at training time. We return the deformation rule itself, which is what makes the two approaches complementary rather than competing.

\subsection{Structured, editable asset generation}
\label{subsec:rw_struct}
A complementary line serialises structured data into discrete token streams generated autoregressively, so the output carries explicit semantics an artist can edit. Recent work covers 3D triangle meshes ~\cite{siddiqui2024meshgpt, chen2024meshanything, chen2024meshxl, tang2024edgerunner, weng2024pivotmesh, wang2024llamamesh, chen2025idea23d, weng2026garmentgpt, weng2026feedforward3deditinglearns}, SVG vector graphics ~\cite{carlier2020deepsvg, lopes2019svgvae, reddy2021im2vec, yang2025omnisvg, wu2025duetsvg, rodriguez2025starvector, wu2025chat2svg, polaczek2025neuralsvg, xing2025llm4svg, chen2025svgthinker, wang2025layertracer}, CAD~\cite{xu2024CADMLLM} assets, and extends to emitting \emph{code} for the artefact~\cite{viga, code2worlds, chen-etal-2026-paircoder, iwbench}, with PairCoder++~\cite{chen2026paircoder} treating pair programming as a paradigm for verified code-driven generation. EdgeRunner ~\cite{tang2024edgerunner} is the most compact mesh tokeniser, adapting EdgeBreaker compression ~\cite{rossignac1999edgebreaker} at four to five tokens per face; closest to us is the line that generates \emph{vector animation}, where LottieGPT ~\cite{Chen_2026_CVPR_LottieGPT} serialises Lottie keyframes and B\'ezier easings into one token stream, OmniLottie ~\cite{yang2026omnilottie} refines that vocabulary into parameterised Lottie tokens, and a further line ~\cite{wu2024aniclipart, gal2023livesketch, gao2025linrbridge} takes the opposite route by distilling motion out of video priors to animate vector artwork, with LiveSVG ~\cite{levy2026livesvgzeroshotsvganimation} doing so zero-shot, with no animation supervision at all. Our output belongs to this family, an asset with a typed parameter interface that a standard runtime replays, and our first design followed the family's method too. That is where we depart, and we report it as a negative result: for keypose \emph{deformation} of an already-given mesh, discretising displacements into a token vocabulary buys nothing, and the accuracies it reports are inflated by teacher forcing (Appendix~\ref{app:mesh_tokenization}). The deeper limit is representational and applies to both routes. The tokenised route expresses motion as transforms of whole primitives plus keyframed parametric paths, which cannot describe a free-form per-vertex warp of hand-painted raster artwork; the video-distillation route needs vector artwork and takes its motion from a raster prior rather than from the parameter-to-deformation mapping an artist authors, so it returns one animation rather than a re-drivable rig. We generate that mapping itself, as a continuous geometric field produced in one pass.

\subsection{Live2D and mesh-based 2D deformation}
\label{subsec:rw_l2d}
Two structured representations dominate anime character animation. \emph{3D mesh, skeleton and skin} (rigging in Maya, Blender or Unity, with auto-rigging ~\cite{anand2024rignet, ma2024rigformer, sun2025drive, Sun_2026_CVPR_Animator}) is fully editable but demands manual skin-weight authoring on stylised topologies and a full 3D pipeline ill-suited to hand-painted art, even where the textured 3D asset itself can be reconstructed from a single image ~\cite{chen2026ultraman}, whereas \emph{Live2D} ~\cite{live2d2024cubism} needs no skeleton or skinning, preserves the artist's strokes, and stays editable per layer and per vertex. Deforming 2D artwork by warping a triangle mesh is long established, from free-form deformation~\cite{sederberg1986freeformdeformation} and as-rigid-as-possible modelling~\cite{sorkine2007arap} to skinning with solved weights~\cite{jacobson2011bbw, morimoto2019rigiddeformation}, so a mesh alone distinguishes nothing. Existing Live2D automation covers only fragments: CartoonAlive ~\cite{huang2025cartoonalive} animates \emph{existing} face models, Textoon ~\cite{liu2024textoon} and Text2AC ~\cite{chen2024text2ac} generate characters on template meshes, SPIRITUS ~\cite{deng2025spiritus} and Outline-and-Detail ~\cite{guo2025outlinedetail} explore layered generation, and lip-sync work ~\cite{aneja2019realtimelipsync, soni2023deeplearninglipsync} handles visemes given pre-decomposed layers. Closest in spirit is PhysAnimator~\cite{yang2024physanimator}, which also starts from one static illustration and also deforms an extracted mesh, but obtains motion from image-space physics simulation and bakes it into a rendered clip through a sketch-guided video model. We adopt Live2D as our target representation for exactly the reasons above, and we differ from every entry here in what we generate: not a clip and not one fragment, but the per-layer mesh \emph{and} the parameter-to-displacement mapping learned from real artist rigs, predicted for all layers jointly so they stay mutually consistent.

\section{Method}
\label{sec:method}

\begin{figure*}[!tp]
  \centering

  \includegraphics[width=\linewidth]{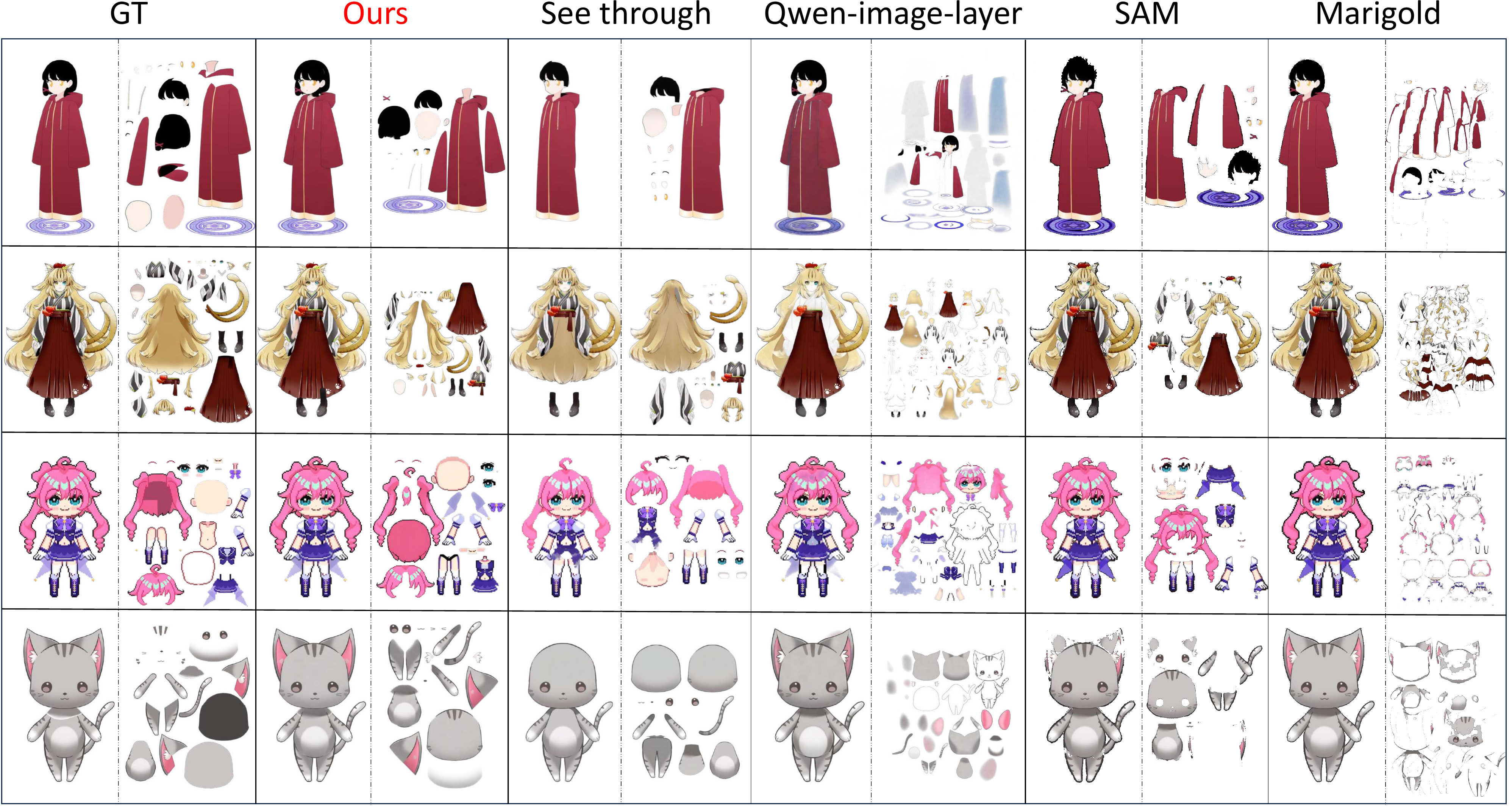}
  \caption{Qualitative comparison against segmentation baselines on the same input illustration.All methods take the same GT illustration as input. From left to right (GT, ours, See through, Qwen-image-layer, SAM, and Marigold), each column pairs the reconstructed character with its decomposed layer textures.}
  \label{fig:video_qual_comp} 

  \vspace{1.5em} 

  \includegraphics[width=\linewidth]{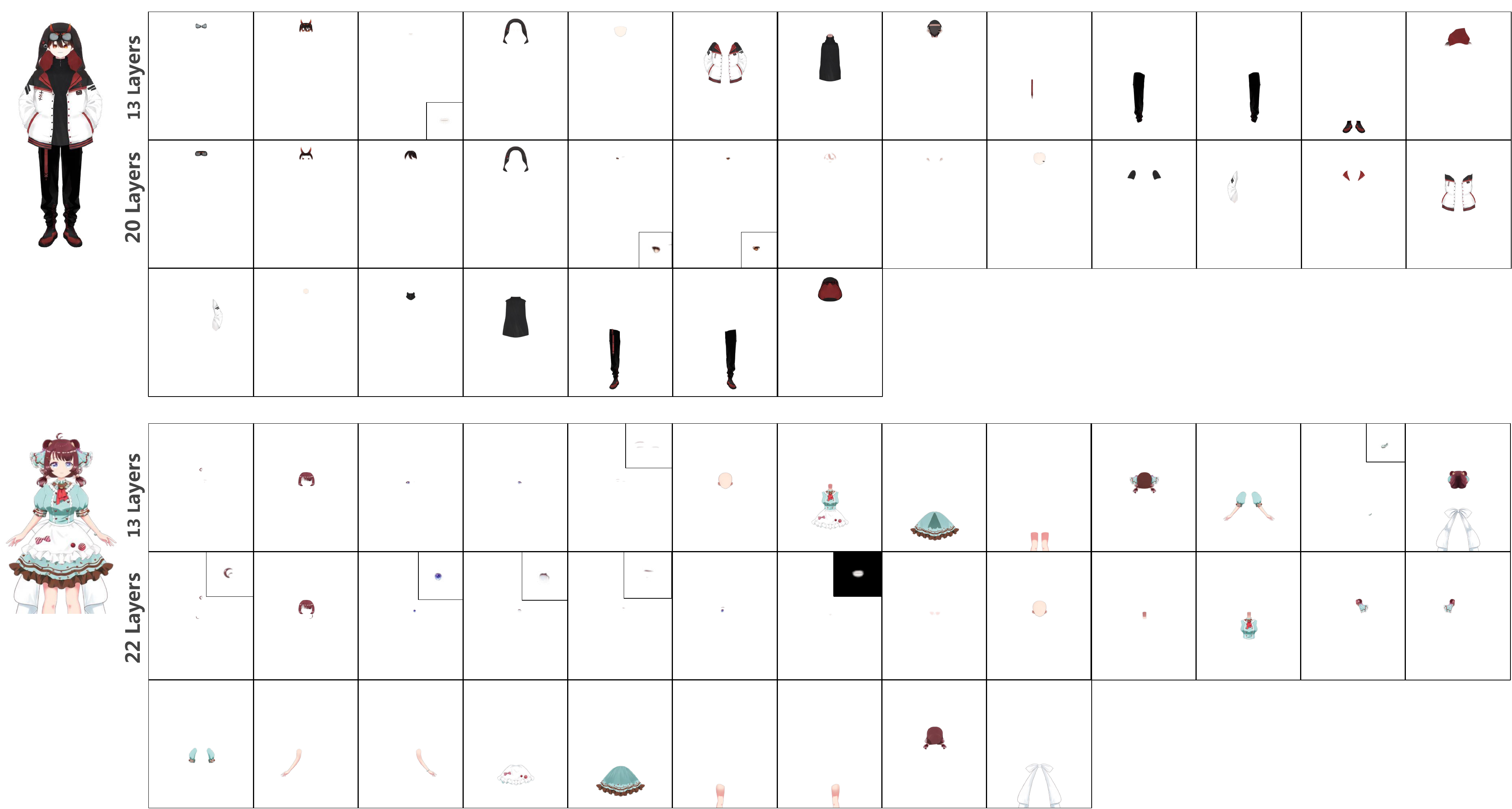}
  \caption{The reconstructed characters (left) are shown alongside their decomposed layer textures (right) under different specified layer counts. Our approach naturally adapts to free layer numbers, supporting both coarse and fine-grained separation.}
  \label{fig:video_qual_demo} 
\end{figure*}

The overview figure is drawn from one character's own data rather than illustrated, and three of its
annotations are measured quantities worth naming here. The input illustration was itself generated from a
text prompt by an image-generation model, so the figure shows the whole chain from text to a drivable rig;
nothing about either stage depends on that, and Figs.~\ref{fig:itw_poses}--\ref{fig:itw_poses_7} show the
same pipeline on artwork found in the wild. The two layers marked in the stack had $33\%$ and $42\%$ of
their area hidden by a layer above and therefore had to be completed rather than copied. The token strip's
segment widths are the real per-layer vertex counts, the arcs under it are the three strongest measured
cross-layer attention pairs, and $92\%$ of that character's attention mass falls outside the within-layer
blocks; measured on five characters the diagonal blocks hold between $4.6\%$ and $6.7\%$ of the matrix by
area and never more than $12.8\%$ of the mass. The
fifth rig cell is a real composite render: the runtime resolves a simultaneous drive by summing the stored
per-parameter fields, so three parameters at once needs no new prediction.

\subsection{Problem formulation}
Given a single illustration $\I\!\in\!\R^{H\times W\times 3}$, we generate a structured Live2D asset
\begin{equation}
\mathcal{A} = \Big(\{\layers_i\}_{i=1}^{N},\ \{\mesh_i\}_{i=1}^{N},\ \big\{\kf_i^{\,p,0}, \kf_i^{\,p,1}\big\}_{i\le N,\ p\in\mathcal{P}}\Big),
\label{eq:l2d_model}
\end{equation}
where $\layers_i\!\in\!\R^{H\times W\times 4}$ is the $i$-th RGBA layer with hidden regions completed, $\mesh_i\!=\!(V_i, F_i)$ is its 2D triangle deformation mesh, $\mathcal{P}$ is the set of modelled animation parameters, and $(\kf_i^{\,p,0},\kf_i^{\,p,1})$ are the start / end keypose vertex offsets of layer $i$ under parameter $p$, so the offset table holds one pair per layer \emph{and} per parameter rather than one pair per layer. Cubism also stores an interpolation type per keyframe; we do not model it and we generate linear tracks throughout, which is what the runtime equation of \S\ref{sec:stage2} assumes, so it is a constant of our output rather than a component of $\mathcal{A}$. $\mathcal{A}$ otherwise contains exactly the information that the Live2D Cubism runtime needs. We decompose into two sequential subtasks handled by two specialised models:
\begin{align}
\textsc{Stage~1:}\ \ &\{\layers_i\} \sim p_{\theta_1}\!\big(\cdot\,\big|\,\I,\,\{\layercaption_i\}\big),
  \label{eq:stage1_factor}\\
\textsc{Stage~2:}\ \ &\mesh_i = \mathrm{Mesh}(\alpha_i), \nonumber\\
  &\big\{\hat\kf_i^{\,p,c}\big\}_{i\le N} = F_{\theta_2}\!\big(\{\layers_i, V_i\}_{i\le N},\, p,\, \bar q_c\big),
  \label{eq:stage2_factor}
\end{align}

The two names for the same object should be said once: a stored keypose offset $\hat\kf_i^{\,p,c}$ is
exactly the per-vertex field the network emits for that layer at that keypose,
$\hat\kf_i^{\,p,c} = \{\widehat{\Delta v}_{i,j}\}_{j\le n_i}$ evaluated at $\bar q_c$, so the
$\widehat{\Delta v}$ of the loss and of the recomposition below is the $\kf$ of the asset definition.
where $\{\layercaption_i\}$ are optional per-layer captions and $c\!\in\!\{0,1\}$ indexes the two keyposes at normalised values $\bar q_c\!\in\!\{-1,+1\}$. The asymmetry between the two lines is
deliberate and worth stating plainly, because it is easy to read the pair as two generative models.
Stage~1 \emph{is} one: it samples a layer stack from a conditional diffusion model and it does use the
captions. Stage~2 is not. Its mesh $\mesh_i$ is a deterministic function of the layer's alpha channel
$\alpha_i$ alone (\S\ref{sec:stage2}), with no learned component and nothing sampled; its keypose
offsets come from a single feed-forward pass of a regressor trained under an $L_1$ objective, so
$F_{\theta_2}$ is a function and not a likelihood; the interpolation type is not predicted at all but
fixed to \textsc{linear}; and Stage~2 receives no
text, its layer conditioning being the frozen image features of $\layers_i$
(Eq.~\ref{eq:stage2_embed}). Here $\bar q_c$ is the keypose value normalised to $[-1,1]$, so $c\!\in\!\{0,1\}$ indexes the two extremes, and one call of $F_{\theta_2}$ returns the field of \emph{every} layer at once (Eq.~\ref{eq:stage2_target}).

\subsection{Dataset construction and layer semantics}
\label{sec:dataset_method}
\paragraph*{Sources and unified extraction.}
We collect publicly available Live2D assets from the internet, yielding approximately ten thousand usable Live2D models after deduplication. Since the runtime binary format is encrypted, we render each model through the official Web SDK in a custom HTML renderer and record per-layer RGBA, art-mesh vertices and faces, and keyframed parameter curves. All vertices of a character are expressed in one \emph{shared} canvas frame normalised to $[-1,1]^2$, so that a layer's position relative to the rest of the character is part of its input rather than being lost to a per-layer crop; coordinates are kept as continuous floating-point values, because Stage~2 (\S\ref{sec:stage2}) regresses displacements directly and therefore needs no quantisation.

\paragraph*{Taxonomy and captions.}
We unify all layers into an $8$-class / $32$-subclass taxonomy (\emph{hair, face skin, eyes, mouth, head accessory, torso, arms, lower body}) used as both Stage~1's layer-class prior and per-layer caption root. Per layer we additionally generate a $\leq 16$-token caption with Qwen-VL ~\cite{bai2025qwen25vl}, for example ``hair/back: long golden twin-tails with purple ribbons'', giving roughly half a million (layer, mesh, caption) triplets.

\subsection{Stage~1: Live2D-aware layered diffusion}
\label{sec:stage1}
We initialise Stage~1 from Qwen-Image-Layered ~\cite{yin2024qwenimagelayered, wang2025qwenimage}, replace its generic ``foreground / background / decoration'' layer tokens with our $8\!\times\!32$ Live2D taxonomy, and supervised fine-tune on $10$\,K Live2D models. The model jointly conditions on (i) the image anchor (VAE latent of $\I$), (ii) Live2D-taxonomy layer-class tokens, (iii) optional per-layer captions ($50\%$ dropout), and (iv) per-layer occlusion masks for hidden-region supervision. The training objective is
\begin{equation}
\loss_\text{Stage~1} = \sum_i \E_{\noise, t}\|\noise - \noise_{\theta_1}(\tilde{\layers}_{i,t}, t, \cond_i)\|_2^2 + \lambda_\text{occ}\sum_i \|m_i^\text{occ}\!\odot\!(\tilde{\layers}_i \!-\! \layers_i^\star)\|_1,
\end{equation}
with $\lambda_\text{occ}\!=\!0.1$. Architecture details are in Appendix~\ref{app:stage1}.

\subsection{Stage~2: joint multi-layer keypose regression}
\label{sec:stage2}

\paragraph*{The Cubism parameter grid.}
A Live2D rig is driven by named scalar \emph{parameters} $p$ (\texttt{ParamAngleX}, \texttt{ParamBody\allowbreak AngleZ}, \texttt{ParamMouth\allowbreak OpenY}, $\dots$), each with an artist-declared range $[p_\text{min}, p_\text{max}]$ and a rest value. For a given layer $i$ and parameter $p$, the artist authors a small set of \emph{keyposes} (parameter values $q^{(p,1)}\!<\!\dots\!<\!q^{(p,K)}$ (typically $K\!\in\!\{2,3\}$: one extreme, rest, the other extreme)) and at each keypose displaces the layer's mesh vertices by hand. This layer-and-parameter-indexed table of vertex displacements is the \emph{ParamVertex} track, and it is the entire animation content of the asset: at runtime the slider value is used only to linearly blend between the two bracketing keyposes (Eq.~\ref{eq:runtime_interp}). Stage~2 therefore has to predict, for every layer, every parameter, and every keypose, a field of 2D vertex displacements.

Concretely, let $V_i\!=\!\{v_{i,1},\dots,v_{i,n_i}\}\!\subset\![-1,1]^2$ be layer $i$'s mesh vertices in the shared character canvas frame, and let $\bar q = 2(q\!-\!p_\text{min})/(p_\text{max}\!-\!p_\text{min})-1$ be the normalised keypose value. We learn a single function that maps a \emph{whole character} at one $(p, q)$ to all of its per-vertex displacements at once,
\begin{equation}
\big\{\widehat{\Delta v}_{i,j}\big\}_{i\le N,\, j\le n_i}
  = F_{\theta_2}\!\Big(\{\layers_i, V_i\}_{i=1}^{N},\ p,\ \bar q\Big),
\label{eq:stage2_target}
\end{equation}
and we predict continuous displacements directly rather than discretised tokens.

\paragraph*{Content-conforming per-layer mesh.}
Each Stage~2 layer needs a base mesh. On rigged training characters we keep the artist's own vertices, which are the native support of the ground-truth displacement field. On a novel illustration there is no artist mesh, so we synthesise one from the layer's alpha channel alone: take every pixel with $\alpha\!>\!4$ (out of $255$) as content, dilate that silhouette by $3$\,px, sample \emph{all} external contours (not only the largest, so a hair layer split into a lock plus a detached tail stays meshed) with a per-contour budget proportional to arc length, add a jittered interior lattice restricted to content pixels, and run Delaunay triangulation ~\cite{barber1996quickhull}. Roughly $55\%$ of the vertex budget goes to the boundary. Two details of this recipe are load-bearing, and both were established by measurement rather than by design.

First, the threshold is pinned between two failures that pull in opposite directions, and it has to be low but \emph{not} zero. Pushing it up fails because Stage~1 emits soft, anti-aliased layer borders: over the in-the-wild layers we measure, pixels with $0\!<\!\alpha\!\le\!16/255$ account for a mean of $10\%$ of all visible pixels and up to $99.8\%$ on a wispy layer, so a nominal cut such as $\alpha\!>\!16/255$ meshes only the opaque core and leaves most of the drawn pixels outside the support; on our worst layer, coverage collapses to $0.18$. Pushing it to zero fails for the opposite reason: Stage~1's diffusion decoder leaves an \emph{imperceptible} noise floor of $1$ to $4/255$ (under $1.6\%$ opacity) across most of a layer's bounding box, so at $\alpha\!>\!0$ the external contour degenerates to the bounding rectangle and the ``content-conforming'' mesh is a box. On real Stage~1 output, $\alpha\!>\!0$ marks $78\%$ of the bounding box as content while $\alpha\!>\!4$ marks $52\%$. We therefore use $\alpha\!>\!4$, which sits above the noise floor and below the perceptible rim ($\alpha\!\geq\!8$): measured over $24$ real Stage~1 layers, thresholds $0$ and $4$ give the \emph{same} coverage of perceptible pixels ($0.9995$ against $0.9996$) while the ratio of mesh area to bounding-box area falls from $0.998$ to $0.788$, i.e.\ only at $\alpha\!>\!4$ does the mesh actually hug the silhouette. An earlier version of this paper argued for $\alpha\!>\!0$ here; that was wrong for the reason just given, and the figures in this version are built at $\alpha\!>\!4$.

Second, we retain \emph{all} Delaunay faces. Throughout this paper \emph{coverage} means one thing: the fraction of a layer's \emph{perceptible} pixels ($\alpha\!\geq\!8$, i.e.\ at least $3\%$ opacity) that fall inside some triangle. Under that definition, discarding faces whose centroid falls off the dilated mask costs little on average, $0.9990$ against $0.9996$, but it is the tail that matters: the fifth percentile falls from $0.9986$ to $0.9945$, the worst layer from $0.9863$ to $0.9682$, and the number of layers below $0.99$ goes from $1$ to $4$ of $151$. A single uncovered rim on one layer is a visible white seam, so we pay the ${\sim}21$ extra faces per layer and keep them all. We note that an earlier version of this paper reported a far larger cost for the discard; that measurement used $\alpha\!>\!0$ as the mask, which as explained above degenerates to the bounding box, and we retract it. The discard also buys nothing visually, because a face lying over transparent texture rasterises to nothing. Retaining faces is therefore free while the gaps that discarding opens are not.

On rigged characters the artist's vertex set alone is likewise not a sufficient support: artists mesh inside the drawn silhouette, so a plain triangulation of their vertices covers a mean of $0.918$ of visible pixels (minimum $0.576$). We therefore \emph{extend} that mesh for rendering (keeping every artist vertex, adding boundary vertices sampled from the dilated contour, and re-triangulating the union) which raises coverage to $0.9998$ (median $1.000$) on that condition. Two threshold details belong here so the numbers are comparable. The artist-mesh extension masks at $\alpha\!>\!0$ while the generated mesh masks at $\alpha\!>\!4$, and this is deliberate: a Stage-1 decoder leaves an imperceptible noise floor covering $26\%$ of a layer's bounding box, whereas artist textures leave only $2.2\%$, so $\alpha\!>\!0$ degenerates to the bounding box on the former but still follows the silhouette on the latter. Each added vertex inherits its displacement from the nearest artist vertex, and the model is never shown the added vertices, so all reported metrics remain defined on the artist's own vertex set. Training, inference and rendering thus share one mesh convention.

\begin{figure}[htbp]
  \centering
  \includegraphics[width=\linewidth]{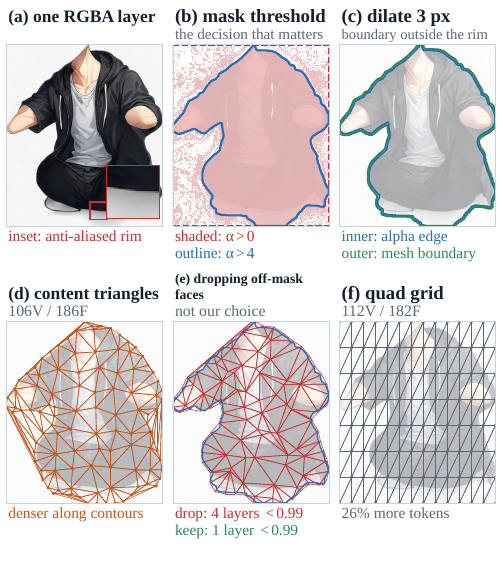}
  \caption{\textbf{Per-layer mesh construction, and what each choice avoids.} Each panel is one real Stage-1
 layer, drawn by the code that builds the shipped meshes: (a)~the RGBA layer, inset on its anti-aliased
 rim; (b)~the $\alpha\!>\!0$ mask (shaded) against $\alpha\!>\!4$ (outline); (c)~the $3$\,px dilation;
 (d)~the triangulation; (e)~keeping every Delaunay face against dropping off-mask faces; (f)~a uniform quad
 grid at equal coverage. The numbers behind each choice are in the text.}
  \Description{Six panels stepping through per-layer mesh construction: the RGBA layer and its
  anti-aliased rim, the two candidate alpha thresholds, the three-pixel dilation, the resulting
  triangulation, the effect of dropping off-mask faces, and a uniform quad grid for comparison.}
  \label{fig:mesh_edge_coverage}
\end{figure}

The $3$\,px dilation is not cosmetic: without it the mesh boundary sits exactly on the alpha edge, the centroid test deletes the rim triangles, and the uncovered opaque pixels render as thin white gaps along every layer border. Measured over $151$ in-the-wild decomposed layers under the definition above, coverage rises from $0.9956$ (median $0.9970$, $14$ layers below $0.99$) without dilation to $0.9996$ (median $1.000$, $1$ layer below $0.99$) with it, which is what closes the gaps. Fig.~\ref{fig:mesh_edge_coverage} shows the effect directly: the generated mesh boundary lies just outside the drawn colour edge on every silhouette, and a layer that is split into several disconnected islands receives one mesh component per island. The alternative support is a uniform quad grid over the layer's bounding box, which is what our own earlier training packs used. We want to be precise about what the contour-conforming mesh does and does not buy over it, because one intuitive argument for it does not survive measurement. Over $46$ in-the-wild layers the two supports are indistinguishable on placement quality: the fraction of vertices landing on content is $0.376$ for the grid versus $0.368$ for ours, the fraction of mesh area lying over transparent pixels is $0.574$ for both, and the IoU between mesh footprint and silhouette is $0.426$ for both. The reason is precisely the soft-alpha property above: scored at $\alpha\!>\!0$, as that comparison was, a Stage~1 layer's content region nearly fills its bounding box, so a grid over that box is not obviously wasteful. What the contour-conforming mesh does buy is \emph{efficiency at equal coverage}: it reaches the same coverage with $83$ vertices and $112$ faces against the grid's $112$ and $182$, i.e.\ $26\%$ fewer tokens. Since every vertex is a token and attention is quadratic in sequence length, that is a direct saving in the dominant cost (\S\ref{app:inference}). Tab.~\ref{tab:stage2_mesh_ab} additionally shows the contour-conforming vertex distribution is somewhat easier to learn a displacement field on, but that comparison is vertex-level only, each condition is evaluated on its own vertex set, and we draw neither a causal nor a rendering conclusion from it.

\paragraph*{Per-vertex conditioning.}
Every vertex of every layer becomes one token. Writing $\gamma_m(x) = \big[x,\,\{\sin(2^\ell\pi x), \cos(2^\ell\pi x)\}_{\ell=0}^{m-1}\big]$ for a Fourier positional encoding, the input embedding of vertex $j$ of layer $i$ is
\begin{equation}
h^{(0)}_{i,j} = W_v\Big[\gamma_6(v_{i,j});\ \mathbf{e}^\text{par}_p;\ \gamma_3(\bar q);\ \mathbf{e}^\text{lay}_i\Big] + W_a\,\phi(\layers_i),
\label{eq:stage2_embed}
\end{equation}
where $\mathbf{e}^\text{par}_p\!\in\!\R^{64}$ is a learned embedding of \emph{which} Live2D parameter is being driven, $\mathbf{e}^\text{lay}_i\!\in\!\R^{64}$ is a learned layer-identity embedding indexed by the layer's position in the stack, $\phi(\layers_i)\!\in\!\R^{384}$ is the frozen DINOv2-small ~\cite{oquab2024dinov2} appearance descriptor of layer $i$ ($256$ patch tokens mean-pooled), and $W_v, W_a$ are linear maps into the model width $d\!=\!256$.

\paragraph*{How image evidence enters, and how it deliberately does not.}
The appearance term is the only path by which pixels reach Stage~2, so it is worth stating precisely what it is. Each layer is cropped to its alpha bounding box, aspect-padded into a $128\!\times\!128$ tile, bilinearly resized to the $224\!\times\!224$ the backbone expects, and passed once through a \emph{frozen} DINOv2-small encoder; the resulting $16\!\times\!16$ grid of $256$ patch tokens (ViT-S/$14$) is mean-pooled into a single $384$-d vector, which is projected and \textbf{added to every token of that layer}. So a layer contributes one global appearance summary shared by all of its vertices, and a vertex is told what kind of part it sits on (long twin-tails rather than a rigid shoulder pad) but not what the texture looks like at its own position. There is deliberately \emph{no} cross-attention between vertex tokens and image patch tokens, and no encoding of the composited character image: the geometry stream and the appearance stream meet only through this additive summary.

That is a design decision we tested rather than assumed. We also implemented the richer alternative, bilinearly sampling the DINOv2 patch grid at each vertex's own position inside its layer tile, so every vertex receives a \emph{local} appearance feature, plus a per-vertex alpha value and an area-weighted composite descriptor of the whole character as a global context token. On identical data and held-out characters this hurts: cos $0.7269$ against $0.7676$, and magnitude calibration degrades badly (median ratio $1.735$ against $1.234$). Our reading is that per-vertex image features let the network key displacement on local texture detail, which does not transfer across characters, whereas one pooled descriptor per layer forces it to rely on part identity plus the geometric configuration; which does. We therefore keep the pooled form, and we report the negative result because it bounds how much of the remaining error can be blamed on weak image conditioning.

\paragraph*{Joint multi-layer self-attention.}
The central design choice is that \emph{all} layers of a character are concatenated into one sequence of $T\!=\!\sum_i n_i$ tokens and processed by a single Transformer, so self-attention spans layer boundaries:
\begin{equation}
H = \mathrm{Transformer}_{\theta_2}\!\Big(\big[h^{(0)}_{1,1},\dots,h^{(0)}_{1,n_1},\ \dots,\ h^{(0)}_{N,1},\dots,h^{(0)}_{N,n_N}\big]\Big),
\label{eq:stage2_joint}
\end{equation}
with $6$ pre-norm blocks, $4$ heads, and no causal mask ($5.1$\,M trainable parameters in total). Each block is $x\!\leftarrow\!x+\mathrm{MHA}(\mathrm{LN}(x))$ followed by $x\!\leftarrow\!x+\mathrm{MLP}(\mathrm{LN}(x))$ with a $4\times$ expansion; padding is handled by a key-padding mask so batched characters of different sizes never attend across characters.

It is worth being concrete about what "spans layer boundaries" means, because it is the paper's central mechanism. The sequence is \emph{vertices}, not layers: a $40$-layer character with ${\sim}80$ vertices per layer yields $T\!\approx\!3200$ tokens, and every one of the $T^2$ attention pairs is permitted. Grouping the tokens by layer partitions that attention matrix into an $N\!\times\!N$ arrangement of blocks. The $N$ diagonal blocks are within-layer attention; and are \emph{all} a per-layer model can ever use, since it processes each layer in isolation. The off-diagonal blocks are exactly what the joint formulation adds: they let an iris vertex read the position of the eye-white vertices it must stay inside, and a fringe vertex read the forehead it must stay attached to. Because there is no causal mask, the coupling is bidirectional, the eye white is equally free to condition on the iris. The layer-identity embedding $\mathbf{e}^\text{lay}_i$ is what makes these blocks addressable: it is indexed by depth position in the stack, so attention can be modulated by \emph{which} layer a token belongs to and by how far apart in depth two layers are, rather than treating the sequence as an unordered bag of vertices. Nothing in the mechanism is layer-count-specific, so a character with $8$ layers and one with $125$ use the same weights.

\begin{figure*}[t]
  \centering
  \includegraphics[width=\linewidth]{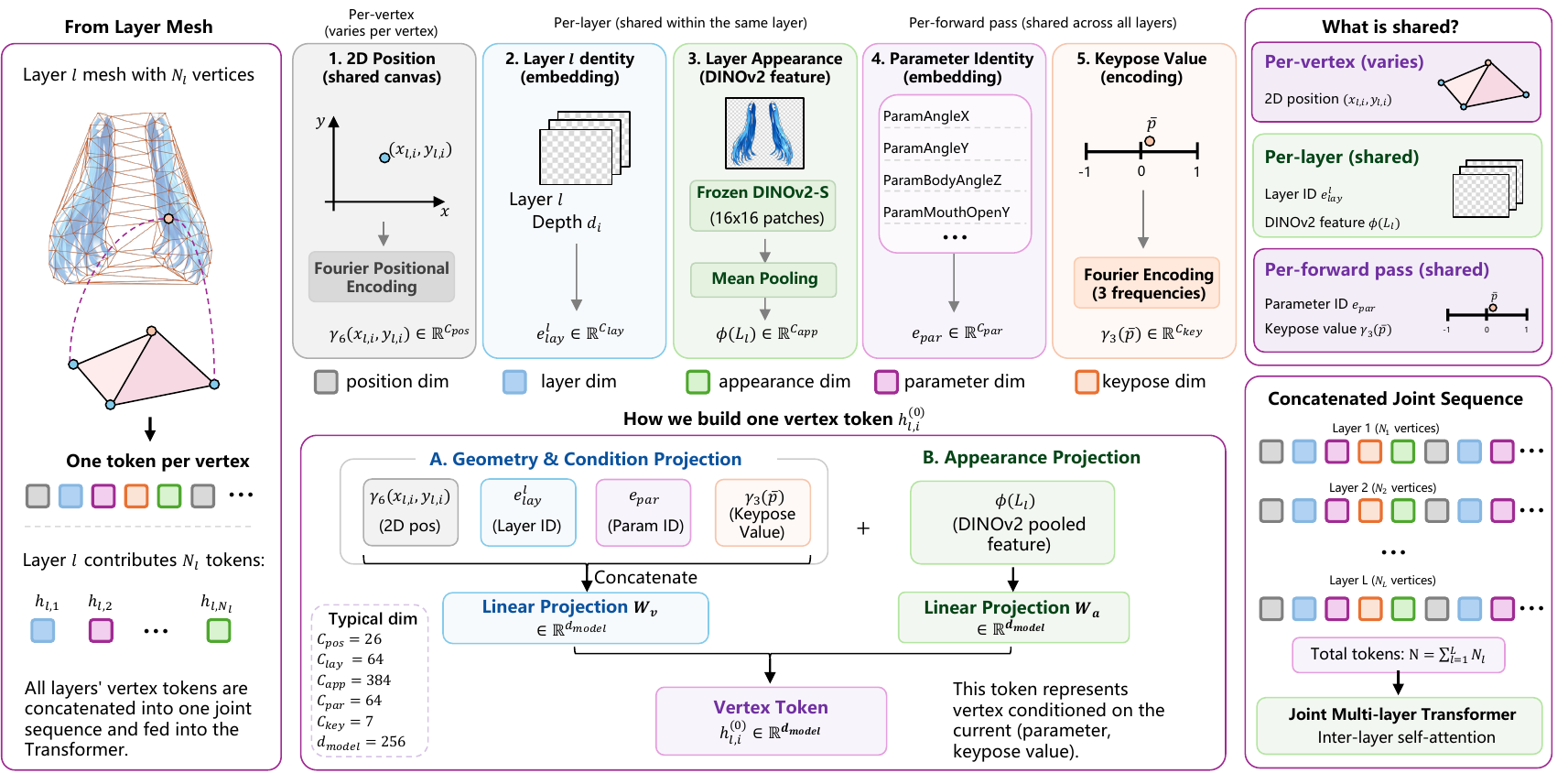}
  \caption{\textbf{Per-vertex token construction: one token per mesh vertex.} The five signals that make up a
 token, grouped by what they vary over: only the $2$D position varies per vertex, the layer embedding and
 the DINOv2 appearance vector are shared within a layer, and the parameter embedding and keypose value are
 shared across the forward pass. The geometry-and-condition signals are concatenated and projected by
 $W_v$; the appearance vector is projected by $W_a$ and \emph{added} (Eq.~\ref{eq:stage2_embed}).
 The per-signal widths are printed in the figure; they sum to $161$, so $W_v$ maps
 $161\!\to\!256$.}
  \Description{A diagram of how one vertex token is built: a 2D position, a layer identity embedding, a
  frozen DINOv2 appearance vector, a parameter identity embedding and an encoded keypose value, projected
  and summed, then concatenated across all layers into one sequence for a joint transformer.}
  \label{fig:pipeline_detail_schematic}
\end{figure*}

Predicting each layer independently (the obvious per-layer formulation, and what our earlier model did) gives the network no way to represent how layers move \emph{relative} to one another, so the iris drifts off its eye white and the fringe separates from the forehead: the tearing artefacts that dominated our previous results. Joint attention removes that failure mode by construction, and it is the single largest quality win we measure (Tab.~\ref{tab:stage2_ablations}). The cost is that attention is quadratic in $T$: measured on one GPU, going from $1052$ to $5476$ tokens ($5.2\times$) raises peak activation memory $24.7\times$, and the joint forwards are the only phase of Stage~2 whose cost grows appreciably with character complexity (\S\ref{app:inference}). This is also why the token saving from the contour-conforming mesh matters. Fig.~\ref{fig:pipeline_detail_schematic} shows the token construction that makes this possible; the block structure of the attention matrix itself is quantified above.

\paragraph*{The attention operator, and what per-layer prediction is in these terms.}
It is worth writing the operator out, because it makes the ablation exact rather than rhetorical. Let $H\!\in\!\R^{T\times d}$ stack the current token states, $d\!=\!256$, and let $S_i\!\subset\!\{1,\dots,T\}$ be the index set of layer $i$'s vertices, so $\{S_i\}_{i=1}^{N}$ partitions the sequence and $|S_i|\!=\!n_i$. Each of the $4$ heads projects $Q^{(k)}\!=\!H W^{(k)}_Q$, $K^{(k)}\!=\!H W^{(k)}_K$, $V^{(k)}\!=\!H W^{(k)}_V$ with $W^{(k)}_\bullet\!\in\!\R^{d\times d_h}$, $d_h\!=\!64$, and forms
\begin{equation}
\begin{aligned}
A^{(k)} &= \mathrm{softmax}\!\Big(\tfrac{Q^{(k)}{K^{(k)}}^{\!\top}}{\sqrt{d_h}} + M\Big) \in \R^{T\times T},\\[2pt]
M_{ab} &= \begin{cases} 0 & \text{$b$ real}\\ -\infty & \text{$b$ padding,}\end{cases}
\end{aligned}
\label{eq:attn_blocks}
\end{equation}
the only mask being the padding mask that keeps batched characters from attending to one another. Reading $A^{(k)}$ through the partition gives $N^2$ sub-blocks $A^{(k)}[S_i, S_j]$: the $N$ diagonal blocks $A^{(k)}[S_i,S_i]$ are within-layer attention, and the $N^2\!-\!N$ off-diagonal blocks $A^{(k)}[S_i,S_j]$, $i\!\neq\!j$, are cross-layer attention. Per-layer prediction is then \emph{exactly} this model with the additional block-diagonal constraint $A^{(k)}[S_i,S_j]\!=\!0$ for $i\!\neq\!j$; the two formulations differ in nothing else, which is why the comparison in Tab.~\ref{tab:stage2_ablations} isolates cross-layer information rather than any change of capacity, data or objective. Concretely, on a $40$-layer character with ${\sim}80$ vertices per layer ($T\!\approx\!3200$), the diagonal blocks account for only ${\sim}2.5\%$ of the $T^2$ entries, so the joint formulation makes roughly $97.5\%$ of the attention budget available to inter-layer reasoning that a per-layer model structurally cannot access.

Two properties of this design are worth making explicit. First, the operator is \emph{permutation-equivariant} over tokens: no sequence-position encoding is used anywhere, so nothing depends on the order in which vertices are concatenated. The grouping into blocks above is a way of \emph{reading} the attention matrix, not a constraint imposed on it. A token's identity is carried entirely by its input embedding (Eq.~\ref{eq:stage2_embed}) (its canvas position, the driven parameter, the keypose value, its layer-identity code) which is what lets the same weights serve characters with $8$ and with $125$ layers, and what lets the layer-identity embedding modulate attention by depth relationship rather than by arbitrary index. Second, because a vertex's canvas coordinate is its own positional signal and all layers share one canvas frame, geometric adjacency across layers is directly available: an iris vertex and the eye-white vertices behind it are nearby in the same coordinate system, so attention can key on spatial proximity across a layer boundary without any explicit correspondence being supplied. This is the mechanism behind the coordination effect; the model is never told which layers are related, only that they inhabit one frame.

Two things this design deliberately does \emph{not} do, for completeness. It does not attend over image patches (\S\ref{sec:stage2}, ``How image evidence enters''), and it does not attend over parameters or keyposes: each $(p,\bar q)$ is an independent forward pass, so the model cannot reason jointly about, say, a head turn and a body lean. Composite poses are instead formed at runtime by summing the per-parameter displacement fields, which is the same superposition the Cubism runtime performs and is therefore the correct behaviour for our target format, but it does mean genuinely coupled multi-parameter deformation is outside the current formulation.

\paragraph*{Decoupled direction and magnitude.}
Live2D displacement magnitudes are extremely heavy-tailed: most vertices move by well under $1\%$ of the canvas while a large head turn moves silhouette vertices by tens of percent (Appendix~\ref{app:alpha}). Regressing raw displacements therefore lets a handful of large-motion vertices dominate the loss. We instead factor each displacement into a bounded shape term and a log-scale term. With $a_i = \max_{j}\|\Delta v_{i,j}\|_\infty$ the layer's peak displacement at the current $(p,\bar q)$ and $\bar u_{i,j} = \mathrm{clip}(\Delta v_{i,j}/a_i, -1, 1)$ the scale-normalised target, two small MLP heads read every token,
\begin{equation}
u_{i,j} = \tanh\!\big(f_\text{dir}(H_{i,j})\big) \in [-1,1]^2, \qquad
s_{i,j} = f_\text{mag}(H_{i,j}) \in \R,
\label{eq:stage2_heads}
\end{equation}
and the predicted offset is recomposed multiplicatively,
\begin{equation}
\widehat{\Delta v}_{i,j} = u_{i,j}\cdot\exp\big(s_{i,j}\big).
\label{eq:stage2_recompose}
\end{equation}
The $\tanh$ makes the shape term exactly as bounded as its target, and putting the scale in $\log$ space makes a $2\times$ magnitude error cost the same whether the true motion is $0.002$ or $0.2$.

\paragraph*{Training objective.}
Both terms are trained with $L_1$ losses, masked to real (non-padding) vertices:
\begin{equation}
\loss_\text{Stage~2} = \E_{i,j}\Big[w_i\,\big\|u_{i,j} - \bar u_{i,j}\big\|_1\Big]
 + \lambda_a\,\E_{i,j}\Big[w_i\,\big|\,s_{i,j} - \log a_i\,\big|\Big],
\label{eq:stage2_loss}
\end{equation}
with $\lambda_a\!=\!1$ and per-layer weights $w_i \propto a_i^{\rho}$ normalised to unit mean. The default model uses $\rho\!=\!0$ (uniform weighting); $\rho\!>\!0$ up-weights large-amplitude layers and is ablated in Tab.~\ref{tab:stage2_ablations}. We train $30$ epochs of AdamW at lr $2{\cdot}10^{-4}$ on one H$800$. For a bounded sequence length, training subsamples at most $28$ vertices per layer and $48$ layers per character (cap $T\!\le\!1200$); at inference the model runs on the full vertex set of every layer, which the attention handles without retraining because Eq.~\ref{eq:stage2_embed} contains no absolute token index.

\paragraph*{Inference and runtime interpolation.}
Inference is one forward pass per $(p,\bar q)$ over the whole character, no autoregressive rollout, hence no exposure-bias gap between training and test, and every number we report in \S\ref{sec:results} is measured in this true-generation regime. Sweeping the parameters we model over their keyposes yields the complete
$\{\kf_i^{\,p,0},\kf_i^{\,p,1}\}$ table of Eq.~\ref{eq:l2d_model}; the released models cover the $8$
parameters of \S\ref{sec:results_stage2} and the $24$-parameter extension of
Appendix~\ref{app:p24}. Writing $\hat V_i^{(p,k)} = V_i + \hat\kf_i^{\,p,k}$ for the deformed
vertex positions that a stored keypose offset produces, the Live2D Cubism viewer (or our WebGL
renderer) drives per-vertex linear interpolation between the two keyposes bracketing the slider value
$q$,
\begin{equation}
V_i(q) = (1-t)\,\hat V_i^{(p,k)} + t\,\hat V_i^{(p,k+1)},\ \ t = (q \!-\! q^{(p,k)})/(q^{(p,k+1)} \!-\! q^{(p,k)}),
\label{eq:runtime_interp}
\end{equation}
followed by per-triangle rasterisation with premultiplied alpha. This is the \textsc{linear}
case, which is the only one we generate; a \textsc{step} track replaces the blend by
$V_i(q)=\hat V_i^{(p,k)}$ for $q<q^{(p,k+1)}$.

\paragraph*{A property the representation gives us for free: texture editing.}
Because a layer's mesh UVs are derived from its alpha bounding box, any edit that preserves a layer's alpha channel and pixel dimensions leaves the mesh (and hence the already-predicted animation) exactly valid. This is a property of the output representation rather than a component we train: it means an off-the-shelf instruction-guided image editor ~\cite{qwenimageedit2511} can repaint one clothing layer's RGB, and once the original alpha and size are restored the rig is re-exported with the new texture and \emph{no} re-animation. We treat this as an application of the generated asset and report it in \S\ref{sec:results_editability}; nothing in Stage~1 or Stage~2 is modified for it.

\section{Experiments and Results}
\label{sec:results}
\subsection{Setup}
\label{sec:setup}

\paragraph*{Dataset.}
Our corpus contains a diverse collection of Live2D models (\S\ref{sec:dataset_method}), categorized into human / humanoid and non-human classes by Qwen3.6-Plus tags. Model-action replacement and color / texture replacement significantly expand the Stage~1 supervision and the animation subset before per-layer / per-parameter expansion. All specific numerical statistics regarding the dataset scale are summarized in Table~\ref{tab:dataset_num}. To mitigate computational overhead from extreme layer counts, we systematically compress the training set. Specifically, we merge layers based on inherent Live2D semantic groupings while strictly preserving the original PSD depth order. This reduces structural complexity without compromising vital occlusion relationships. For comprehensive details regarding our data augmentation strategies and the statistical distribution of the dataset, please refer to Appendix~\ref{app:dataset}.

\paragraph*{Dynamic resolution and layer grouping.}
To efficiently balance computational cost and generation fidelity, we employ a dynamic, layer-count-aware resolution scaling strategy. Instead of rigid resolution buckets, we allocate a target pixel budget based on the effective output layer count $L$, preserving the original aspect ratio. 

The target pixel budget $P_{\text{target}}$ is inversely proportional to $L$, clamped by empirical bounds:
$$P_{\text{target}} = \max \left( 196608, \min \left( 1048576, \frac{6400000}{L + 2} \right) \right)$$

To prevent upsampling artifacts, the scaling factor $s$ is bounded by the original pixel count $P_{\text{orig}} = W_{\text{orig}} \times H_{\text{orig}}$ of the input illustration:
$$s = \sqrt{\frac{\min(P_{\text{target}}, P_{\text{orig}})}{P_{\text{orig}}}}$$

Target dimensions are then uniformly scaled and rounded to the nearest integer: $W_{\text{target}} = \text{round}(W_{\text{orig}} \cdot s)$ and $H_{\text{target}} = \text{round}(H_{\text{orig}} \cdot s)$. 

\paragraph*{Live2D-Bench.}
Existing benchmarks stop at either image-to-multi-layer decomposition or raster / video animation; none evaluates whether a method generates a complete Live2D asset. We therefore build \emph{Live2D-Bench}, a unified benchmark that combines image-to-multi-layer evaluation for semantic RGBA decomposition with per-layer mesh-animation evaluation for topology, keypose offsets, and rendered motion. The fixed pool has $120$ examples: $100$ human / humanoid and $20$ non-human cases, stratified by layer count into $10$ to $20$, $20$ to $35$, and $35{+}$ bins in a $1{:}2{:}1$ ratio. For more details, please refer to Appendix~\ref{app:benchmark}.
\begin{table}[hbtp]
  \centering
  \caption{Live2D corpus and benchmark composition.}
  \label{tab:dataset_stats}
  \small
  \setlength{\tabcolsep}{5pt}
  \renewcommand{\arraystretch}{1.05}
  \begin{tabular*}{\linewidth}{@{\extracolsep{\fill}}lr@{}}
    \toprule
    Quantity & Count \\
    \midrule
    Raw usable Live2D models & $8{,}884$ \\
    Human / humanoid models & $7{,}773$ \\
    Non-human models & $1{,}111$ \\
    Augmented layer-decomposition examples & $\sim 50{,}000$ \\
    Animation examples before per-layer expansion & $\sim 35{,}000$ \\
    \midrule
    Live2D-Bench examples & $120$ \\
    Human / humanoid benchmark examples & $100$ \\
    Non-human benchmark examples & $20$ \\
    Layer-count bins $(10$ to $20)/(20$ to $35)/(35{+})$ & $30/60/30$ \\
    \bottomrule
  \end{tabular*}
  \label{tab:dataset_num}
\end{table}

\paragraph*{Baselines.}
Stage~1 baselines are depth-based segmentation using Marigold-depth~\cite{ke2024marigold}, SAM segmentation without inpainting~\cite{kirillov2023sam, ravi2024sam3}, zero-shot Qwen-Image-Layered~\cite{yin2024qwenimagelayered}, and See-through~\cite{lin2026seethrough}. Specifically, for Marigold-depth, we employ the original model and partition the estimated depth into discrete layers according to the ground-truth layer count. Our SAM baseline utilizes the fine-tuned model introduced in See-through, which performs a 19-class segmentation based on predefined semantic labels. For See-through itself, we adopt its latest v3 release for evaluation. Stage~2 baselines cover oracle classical fitters (rigid, FFD~\cite{sederberg1986freeformdeformation}, ARAP~\cite{sorkine2007arap}), learning-free methods~\cite{yang2024physanimator,siddiqui2024meshgpt}, image-blind retrieval, DINOv2-conditioned regressors, and image-to-video systems~\cite{xing2024tooncrafter,wang2025wan22,yang2024cogvideox,meng2025anidoc}.

\paragraph*{Metrics.}
We report three metric families: full-image / layer-level scores, mesh-level cosine / magnitude / RMSE / PCK, and animation-level warped-frame $L_1$ / PSNR / SSIM / LPIPS over $24$-frame Body and Face Angle X loops. Crucially, to evaluate predicted outputs with variable layer counts against the ground truth, our layer-level evaluation employs the Hungarian algorithm (balancing $\alpha$-IoU and RGB $L_1$ distance) to establish optimal bipartite layer matching. This design allows us to simultaneously assess layer ordering correctness and compute per-layer fidelity metrics. Furthermore, we introduce a novel \emph{per-pixel Cov-MAE} metric to explicitly quantify the granularity of the image decomposition. The Live2D-Bench HTML inspector complements the automatic metrics. Detailed formulations for all evaluation metrics are provided in Appendix~\ref{app:benchmark}.

\paragraph*{Implementation.}

Stage~1 is fine-tuned on $8$ A$100$ GPUs for $6$K steps. Stage~2 is the $5.1$\,M-parameter joint Transformer of \S\ref{sec:stage2} ($d\!=\!256$, $6$ blocks, $4$ heads), trained from scratch for $30$ epochs of AdamW at lr $2{\cdot}10^{-4}$ on one H$800$ over $1{,}443$ characters and $357{,}381$ (character, parameter, keypose, layer) records, with $50$ characters held out. Stage~2 inference is a single forward pass per (parameter, keypose) pair over the entire character (all layers at once, no autoregressive rollout) and the generated bundle plays in our WebGL renderer at $30$\,fps. Stage~2 end-to-end cost, timed phase by phase with \texttt{cuda.synchronize()} around each phase and
the median of nine repetitions reported, is $0.71$\,s for a small character ($10$ layers, $T\!=\!1052$
vertex tokens) and $2.84$\,s for a large one ($63$ layers, $T\!=\!5476$), with a real Stage-1 output
($19$ layers, $T\!=\!1873$) at $1.39$\,s. Peak activation memory is $189$\,MiB and $1982$\,MiB
respectively on top of $103.9$\,MiB of resident weights ($19.4$ for the joint model, $84.5$ for the
frozen DINOv2), so the whole of Stage~2 fits comfortably on a consumer card. The $5.2\times$ token
growth costs $24.7\times$ activation memory, which is the expected $O(T^2)$ attention scaling and the
one place where very large characters will eventually need windowing. Of the large character's
$2.84$\,s, the joint forward passes are $37\%$, DINOv2 feature extraction $22\%$, rig serialisation
$20\%$, image I/O $19\%$ and \emph{mesh construction only $2.7\%$}: the triangulation whose design
occupies \S\ref{sec:stage2} is computationally free. A complete rig is $13$ forward passes ($8$
parameters, three keyposes for the five angle parameters and two for the three $[0,1]$ parameters, with
the rest pose needing none), each pass covering \emph{all} layers of the character at once. These were
measured on a shared A800 with other tenants occupying $58.5$\,GB and the contention is visible in the
spread (the small case's forward phase ranges $0.16$ to $0.35$\,s), so the minima are the better
estimate of an uncontended run; full per-phase tables are in Appendix~\ref{app:runtime}.

\subsection{Stage~1: layer decomposition}
\label{sec:results_stage1}

Live2D-Bench scores layer stacks with full-image, Hungarian-matched per-layer metrics. Tab.~\ref{tab:stage1_main} reports a representative subset.

\begin{table*}[t]
  \centering
  \caption{Stage~1 layer decomposition on Live2D-Bench. Group~I is full-image composite; Group~II is per-layer Hungarian matching with unmatched-layer penalties; Group~III is global set-theoretic.}
  \label{tab:stage1_main}
  \scriptsize
  \setlength{\tabcolsep}{5pt}
  \renewcommand{\arraystretch}{1.05}
  \newcolumntype{Y}{>{\centering\arraybackslash}X}
  \begin{tabularx}{\linewidth}{@{}lYYYYY@{}}
    \toprule
    Metric & Depth seg. & SAM seg. & Qwen-Image-Layered & See-through & Ours \\
    \midrule
    \multicolumn{6}{@{}l}{\emph{Group~I: full-image (composite vs.\ input)}} \\
    PSNR $\uparrow$                 & \textbf{28.24} & \underline{25.58} & 16.58 & 19.88 & 24.28 \\
    SSIM $\uparrow$                 & \textbf{0.977} & \underline{0.953} & 0.767 & 0.890 & 0.813 \\
    LPIPS $\downarrow$              & \textbf{0.061} & \underline{0.082} & 0.241 & 0.198 & 0.149 \\
    $\alpha$-IoU $\uparrow$    & \textbf{0.957} & 0.851 & 0.916 & 0.873 & \underline{0.950} \\
    RGB-L1 $\downarrow$    & \textbf{0.0161} & \underline{0.0520} & 0.2206 & 	0.1419 & 0.0627 \\
    \midrule
    \multicolumn{6}{@{}l}{\emph{Group~II: per-layer}} \\
    matched-layer LPIPS $\downarrow$       & 0.210 & 0.066 & 0.143 & \underline{0.064} & \textbf{0.055} \\
    penalised $\alpha$-IoU $\uparrow$      & 0.081 & 0.137 & 0.175 & \underline{0.265} & \textbf{0.298} \\
    penalised cost $\downarrow$            & \underline{0.546} & 0.616 & 0.573 & 0.640 & \textbf{0.491} \\
    Order (pair-weighted) $\uparrow$       & 0.391 & 0.302 & \underline{0.431} & 0.266 & \textbf{0.674} \\
    Cov-MAE per-pixel $\downarrow$         & \underline{0.166} & 0.197 & 0.445 & 0.167 & \textbf{0.090} \\
    Mask Dice (penalised) $\downarrow$     & 0.867 & 0.799 & 0.725 & \underline{0.653} & \textbf{0.514} \\
    \bottomrule
  \end{tabularx}
\end{table*}

Three points stand out. First, full-image metrics are misleading on their own: depth segmentation obtains strong PSNR / SSIM by producing a few globally similar layers whose composite matches the input, but its penalised $\alpha$mIoU is only $0.081$. Second, unmatched layers matter; See-through's matched $\alpha$mIoU drops once its missing layers are penalised. Third, coverage-corrected, Stage~1 leads the meaningful metrics: penalised $\alpha$mIoU $0.298$, penalised cost $0.491$, order $0.674$ ($+56\%$ over the best baseline), and the best Group~III scores. Matched-layer LPIPS $0.055$ is $14\%$ better than See-through and $42\%$ better than SAM segmentation.

\begin{figure}[htbp]
  \centering
  \includegraphics[width=\linewidth]{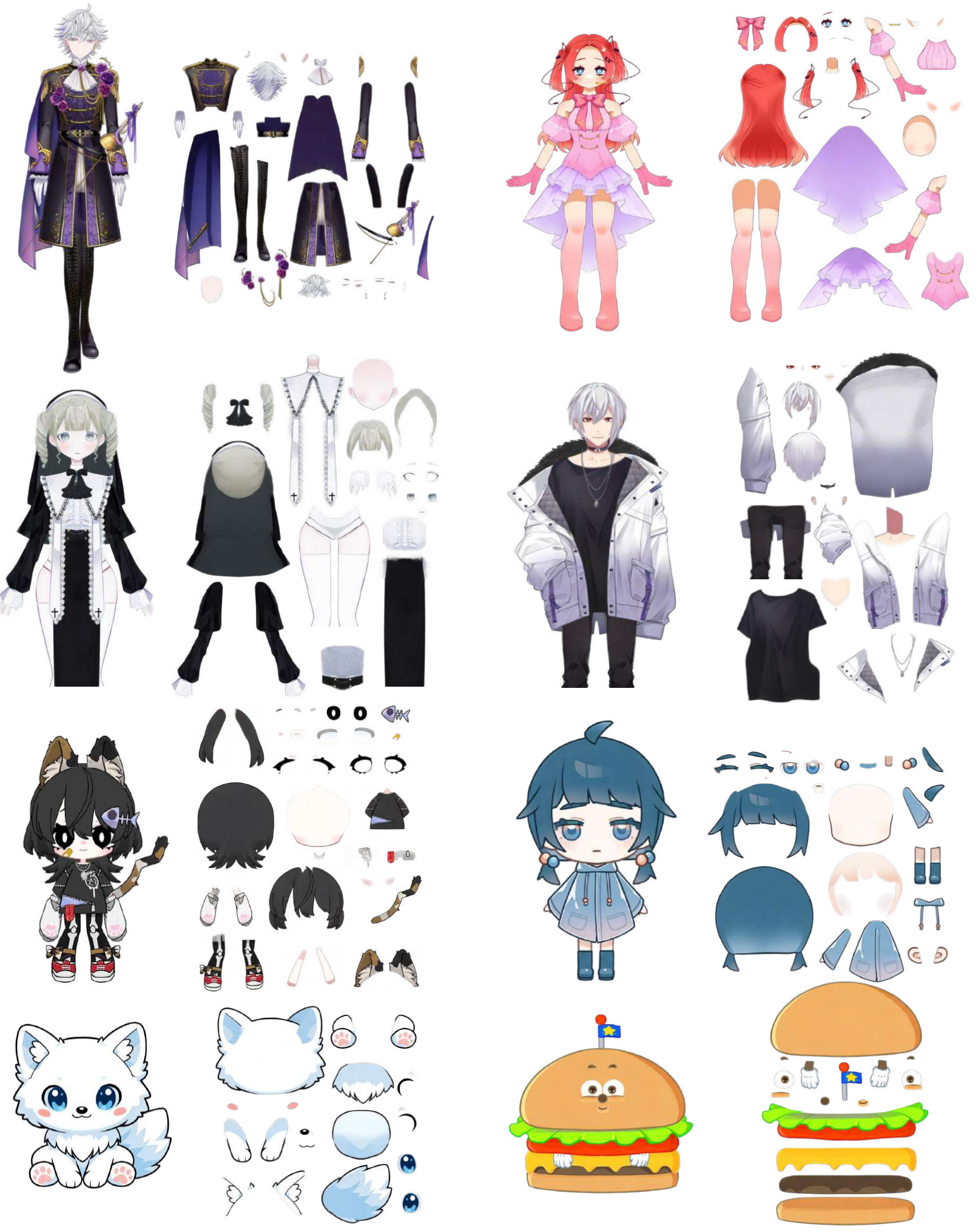}
  \caption{Stage~1 qualitative results on Live2D-Bench. Each example shows how the input illustration is decomposed into an ordered Live2D layer stack with editable RGBA textures and completed occluded regions, rather than a single flat segmentation mask.}
  \label{fig:stage1_qual_lite}
\end{figure}

\subsection{Stage~2: mesh and animation}
\label{sec:results_stage2}

\paragraph*{Protocol and metric.}
Stage~2 is evaluated in the regime in which it is deployed: \emph{true generation}. For each held-out character we feed the model the character's layers and mesh vertices together with a (parameter, keypose) pair, take one forward pass, and compare the predicted displacement field against the artist's. No ground-truth displacement is ever fed back to the model, and no teacher forcing of any kind is used, so the numbers below are what a user actually gets. The primary metric is the \emph{per-vertex direction cosine} between predicted and ground-truth displacement, restricted to vertices the artist actually moves ($\|\Delta v\|\!>\!0.005$ of canvas extent, since the direction of a numerically zero displacement is undefined), averaged within a (layer, parameter, keypose) triple, then within a character, and finally reported as a mean and median over characters. The companion metric is the \emph{magnitude ratio}, the mean predicted displacement norm over the mean ground-truth norm on the same vertices, whose ideal value is $1.0$: below $1$ means the motion is too timid, above $1$ too violent. Because a rig is only useful if the whole character animates, we additionally report how many held-out characters clear a direction cosine of $0.80$.

\begin{table}[t]
  \centering
  \caption{Stage~2 mesh-animation results under \textbf{true generation} on the $50$ held-out
  characters. Metrics per \S\ref{sec:results_stage2}. The FFD rows are expressiveness oracles, not
  competing methods; see the text.}
  \label{tab:stage2_mesh}
  \scriptsize
  \setlength{\tabcolsep}{3pt}
  \renewcommand{\arraystretch}{1.1}
  \begin{tabular*}{\linewidth}{@{\extracolsep{\fill}}lcccc@{}}
    \toprule
    & \multicolumn{2}{c}{dir-cos $\uparrow$} & mag & chars \\
    \cmidrule(lr){2-3}\cmidrule(lr){4-4}\cmidrule(lr){5-5}
    Method & mean & med. & med.\,$\to\!1$ & $\geq\!0.8$ \\
    \midrule
    Ours (joint, $5.1$\,M) & \textbf{0.7676} & \textbf{0.8278} & $1.234$ & $\mathbf{31}/50$ \\
    \midrule
    \multicolumn{5}{@{}l}{\emph{Expressiveness oracles on the same pool and metric$^{\dagger}$}} \\
    FFD $2\!\times\!2$ lattice oracle & $0.9909$ & $0.9955$ & $0.989$ & $46/46$ \\
    FFD $3\!\times\!3$ lattice oracle & $0.9982$ & $0.9992$ & $0.998$ & $46/46$ \\
    FFD $4\!\times\!4$ lattice oracle & $0.9993$ & $0.9997$ & $1.000$ & $46/46$ \\
    \bottomrule
  \end{tabular*}\\[2pt]
\end{table}

\begin{table}[t]
  \centering
  \caption{Stage~2 against image-to-video baselines on the $120$-example Live2D-Bench pool. Baselines see only
 the still image; AniDoc additionally receives a sketch of the ground-truth animation. Bold marks the best
 entry per column. The rows are \emph{not} all on one protocol; see the text.}
  \label{tab:stage2_video}
  \scriptsize
  \setlength{\tabcolsep}{5pt}
  \renewcommand{\arraystretch}{1.05}
  \begin{tabular*}{\linewidth}{@{\extracolsep{\fill}}lccccc@{}}
    \toprule
    Method                & $L_1$ & PSNR & SSIM & LPIPS & Edit. \\
    \midrule
    ToonCrafter           & 0.438 & 3.94  & 0.387 & 0.591 & no \\
    Wan2.2-I2V            & 0.064 & 17.06 & 0.767 & 0.224 & no \\
    CogVideoX-I2V         & 0.044 & 19.13 & 0.806 & 0.193 & no \\
    AniDoc (+GT sketch)   & 0.027 & 23.13 & 0.898 & 0.059 & no \\
    \midrule
    Ours, superseded AR Stage~2$^{\ddagger}$ & 0.015  & 37.63  & 0.965  & 0.049     & \textbf{yes} \\
    \midrule
    \multicolumn{6}{@{}l}{\emph{Current joint Stage~2, different protocol (App.~\ref{app:crossdecomp}), not comparable to the rows above}} \\
    \textbf{Ours (joint, $5.1$\,M)} & n/a & \textbf{41.8} & \textbf{0.955} & \textbf{0.028} & \textbf{yes} \\
    \bottomrule
  \end{tabular*}
\end{table}

The provenance of Tab.~\ref{tab:stage2_video} needs stating, because its rows are not all on one
protocol. The four baseline rows and the ``Ours (superseded)'' row are this paper's previous evaluation,
computed with the autoregressive token Stage~2 of Appendix~\ref{app:mesh_tokenization} on the
$120$-example pool. We report the current joint model \emph{separately} rather than overwriting that row,
because its number comes from a different rendering protocol (Appendix~\ref{app:crossdecomp}: $3$
characters that ship an artist rig, $13$ non-rest keyposes, each condition framed to its own rest content
bbox), and splicing two protocols into one column is exactly the confound we document in
\S\ref{sec:results_p24}. Pixel PSNR on this protocol is also floor-dominated, so it should be read as a
coarse sanity check rather than a motion-quality metric. Re-running the four baselines under the current
protocol is the one measurement this paper still owes; until then the comparison is indicative of scale
only, and Tab.~\ref{tab:stage2_mesh} with Tab.~\ref{tab:crossdecomp_render} are the numbers we ask to be
judged on.

Tab.~\ref{tab:stage2_mesh} reports the per-vertex direction cosine (mean and median over characters),
the magnitude ratio (ideal $1.0$) and the number of characters whose cosine reaches $0.80$, on the $50$
held-out characters, which have zero overlap with training.

The FFD rows need reading carefully, because they are not competing methods. Each is handed the artist's
own displacement field and fits a free-form-deformation lattice to it by least squares, so it measures
what a classical lattice rig can \emph{express}, not what anything can predict from an image; they are
measured on the identical pool through the identical metric code path as the row above. The result is
worth stating plainly: even a $2\!\times\!2$ lattice reproduces the artist's per-layer field to a
direction cosine of $0.99$, so the representation is essentially never the limiting factor and the whole
gap to $1.0$ in the first row is a \emph{prediction} gap. Four of the $50$ held-out characters carry no
records in this pack and are absent from every FFD row. A per-vertex \emph{independent} regression
baseline is reported separately in Tab.~\ref{tab:deploy_prior_ablation}, trained on the same data as our
model rather than borrowed from an earlier run.

\begin{figure*}[!tp]
  \centering
  \includegraphics[width=\linewidth]{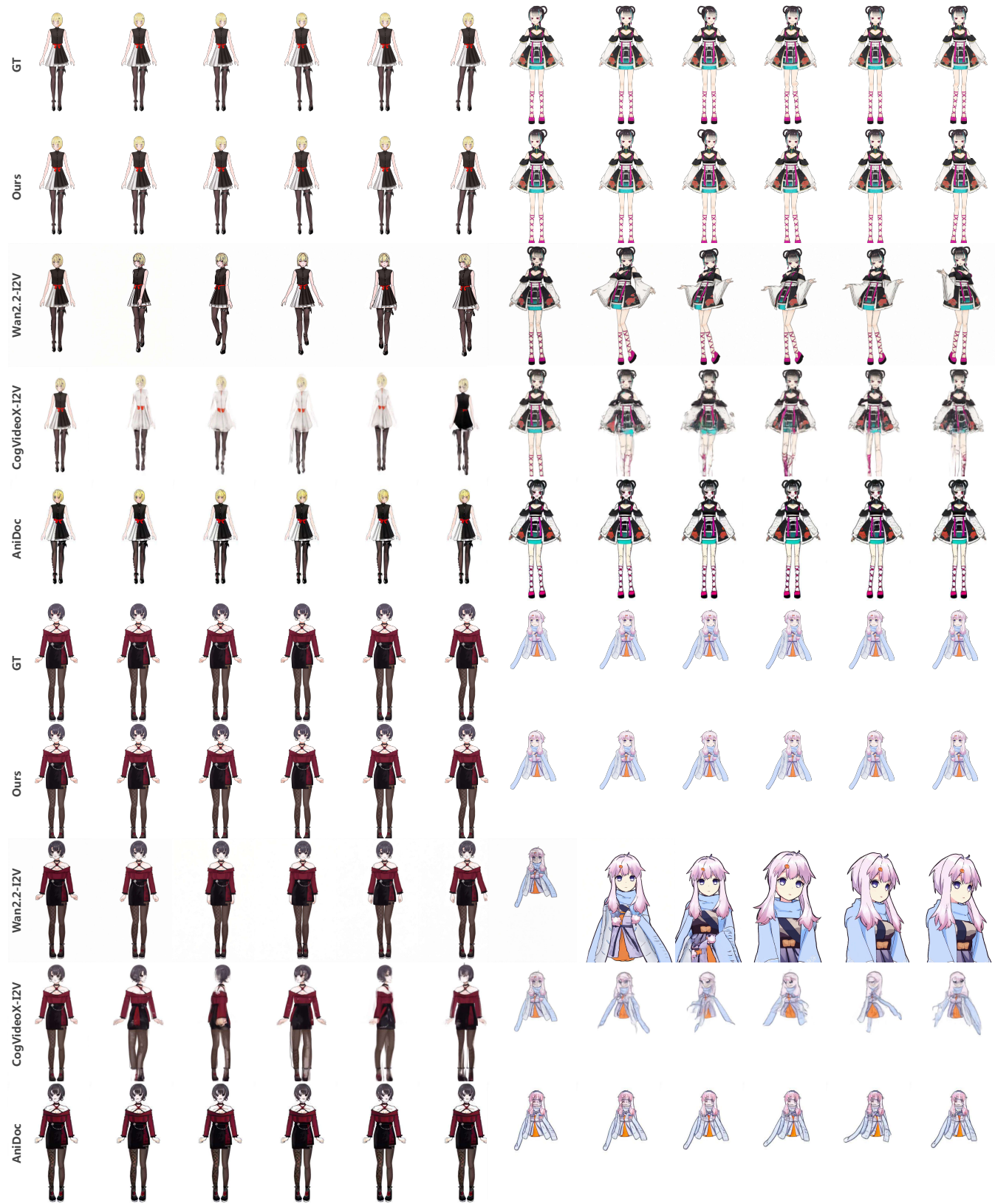}
  \caption{Qualitative comparison against image-to-video baselines on the same input illustration and animation prompt. Each row shows six frames evenly sampled from the generated clip; from top to bottom: GT, Wan2.2-I2V, CogVideoX-I2V, AniDoc, and ours.}
  \label{fig:video_qual}
\end{figure*}

With only (layer stack, mesh vertices, parameter id, keypose value) as input and a single forward pass, Stage~2 reaches a per-vertex direction cosine of $0.7676$ on average and $0.8278$ at the median across the $50$ held-out characters, and $31$ of the $50$ ($62\%$) clear $0.80$. The median-versus-mean gap is informative: the distribution is left-skewed, i.e.\ most characters animate well and a minority fail badly, rather than all characters being mediocre. Amplitude is the weaker axis (the median magnitude ratio is $1.234$, so aggregate motion is if anything slightly too large) but this aggregate hides a systematic effect we discuss below: small motions are over-shot while large turns are under-shot, which is the signature of a pointwise regression objective committing to a central tendency of an ambiguous conditional distribution. We stress that $0.7676$ is a \emph{true-generation} number. Our own earlier autoregressive token model reported cosines above $0.99$ on a comparable task, but only under teacher forcing, where the ground-truth prefix is fed back at every step; that setting measures next-token accuracy, not the quality of a generated rig, and we no longer report it. Rendering itself was a reviewer concern rather than a method concern, and we treat it as one: the renderer's silhouette aliasing came from texture minification without a mip chain rather than from missing multisampling, and fixing it reduces edge-band error against a $16\times$-supersampled reference by $2.34\times$ (Appendix~\ref{app:antialias}); the primary metric is computed on geometry and never touched the rasteriser. At the pixel level (Tab.~\ref{tab:stage2_video}), Stage~2 leads image-to-video baselines by a wide margin even against AniDoc with oracle GT-derived sketches, but we caution that this protocol's PSNR is floor-dominated (Appendix~\ref{app:crossdecomp}) and treat the mesh metrics as primary. Qualitative comparison is in Fig.~\ref{fig:video_qual}.

\paragraph*{Mesh representation: content-conforming triangles are the easier support to learn on.}
The mesh is the support on which the whole animation lives, so we compare the two natural choices under otherwise identical conditions, the same $50$ held-out characters, the same $5.1$\,M architecture, the same objective and schedule, with only the mesh representation on which the model is trained and evaluated changed (Tab.~\ref{tab:stage2_mesh_ab}). Content-conforming triangulation is ahead on direction ($0.7676$ vs.\ $0.7542$ mean, $0.8278$ vs.\ $0.8049$ median), but we must be honest about the size of that gap: $0.013$ on the mean is \emph{inside} the $0.024$ seed floor established above, so this pair of single runs does not establish the ordering on direction alone. The count of usable characters is the more robust signal, since it aggregates a per-character threshold rather than a mean: $31$ vs.\ $26$ of $50$ clear a cosine of $0.80$, a $+19\%$ difference in the quantity a user actually cares about. The quad grid is better on amplitude calibration ($1.116$ vs.\ $1.234$, closer to the ideal $1.0$). Taken together we read this as a mild preference for triangles supported mainly by the usable-character count and by the token-cost argument below, not as a decisive win on cosine. The mechanism is \emph{not} the one we first assumed. We measured both meshes on $46$ real held-out layers and the fraction of vertices landing on opaque content is statistically indistinguishable ($0.376\!\pm\!0.152$ for the grid against $0.368\!\pm\!0.142$ for ours), as are mesh area over transparent region ($0.574$ both) and intersection over union with the silhouette ($0.426$ both); a quad grid does \emph{not} waste vertices on emptiness, because a layer's alpha bounding box is mostly filled by that layer. What differs is cost at equal coverage: content-conforming triangulation reaches the same opaque coverage with $83$ vertices and $112$ faces per layer where the grid needs $112$ and $182$, i.e.\ $26\%$ fewer tokens for the same support, and it places its budget adaptively, densely along contours where the displacement field turns and sparsely in flat interiors where it does not.

Two caveats bound what this experiment establishes, and we state both plainly. First, the per-vertex metric is computed on each condition's \emph{own} vertex set (artist vertices vs.\ resampled grid vertices), so this is not a vertex-identical comparison; it is the same characters, task, objective and architecture, and the artist mesh is the native support of the ground truth while the grid is our synthetic resampling of it, but the two rows are not two readings of one ruler. Second, and importantly, \textbf{this is a vertex-level result only and we make no rendering claim from it}: the artist-mesh data pack stores the artist's vertices and displacements but no per-layer UV / texture correspondence, so artist-mesh rigs cannot be rasterised faithfully; we tried, and the renders show visible edge truncation. Every rendered figure and every qualitative comparison in this paper therefore comes either from the grid-mesh rigs, whose UVs are exact by construction, or from the in-the-wild pipeline, where the mesh vertices are generated \emph{from} the layer's own alpha channel and coverage of perceptible pixels is $0.9996$ on average with a median of $1.000$ (\S\ref{sec:stage2}). The mesh ablation should be read as evidence about which vertex distribution is easier to learn a displacement field on, not as evidence about final image quality.

\begin{table}[t]
  \centering
  \caption{\textbf{Ablation A, mesh representation.} Identical $50$ held-out characters, identical $5.1$\,M architecture, identical objective and schedule; only the mesh changes. True generation. Two caveats: each row is measured on its own vertex set, so the comparison is same-character and same-task but not same-vertex; and this is a \emph{vertex-level} result only, the artist-mesh pack carries no per-layer UV correspondence, so it supports no claim about rendered image quality.}
  \label{tab:stage2_mesh_ab}
  \scriptsize
  \setlength{\tabcolsep}{3pt}
  \renewcommand{\arraystretch}{1.1}
  \begin{tabular*}{\linewidth}{@{\extracolsep{\fill}}lcccc@{}}
    \toprule
    & \multicolumn{2}{c}{dir-cos $\uparrow$} & mag & chars \\
    \cmidrule(lr){2-3}\cmidrule(lr){4-4}\cmidrule(lr){5-5}
    Mesh & mean & med. & med.\,$\to\!1$ & $\geq\!0.8$ \\
    \midrule
    Content-conforming triangles (ours) & \textbf{0.7676} & \textbf{0.8278} & $1.234$ & $\mathbf{31}/50$ \\
    Quad grid re-mesh                   & $0.7542$ & $0.8049$ & $\mathbf{1.116}$ & $26/50$ \\
    \bottomrule
  \end{tabular*}
\end{table}

\begin{figure*}[t]
  \centering
  \includegraphics[width=\linewidth]{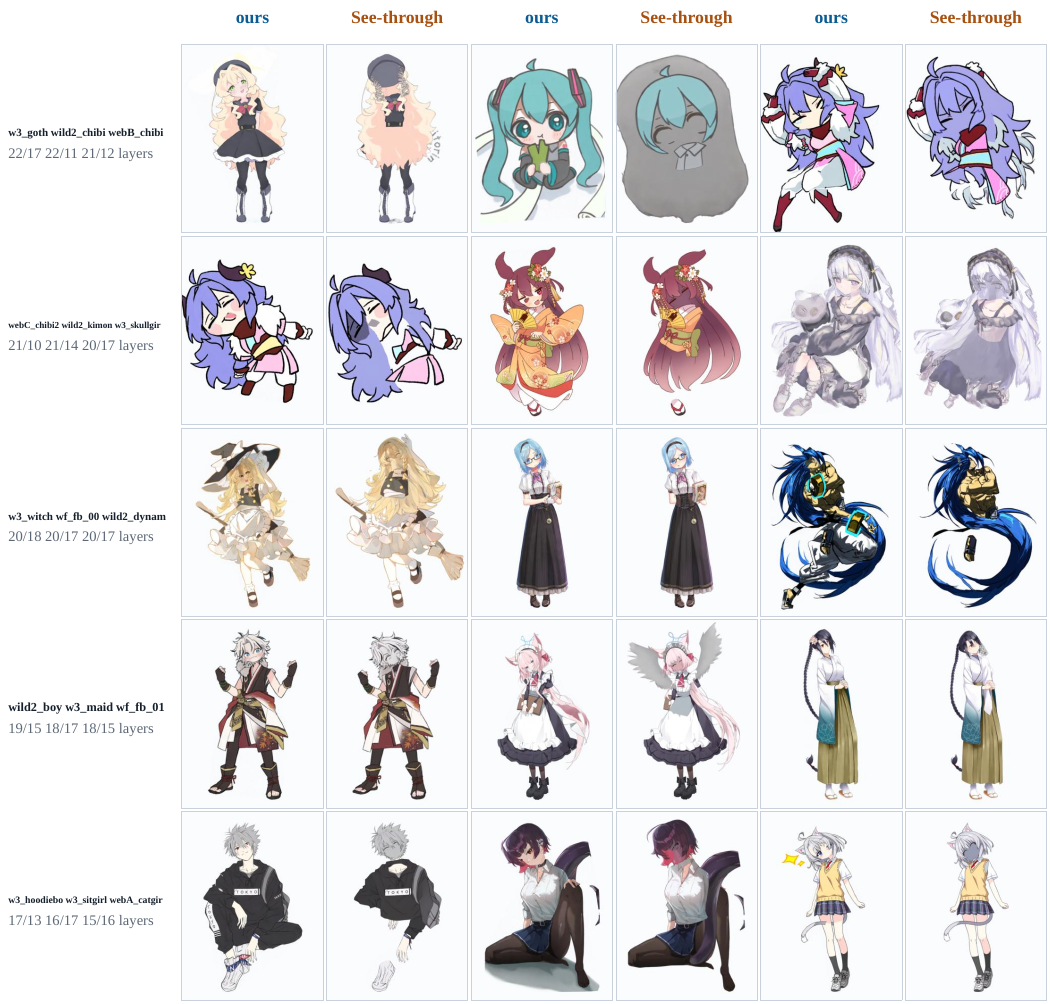}
  \caption{\textbf{Stage~1 is the bottleneck: identical animation model, two decomposers.} Each of the $13$
 in-the-wild illustrations was decomposed twice (left tile ours, right tile See-through~\cite{lin2026seethrough})
 and both stacks were rigged and animated by the \emph{same} frozen Stage~2 checkpoint at the same
 parameter value; layer counts are printed under each tile. Since only the layer source changes, every
 visible difference is Stage-1 error propagating through an unchanged model. Raw outputs; the per-case
 failures are listed in the text.}
  \Description{Thirteen in-the-wild characters animated from two different layer decompositions.}
  \label{fig:itw_two_decomposers}
\end{figure*}

The under-segmented stacks in Fig.~\ref{fig:itw_two_decomposers} fail in the ways a rigger would
predict: \texttt{wild2\_chibi} ($10$ layers) emits the paper sheet behind the character as one opaque grey
layer that then animates as if it were the body; \texttt{wild2\_kimono} ($13$) loses the kimono torso and
legs; \texttt{w3\_hoodieboy} ($12$) loses both legs and detaches a shoe; \texttt{webB\_chibi} ($11$) loses
the boots and hair ornament; \texttt{webC\_chibi2} ($9$) loses the fur cuff; \texttt{w3\_witch} loses the
hat; \texttt{w3\_goth} loses the head-bow and exposes a source watermark our stack keeps in a back layer.
Our stacks stay complete on all $13$. See-through is used as published with no tuning by us and was not
designed for rigging, so this is evidence that decomposition quality dominates the end-to-end result, not
a claim about that method's own task.

\paragraph*{Stage~1 error propagation on in-the-wild illustrations.}
The end-to-end question a user cares about is whether the pipeline works on art that was never a Live2D model. We collected $13$ real web illustrations with no associated rig, decomposed each one \emph{twice} (once with our Stage~1 and once with the third-party See-through decomposer ~\cite{lin2026seethrough}) and animated all $26$ resulting layer stacks ($446$ layers) with the identical frozen Stage~2 model, adding $5$ further illustrations processed by our Stage~1 alone for a total of $31$ rigs and $541$ layers. Because the animation model, its weights, the mesh procedure and the parameter sweep are held fixed, any difference between the two conditions is attributable to the decomposer alone, which is exactly the Stage-1-to-Stage-2 error-propagation channel: Stage~2 can only move what Stage~1 gave it, so a layer that was cut through the middle of a sleeve, or whose occluded region was left hollow, tears or reveals background no matter how good the predicted displacement field is. Fig.~\ref{fig:itw_two_decomposers} shows all $13$ pairs at one identical parameter value, and the failures are exactly of this kind: an under-segmented stack loses a limb, a garment, or the character/background separation, and the animation model then faithfully animates whatever it was handed. The quantitative version of this experiment, on the three held-out characters that ship an artist rig so that a ground-truth animation exists, is reported in Appendix~\ref{app:crossdecomp}: with layer source as the only variable, our Stage~1 reaches rendered PSNR $23.4$\,/\,SSIM $0.883$\,/\,LPIPS $0.088$ against See-through's $21.1$\,/\,$0.865$\,/\,$0.120$, and a rest-pose control shows the residual gap to the artist-layer condition is dominated by decomposition rather than by motion error. One protocol detail matters for fairness: See-through emits a $1280^2$ padded canvas whereas our Stage~1 emits a tight crop, so we normalise every rendered canvas to the character's own content bounding box. Without that normalisation a side-by-side would simply shrink whichever method pads more, which would flatter our method for the wrong reason. The complete in-the-wild record (for each of the $18$ characters, the input illustration, the layers Stage~1 produced, all per-layer generated meshes with silhouette close-ups, and six deterministic poses of the finished rig, plus the two-decomposer comparison at further parameter values, the texture re-skins, and three reproducible failure modes) is provided as a $48$-page supplementary PDF (	exttt{supplementary\_in\_the\_wild.pdf}); every panel there is a raw render with no per-example fix.

\subsection{Ablations}
\label{sec:results_ablations}

\begin{table}[t]
  \centering
  \caption{\textbf{Ablations B and C} on the clean $46$-character benchmark (zero training overlap), true generation, same data / objective / $30$ epochs throughout. Numbers are comparable \emph{within} this table only: this pool and mesh pack differ from Tab.~\ref{tab:stage2_mesh}, which is why the $5.1$\,M row reads $0.7397$ here and $0.7676$ there.}
  \label{tab:stage2_ablations}
  \scriptsize
  \setlength{\tabcolsep}{3pt}
  \renewcommand{\arraystretch}{1.1}
  \begin{tabular*}{\linewidth}{@{\extracolsep{\fill}}lcccc@{}}
    \toprule
    & \multicolumn{2}{c}{dir-cos $\uparrow$} & mag & chars \\
    \cmidrule(lr){2-3}\cmidrule(lr){4-4}\cmidrule(lr){5-5}
    Variant & mean & med. & med.\,$\to\!1$ & $\geq\!0.8$ \\
    \midrule
    \multicolumn{5}{@{}l}{\emph{B: model capacity (identical data and objective)}} \\
    $5.1$\,M ($d\,256$, $6$ blk; ours) & \textbf{0.7397} & \textbf{0.7944} & $1.130$ & $\mathbf{22}/46$ \\
    $38.7$\,M ($d\,512$, $12$ blk)     & $0.6784$ & $0.7605$ & $0.975$ & $20/46$ \\
    $115.0$\,M ($d\,768$, $16$ blk)    & $0.7333$ & $0.7748$ & $1.063$ & $\mathbf{22}/46$ \\
    $204.1$\,M                         & $0.7097$ & $0.7807$ & $1.171$ & $20/46$ \\
    $571.6$\,M                         & $0.6853$ & $0.7565$ & $\mathbf{1.008}$ & $19/46$ \\
    $1.0$\,B                           & \multicolumn{4}{c}{diverged (val cos $-0.085$)} \\
    \midrule
    \multicolumn{5}{@{}l}{\emph{C: amplitude reweighting $w_i \propto a_i^{\rho}$ (Eq.~\ref{eq:stage2_loss})}} \\
    $\rho = 0$ (default)               & $0.7397$ & $0.7944$ & $1.130$ & $22/46$ \\
    $\rho = 0.25$                      & $0.7392$ & n/a & $1.150$ & n/a \\
    $\rho = 0.5$                       & $0.7301$ & n/a & $1.247$ & n/a \\
    \bottomrule
  \end{tabular*}
\end{table}

\paragraph*{Joint cross-layer coordination is the decisive ingredient.}
Replacing per-layer independent prediction with joint prediction over all layers of a pose is the largest single improvement we measure. On the same $46$-character benchmark under true generation, a paired run lifts direction cosine from $0.693$ to $0.736$ and simultaneously corrects amplitude from a magnitude ratio of $0.86$ to $0.99$ (Appendix~\ref{app:crossdecomp}, Tab.~\ref{tab:deploy_prior_ablation}); the joint entry of that pair reads $0.736$ against the committed checkpoint's $0.7397$, a gap well inside the seed floor we measure below. The qualitative effect is larger than the scalar suggests, because the errors that per-layer prediction makes are the visually intolerable kind: an iris that slides off its eye white, or a fringe that separates from the forehead, breaks the illusion of a single character even when the average displacement error is small. Cross-layer self-attention removes this failure mode structurally, since the displacement of a vertex is now conditioned on every other vertex of the character.

\paragraph*{How large a difference is meaningful here.}
Before reading any ablation it is worth establishing the noise floor, because several of the
differences below are small. Two runs of the committed configuration that differ \emph{only} in random
seed score $0.7289$ and $0.7533$ on the clean $46$-character benchmark, a range of $0.024$. We
therefore treat a mean-cosine difference below roughly $0.02$ as uninformative on its own, and we say
so at each point where a comparison falls inside that band rather than reporting the ordering as if it
were established. Two caveats bound this estimate itself. It comes from two replicates, so it is a
crude estimate of a standard deviation rather than a confidence interval, and a third run of a related
configuration scores $0.6918$ but has no surviving training log, so we exclude it rather than use an
unverifiable run to widen the floor. An earlier version of this paper quoted $\approx\!0.006$ for this
quantity; that figure came from a per-record metric rather than the per-character metric we report, and
we retract it.

\paragraph*{What the attention actually does.}
The formalisation above says per-layer prediction is this model plus a block-diagonal constraint, but it
does not say whether the model uses the freedom that removing the constraint gives it. That is
measurable, so we measured it: we captured every block's attention weights on a real character
($15$ layers, $T\!=\!1435$ vertex tokens) and computed how much attention \emph{mass} falls inside the
diagonal blocks against how much of the matrix they occupy. The diagonal blocks are $6.7\%$ of the matrix
by area, and they receive $4.4\%$ of the mass in the first block, rising monotonically to $12.2\%$ in the
sixth: on average $91.9\%$ of the attention mass is cross-layer. Two readings are worth separating. The
first is that attention is \emph{diffuse rather than diagonal}, so the model is not quietly recovering
the per-layer baseline inside a joint architecture. The second is a depth trend we did not anticipate:
early blocks attend broadly across the whole character and later blocks progressively localise, which is
the pattern one would expect if the early layers establish a global frame and the later ones refine
within a part. We are careful not to over-read the first number: a uniform attention would also put mass
in proportion to area, so this measurement shows the cross-layer capacity is used, not that using it
helps. The evidence that it helps is the ablation, $0.693$ against $0.736$ under an identical protocol.
Reading the attention matrix directly on a held-out character ($15$ layers, $T\!=\!1435$), the
within-layer diagonal blocks occupy $6.7\%$ of the matrix by area and receive only $4.4\%$ of the
attention mass in the first transformer block, rising to $12.2\%$ in the last: attention is diffuse
rather than diagonal, and it localises with depth rather than starting localised. Some layers are also
attended to by all the others, appearing as bright columns in the matrix.

\paragraph*{Capacity is not the bottleneck.}
Ablation B scales the same architecture, data and objective across six points from $5.1$\,M to $571.6$\,M parameters (a $112\times$ range) and direction accuracy never improves: $0.7397$ at $5.1$\,M, $0.7333$ at $115.0$\,M, $0.7097$ at $204.1$\,M, $0.6853$ at $571.6$\,M. Individual adjacent gaps here are comparable to the $0.024$ seed floor, so we do not claim that any one pair is separated; the finding is the \emph{absence of improvement across two orders of magnitude} together with outright divergence at the top, which no seed effect explains. The smallest model is the best on both mean and median cosine and ties for the most characters above $0.80$, and the trend beyond $115$\,M is degradation; interestingly the largest converged models are the best-calibrated on amplitude ($1.008$ at $571.6$\,M, $0.975$ at $38.7$\,M), so scale buys magnitude at the cost of direction. At $1.0$\,B the run diverged outright (validation cosine $-0.085$), so the extra capacity destabilised optimisation rather than helping it. Training loss keeps falling as models grow while held-out cosine does not, which identifies the gap as generalisation rather than under-fitting. We attribute the ceiling to three structural causes, none of which more parameters can fix. First and most fundamentally, the task is \emph{ambiguous}: one still image plus one parameter value is consistent with many plausible artist motions, and a pointwise regression loss can only recover a central tendency of that conditional distribution, the median for the $L_1$ objective we use, the mean for an $L_2$ one. Averaging over plausible motions is exactly what produces amplitude compression on the large turns where artists disagree most, and a larger model predicts that same central tendency more precisely rather than escaping it. Second, Stage~1 decomposition quality caps what Stage~2 can express. Third, rig-data diversity is limited: $1{,}443$ characters is large for this domain but small in absolute terms, and the $8$ parameters we model are the common subset rather than the full Cubism vocabulary. The levers that remain are therefore a distributional output (sampled, probabilistic, or flow-matching) instead of a pointwise regression, better layers, and more diverse rigs, not a bigger network.

\paragraph*{Conditioning signals that did not help.}
Beyond the ablations above we tried four further conditioning signals, three of them suggested by reviewers of an earlier version, and none exceeded the seed noise floor: explicit per-vertex draw-order conditioning ($0.7148$), a zero-initialised 2.5D parallax residual (two seeds, $0.7225$ and $0.7170$), the two combined at $40$ epochs ($0.7153$), and per-vertex image-feature sampling in place of one pooled vector per layer ($0.6918$). Two baseline seeds give $0.7289$ and $0.7533$ on the same benchmark, so the noise floor is $0.024$ and none of these moves the mean beyond it. The parallax variants are the interesting near-miss: parallax with draw order reaches the best median ($0.8255$), the most characters above $0.80$ ($25/46$) and by far the best amplitude ($1.054$ against $1.238$), i.e.\ it improves the typical character and the amplitude while losing badly on a few, which is the one negative result that points at our amplitude-compression failure mode. Full table, per-variant reasoning, and a measurement error we made and corrected are in Appendix~\ref{app:negative}.

\paragraph*{Amplitude reweighting is a wash.}
Since amplitude is the weaker axis, we tried the obvious fix of up-weighting large-displacement layers in the loss (ablation C). It does not help: $\rho\!=\!0.25$ leaves direction unchanged ($0.7392$ vs.\ $0.7397$) and pushes the magnitude ratio further above target ($1.150$ vs.\ $1.130$), and $\rho\!=\!0.5$ degrades direction to $0.7301$ while overshooting badly ($1.247$). The reason is visible in the baseline number itself: aggregate amplitude is already slightly \emph{above} $1.0$, so a global up-weighting overshoots. The deficit is specific to large turns, and an aggregate objective gives no credit for fixing only those. A correct fix has to be conditional on motion scale rather than a global reweighting; we leave it open.

\begin{figure*}[t]
  \centering
  \includegraphics[width=\linewidth]{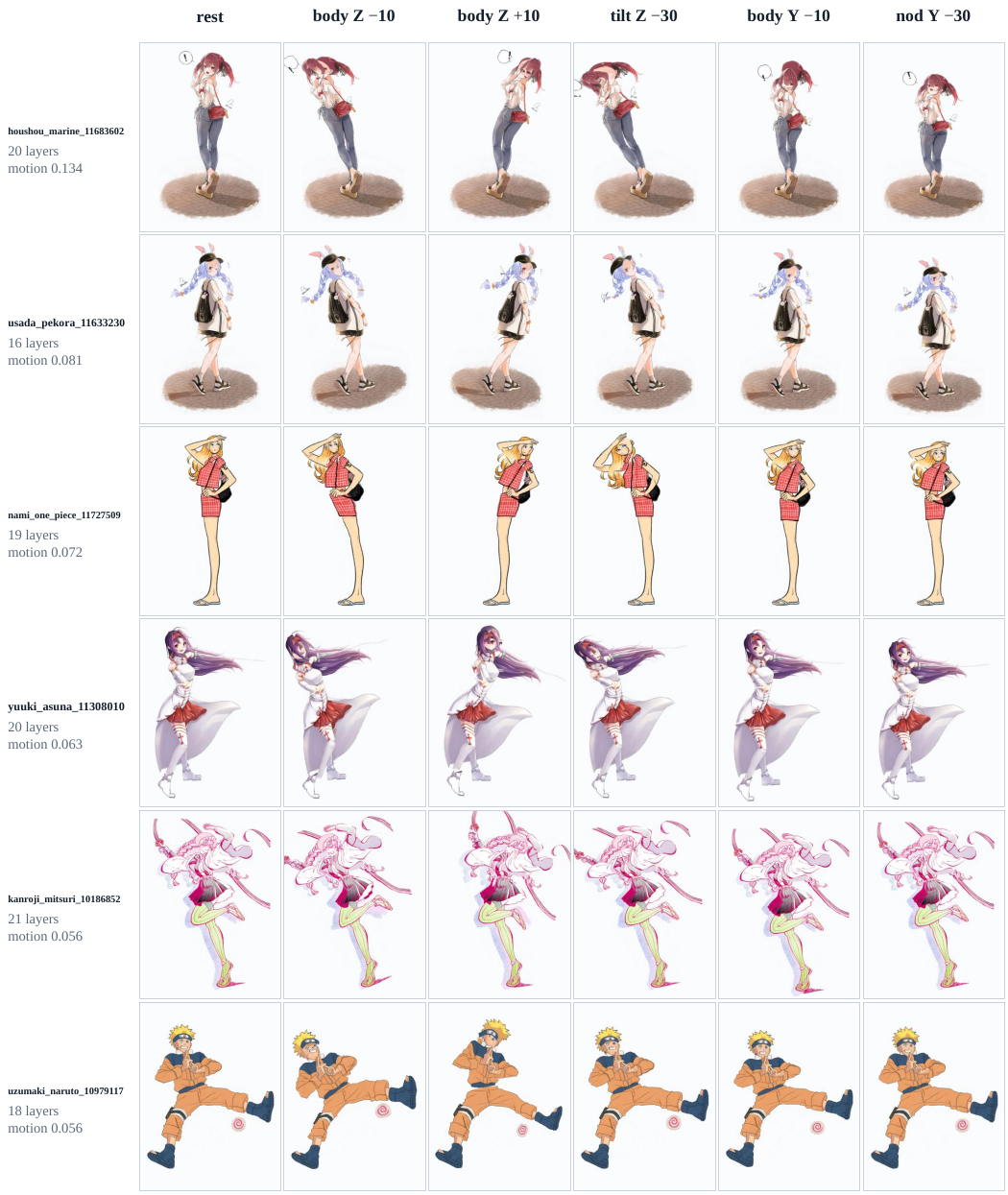}
  \caption{\textbf{In-the-wild rigs, raw output.} Web illustrations that were never Live2D models, each
 decomposed by Stage~1 and animated by Stage~2 at six \emph{absolute} parameter values, so every cell is
 reproducible from the released rig. Rows are ordered by measured motion magnitude (printed beside each
 row). One frozen checkpoint, one triangulation procedure and one parameter sweep produced every cell. The
 remaining characters are in Appendix~\ref{app:itw_gallery}.}
  \Description{Five in-the-wild characters shown at rest and under five different animation parameters.}
  \label{fig:itw_poses}
\end{figure*}

Two things in Fig.~\ref{fig:itw_poses} are worth looking for. In the crowded cases the coordination
holds: under tilt~Z the hood, hair, face and held plush of \texttt{test5} rotate as one object rather than
shearing against each other, and under turn~X the fringe stays attached to the forehead, which is
precisely the failure mode per-layer prediction produces (\S\ref{sec:results_ablations}). The remaining
visible weakness is amplitude: large turns come out smaller than an artist would draw them
(\S\ref{sec:conclusion}).

\subsection{Scaling the parameter vocabulary: $8 \to 24$}
\label{sec:results_p24}

Everything above models the $8$ Cubism parameters that almost every rig in the corpus declares. That
subset covers head turn, nod and tilt, body sway and eye blink, but it is not a usable idle animation:
a character that never moves its gaze, brows, hair or chest reads as frozen. We therefore ask whether
the same architecture absorbs a three-times larger parameter vocabulary, and what breaks when it does.

\paragraph*{Data.}
We re-extracted the corpus keeping every parameter with at least $200$ layer records, which yields
$24$ parameters over $1{,}306$ characters and $449{,}157$ layer records: the original $8$, plus gaze
(\texttt{EyeBallX/Y}), eye smile, mouth form, four brow controls per side, three hair-sway groups and
breathing. \texttt{ParamCheek} had only $28$ records corpus-wide and was dropped rather than shipped
broken. The held-out character list is copied verbatim from the $8$-parameter split, and we verified
that the re-extraction is faithful: over the $1{,}211$ characters present in both packs the base vertex
positions agree to a median $\max|\Delta|$ of $1.0\!\times\!10^{-6}$.

\paragraph*{The vocabulary triples but the capability does not.}
Table~\ref{tab:p24_tiers} groups the $24$ parameters by the quality they reach, and the grouping is
stark: $11$ parameters land above a direction cosine of $0.70$, $5$ sit between $0.40$ and $0.70$, and
$8$ never become usable. The winners include the two we most wanted, gaze at $0.999$ and $0.997$,
together with breathing at $0.749$ and brow height at $0.84$, so the extension does deliver a
complete idle loop. The losers are equally specific: brow \emph{form} and brow \emph{angle}, mouth
form, and side-hair sway. Table~\ref{tab:p24_perparam} in Appendix~\ref{app:p24} lists all $24$ with
their training support, and the ordering is explained almost entirely by that one column. Brow angle
has $17$ to $18$ moving-layer samples against head turn's $2{,}841$; the two parameters with negative
cosine are the two rarest in the corpus. This is a long-tail data problem, not a capacity problem,
which is consistent with ablation~B: the same architecture scaled $112\times$ does not improve the
parameters it already has, so it will not rescue the ones it barely sees. We report the failures
rather than pruning the vocabulary to the $16$ that work, because a reader deciding whether to
model $24$ parameters needs to know which ones the data can currently support.

One weakness is \emph{not} caused by the extension. \texttt{MouthOpenY} scores $0.239$ in the
$8$-parameter model and $0.235$ in the $24$-parameter one: mouth opening was always our worst common
parameter. Its motion is dominated by a few strongly deforming layers whose displacement is close to a
pure anisotropic scale, which the peak-normalised direction target of Eq.~\ref{eq:stage2_heads}
represents poorly, and no amount of extra parameter coverage changes that.

\begin{table}[t]
  \centering
  \caption{\textbf{Parameter-vocabulary extension, capability tiers.} The $24$-parameter model on $46$ held-out characters, true generation. Tiers are by per-parameter direction cosine; $\ast$ marks a parameter the $8$-parameter model does not have. The full per-parameter table with training support is Tab.~\ref{tab:p24_perparam}.}
  \label{tab:p24_tiers}
  \scriptsize
  \setlength{\tabcolsep}{3pt}
  \renewcommand{\arraystretch}{1.1}
  \begin{tabular*}{\linewidth}{@{\extracolsep{\fill}}llc@{}}
    \toprule
    Tier & Parameters & count \\
    \midrule
    good, $\geq\!0.70$ & \texttt{EyeBallY}$\ast$ \texttt{EyeBallX}$\ast$ \texttt{BodyAngleZ} \texttt{AngleZ} & \\
                       & \texttt{BrowRY}$\ast$ \texttt{BrowLY}$\ast$ \texttt{EyeROpen} \texttt{EyeLOpen} & $11$ \\
                       & \texttt{Breath}$\ast$ \texttt{AngleY} \texttt{AngleX} \texttt{EyeRSmile}$\ast$ & \\
    \midrule
    marginal, $0.40$--$0.70$ & \texttt{EyeLSmile}$\ast$ \texttt{HairBack}$\ast$ \texttt{BodyAngleY}$\ast$ & $5$ \\
                             & \texttt{BodyAngleX} \texttt{HairFront}$\ast$ & \\
    \midrule
    unusable, $<\!0.40$ & \texttt{HairSide}$\ast$ \texttt{BrowLForm}$\ast$ \texttt{BrowRForm}$\ast$ & \\
                        & \texttt{MouthOpenY} \texttt{MouthForm}$\ast$ & $8$ \\
                        & \texttt{BrowRAngle}$\ast$ \texttt{BrowLAngle}$\ast$ & \\
    \bottomrule
  \end{tabular*}
\end{table}

\paragraph*{Warm-starting from the small vocabulary recovers most of the loss.}
Trained from scratch on $24$ parameters the model spends capacity on the tail and gives ground on the
original $8$. Copying the $8$ shared parameter-embedding rows from the $8$-parameter checkpoint
\emph{by parameter name}, leaving the $16$ new rows at random initialisation, and then training on all
$24$ is strictly better on the axis that matters (Tab.~\ref{tab:p24_warm}): the eight warm-started
parameters hold level ($0.6983$ against $0.7009$ cold, inside seed noise) while the $16$ new ones rise
from $0.4792$ to $0.5588$, a $17\%$ relative gain. The gain concentrates exactly where support is thinnest,
\texttt{EyeRSmile} $0.372\!\to\!0.724$ on $17$ samples and \texttt{EyeLSmile} $0.561\!\to\!0.710$ on
$16$, which is what a transfer explanation predicts: a rare parameter cannot learn a general
displacement field from $17$ examples, but it can learn to \emph{re-index} a field the shared trunk
already represents. Warm-starting is what we release.

\begin{table}[t]
  \centering
  \caption{\textbf{Warm-start ablation on the $24$-parameter vocabulary.} Per-parameter direction cosine averaged within groups, $46$ held-out characters, true generation. ``Cold'' trains all $24$ from scratch; ``warm'' copies the $8$ shared parameter embeddings from the $8$-parameter checkpoint by name and trains identically thereafter. The split is exactly the two released vocabularies: the $8$ rows that exist in the $8$-parameter checkpoint against the $16$ that do not and are therefore randomly initialised in both runs.}
  \label{tab:p24_warm}
  \scriptsize
  \setlength{\tabcolsep}{3pt}
  \renewcommand{\arraystretch}{1.1}
  \begin{tabular*}{\linewidth}{@{\extracolsep{\fill}}lccc@{}}
    \toprule
    Parameter group & \#\,params & cold & warm \\
    \midrule
    warm-started ($8$ shared)                     & $8$  & $\mathbf{0.7009}$ & $0.6983$ \\
    newly added                                   & $16$ & $0.4792$ & $\mathbf{0.5588}$ \\
    \midrule
    all                                           & $24$ & $0.5531$ & $\mathbf{0.6053}$ \\
    \bottomrule
  \end{tabular*}
\end{table}

\paragraph*{A metric caveat we had to resolve: per-vertex cosine is not invariant to mesh density.}
The $8$-parameter model reports $0.7676$ and the $24$-parameter model $0.689$, and the tempting
reading, that tripling the vocabulary costs $0.079$, is wrong. Three things changed between those two
evaluations, and only one of them is the model. Table~\ref{tab:p24_controlled} locks them one at a
time. Four held-out characters are absent from the re-extracted pack, worth $0.013$. The
re-extraction also re-meshed $24\%$ of layers more densely, mean vertex count per layer rising from
$47.5$ to $71.9$ and never falling, worth a further $0.063$, which is twice the size of the effect
being studied. Only the last row isolates the model, and there the true cost of the extra $16$
parameters is $0.031$; magnitude calibration in fact \emph{improves}, $|\text{mag}\!-\!1|$ falling from
$0.065$ to $0.047$.

We verified the density term rather than inferring it. Scoring the same model on only the $1{,}273$
layers whose vertex count is \emph{identical} in both packs gives $0.7161$ on one pack and $0.7150$ on
the other, equal to within $0.001$, while the $409$ densified layers score $0.6822$. Everything else
was excluded by direct comparison: the two packs share one physical DINOv2 token file with
$57{,}972/57{,}972$ identical row indices (sampled per-layer feature agreement $1.2\!\times\!10^{-7}$),
parameter ranges are identical, layers per pose differ by a median of $0$, base vertices by
$3.1\!\times\!10^{-6}$ and ground-truth offsets by $4.8\!\times\!10^{-8}$. The mechanism is simple in
hindsight: a denser mesh adds interior vertices whose true displacement is small and whose
\emph{direction} is therefore near-degenerate, yet they still clear the $\|\Delta v\|\!>\!0.005$ gate
and enter the average. \textbf{Any per-vertex cosine comparison across meshes of different density is
confounded}, which affects how our own mesh ablation must be read (\S\ref{sec:results_stage2}) and, we
expect, any future work that adopts this metric. Checkpoints should be compared only on one fixed pack,
character set and parameter set; we added the three switches that make this enforceable to our
evaluation script and report them in Appendix~\ref{app:p24}.

\begin{table}[t]
  \centering
  \caption{\textbf{Why $0.7676$ and $0.689$ are not comparable.} Each row changes exactly one thing from the row above, on the shared held-out characters and the shared $8$ parameters, true generation throughout. Mesh density alone accounts for twice as much as the model change under study.}
  \label{tab:p24_controlled}
  \scriptsize
  \setlength{\tabcolsep}{3pt}
  \renewcommand{\arraystretch}{1.1}
  \begin{tabular*}{\linewidth}{@{\extracolsep{\fill}}llccc@{}}
    \toprule
    Setting changed & model & chars & dir-cos & $\Delta$ \\
    \midrule
    as reported in Tab.~\ref{tab:stage2_mesh}      & $8$p & $50$ & $0.7676$ & -- \\
    $-$ $4$ chars absent from the $24$p pack       & $8$p & $46$ & $0.7543$ & $-0.013$ \\
    $+$ denser re-meshing ($24\%$ of layers)       & $8$p & $46$ & $0.6913$ & $-0.063$ \\
    \midrule
    \textbf{only the model differs}                & $24$p cold & $46$ & $\mathbf{0.6608}$ & $\mathbf{-0.031}$ \\
    same, warm-started (released)                  & $24$p warm & $46$ & $0.6866$ & $+0.026$ \\
    \bottomrule
  \end{tabular*}
\end{table}

\paragraph*{Released models.}
We release both versions, because they are not a replacement pair. The $8$-parameter model is the more
faithful one on the motions it covers ($0.6913$ against $0.6866$ on identical data and parameters) and
is what the headline numbers of this paper refer to; the $24$-parameter warm-started model is the one
that produces a complete idle animation. Each release carries its weights, its parameter vocabulary
(whose list order defines the embedding index), an architecture record, and a model card stating
per-parameter quality including the $8$ parameters that do not work. The reference inference script is
verified numerically identical to the pipeline these metrics were computed with: on a $17$-layer
in-the-wild character all $1{,}634$ base vertices and all $66$ (parameter, keypose) frames match to
$0.0000$\,px.

\subsection{Editability and outfit swap}
\label{subsec:editing}
\label{sec:results_editability}
Our output remains editable per layer, mesh edge, and keyframe; replacing a layer, changing vertex offsets, or swapping a motion template does not require re-running either stage. \begin{figure*}[t]
  \centering
  \includegraphics[width=\linewidth]{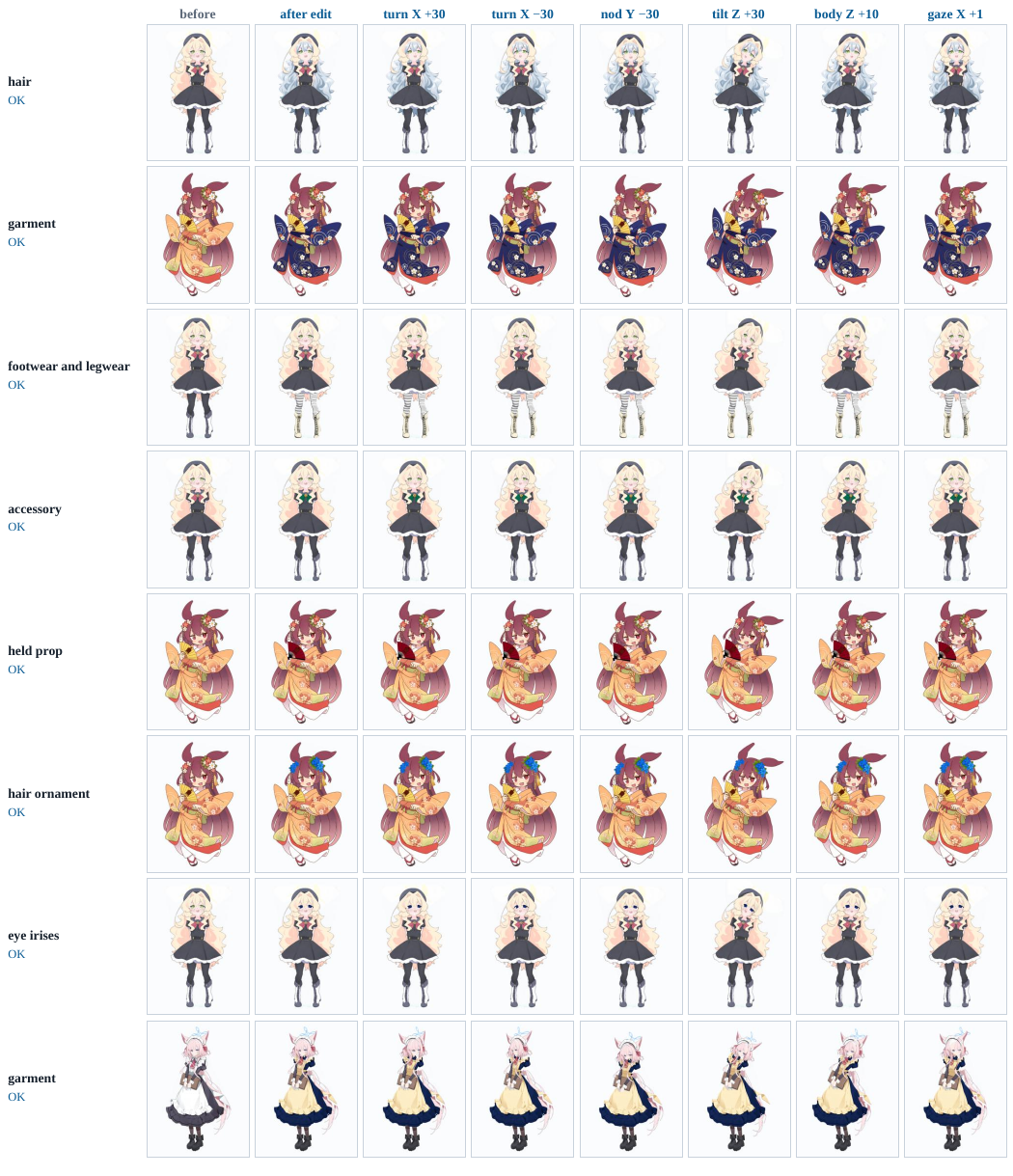}
  \caption{\textbf{Prompt-driven texture editing with zero re-animation.} Each row is one edit: the rig before,
 after repainting one to five layers' RGB from a natural-language instruction, and the \emph{same} rig
 driven to six absolute parameter values. Every pose column replays the stored displacement frames on the
 new texture rather than re-generating them. Sheet~2 is Fig.~\ref{fig:outfit_swap_2}; the verification and
 the category list are in the text.}
  \Description{Eight texture edits on five characters, each shown before, after, and after being driven to six poses.}
  \label{fig:outfit_swap}
\end{figure*}

\begin{figure*}[t]
  \centering
  \includegraphics[width=\linewidth]{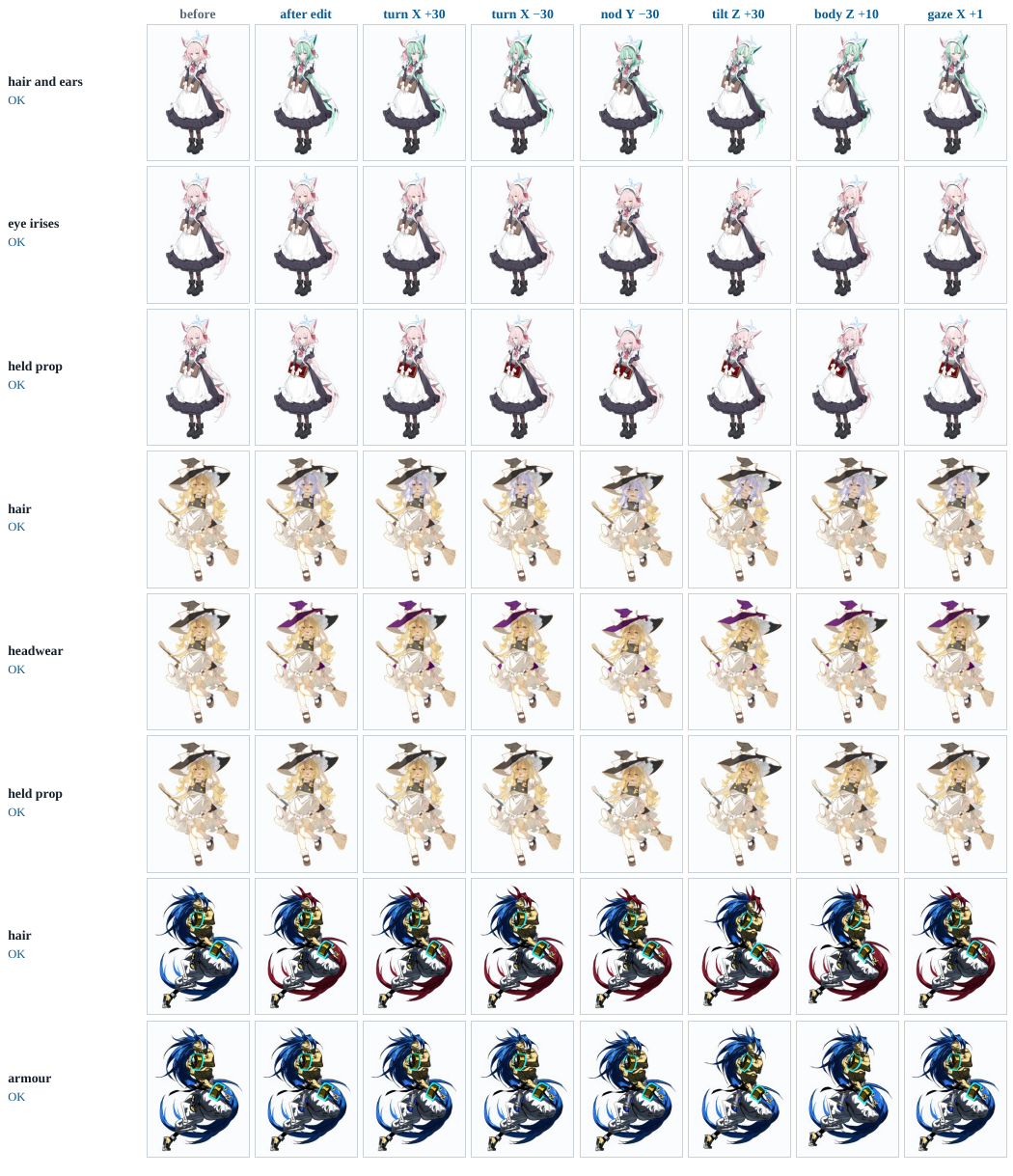}
  \caption{\textbf{Prompt-driven texture editing, sheet~2.} Format and verification as
  Fig.~\ref{fig:outfit_swap}. These eight edits are on the three characters added to test whether the
  invariance is a property of the representation or of two lucky rigs: a maid outfit (garment, hair with
  animal ears over five layers, irises, held prop), a witch (hair, headwear, broom), and a
  heavily armoured fighter (hair over four layers, gauntlet).}
  \Description{Eight further texture edits on three characters, each shown before, after, and after being driven to six poses.}
  \label{fig:outfit_swap_2}
\end{figure*}

Because a layer's mesh UVs derive from its alpha bounding box, an edit that restores the original
alpha and pixel size cannot invalidate the mesh and therefore cannot invalidate the animation. We check
that mechanically rather than by eye: per row, the vertex positions, the triangle index lists and
\emph{every} predicted displacement frame are byte-identical between the source and the edited rig, which
over the $16$ edits of Figs.~\ref{fig:outfit_swap} and~\ref{fig:outfit_swap_2} amounts to $323$ layers,
$31{,}508$ vertices, $50{,}762$ triangles and $21{,}318$ displacement frames compared, with zero
differences. The categories deliberately go beyond clothing: hair, hair with animal ears, a hair ornament,
garment, footwear with legwear, headwear, armour, an accessory, a held prop, and the eye irises, the last
being the hardest case because those layers are only $30\!\times\!20$ and $44\!\times\!15$ pixels.

\paragraph*{Prompt-driven outfit swap (optional Stage~3).}
Editability is easy to claim and hard to demonstrate, so we instantiate it as a concrete capability. Because a layer's mesh UVs are derived from that layer's alpha bounding box, an edit that leaves the alpha channel and the pixel dimensions untouched cannot invalidate the mesh, and therefore cannot invalidate the animation either. We use this to change a character's outfit: one clothing layer's RGB is repainted by an instruction-guided image editor ~\cite{qwenimageedit2511} from a natural-language instruction, the original alpha and size are restored, and the rig is re-exported with the new texture and \emph{no} re-animation and \emph{no} re-decomposition. Figs.~\ref{fig:outfit_swap} and~\ref{fig:outfit_swap_2} show all $16$ edits we ran, on $5$ characters and spanning $10$ layer categories, repainting $31$ layers in total. We checked the invariance mechanically rather than by eye: comparing each edited rig against its source, the vertex positions, the triangle index lists and every predicted displacement frame are byte-identical, which across the $16$ edits is $323$ layers, $31{,}508$ vertices, $50{,}762$ triangles and $21{,}318$ frames compared with zero differences, and every edited texture retains its exact pixel dimensions and an identical alpha channel; the only field that changes anywhere in the file is the texture path. This is what an editable structured asset buys that a video cannot: a downstream appearance change costs one image edit rather than a re-generation.

\begin{figure*}[t]
  \centering
  \includegraphics[width=\linewidth]{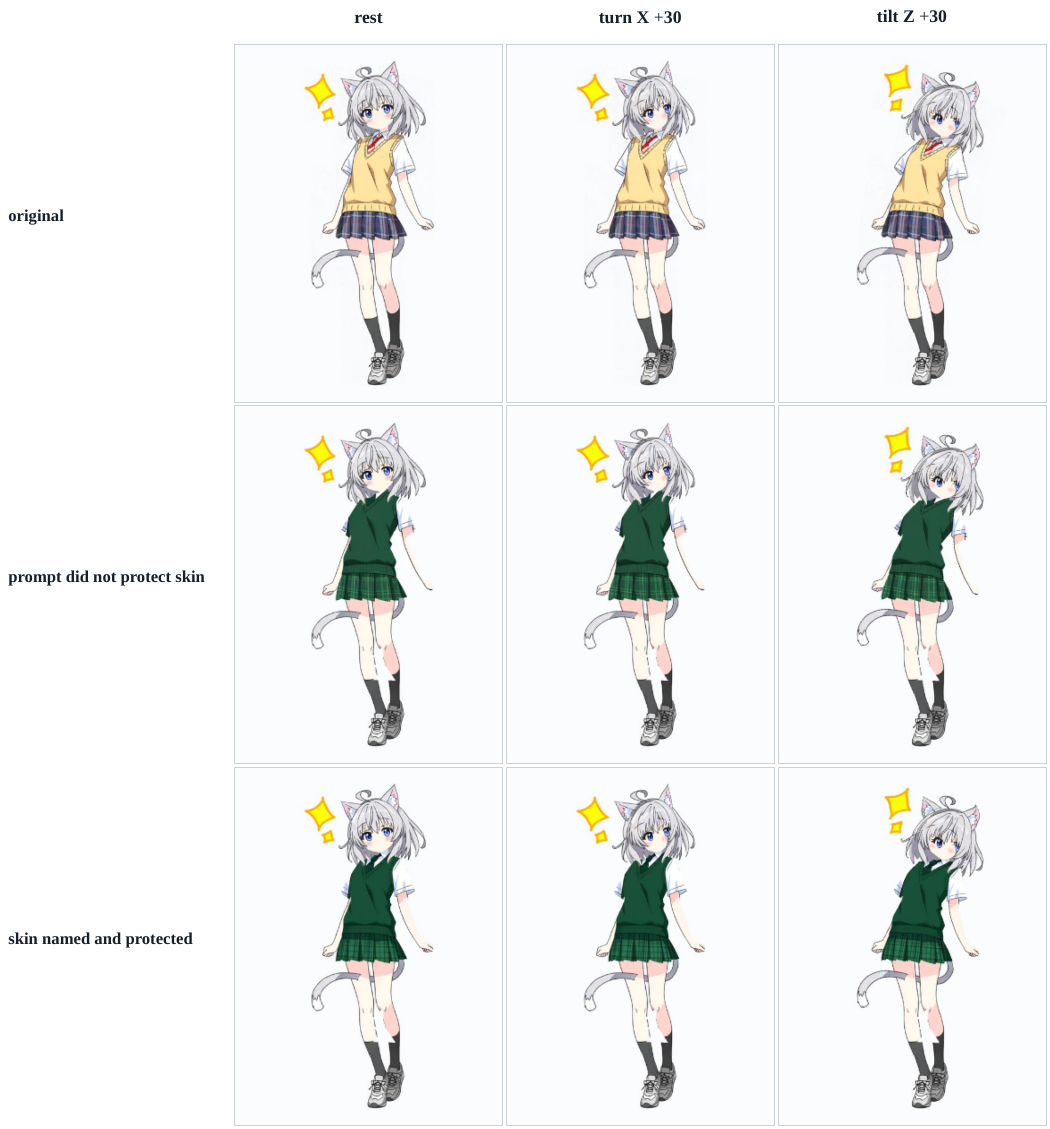}
  \caption{\textbf{Honest re-skin failure and its fix.} On \texttt{q\_webA\_catgirl} a single Stage-1 layer holds both the school uniform and the bare arm. Asking only for a green uniform (centre) desaturated the arm and hand to near-white; naming and protecting the skin colour in the instruction (right) preserved them. An edit can only be as local as the layer it lives in, so this failure is inherited from decomposition granularity rather than from the editing step. The right-hand version is the one used in Fig.~\ref{fig:outfit_swap}.}
  \Description{Original garment layer, a failed re-texture that bleached the arm, and the corrected re-texture.}
  \label{fig:reskin_failure}
\end{figure*}

Two honest limitations. When a garment layer shares pixels with skin (a short sleeve whose layer also contains the bare arm) the editor can desaturate the skin along with the cloth unless the instruction explicitly names and protects the skin colour; our first attempt on one character did exactly that (Fig.~\ref{fig:reskin_failure}), and a skin-protecting instruction fixed it, which means the failure is a prompting failure rather than an architectural one but still a failure a user can hit. And because alpha is deliberately held fixed, only edits that preserve the garment's silhouette are in scope: turning a skirt into trousers changes the alpha support, invalidates the mesh, and requires re-running the pipeline.

\paragraph*{What Live2D-Bench measures.}
The benchmark is deliberately stricter than evaluating a rendered frame. A plausible composite can still be unusable if it merges front hair with the face, omits occluded sleeves, emits no mesh, or bakes motion into pixels rather than editable keyposes; conversely, a mesh predictor can look accurate under oracle layers while failing once the layer alpha support changes. Live2D-Bench therefore exposes the full asset interface: ordered RGBA layers, per-layer artist meshes, parameter ids, keypose values, vertex-offset fields, and rendered Body / Face Angle loops. Its $10$ to $20$, $20$ to $35$, and $35{+}$ layer bins test semantic separation, the common production regime, and ordering / occlusion pressure, while the non-human split checks tails, ears, wings, plush bodies, and accessories. This unified protocol distinguishes a Live2D asset from a movie of a Live2D-like character and makes the benchmark a reusable target for future complete Live2D-generation systems.

\section{Conclusion}
\label{sec:conclusion}

We have presented the first end-to-end system that turns one illustration into a structured, editable, animatable Live2D asset. Stage~1 decomposes the illustration into Live2D-aware RGBA layers with hidden-region completion. Stage~2 builds a content-conforming triangle mesh per layer and regresses per-vertex keypose displacements for \emph{all} layers of a character jointly, in one forward pass, so that self-attention can coordinate layers instead of letting each drift independently; on $50$ held-out characters it reaches a per-vertex direction cosine of $0.7676$ (median $0.8278$) under true generation, with $31$ of $50$ characters above $0.80$. An optional third step re-textures a single clothing layer from a natural-language instruction and reuses the predicted animation verbatim. Together with Live2D-Bench and an $8{,}884$-model corpus, the results show that single-image Live2D generation is tractable as a learnt structured-asset problem.

\paragraph*{Limitations.}
Several limitations are worth stating precisely, because each points at a different missing piece rather than at a tuning deficiency.

\emph{Amplitude compression on large motion.} Aggregate amplitude is well calibrated (the median magnitude ratio is close to $1$) but the error is not uniform across motion scale: small displacements are slightly over-shot and large head or body turns are under-shot. This is the expected behaviour of a pointwise regression objective on an ambiguous task, since one still image and one parameter value admit many plausible artist motions and an $L_1$ loss can only recover a central tendency of that distribution. Scaling the model $40\times$ does not fix it (Tab.~\ref{tab:stage2_ablations}) and neither does globally up-weighting amplitude in the loss; a distributional output (sampling, or a flow-matching / diffusion head over displacement fields) is the principled remedy and is our main planned direction.

\emph{Animated draw order is not modelled, and conditioning on static order does not help.} We tested the obvious remedy of feeding each vertex its layer's normalised draw-order position explicitly; it scored below both baseline seeds (Appendix~\ref{app:negative}), which suggests the ordering is already recoverable from layer identity plus the shared canvas. We predict a static layer ordering and deform within it. Real rigs sometimes keyframe the draw order itself, so that a hair strand passes in front of a face partway through a turn, or an arm crosses the torso. Nothing in our formulation expresses an order that changes with a parameter value, so those transitions are simply unavailable; adding a per-layer, per-keypose depth channel is a natural extension.

\emph{Texture edits cannot change silhouette.} The outfit-swap step preserves the animation precisely because it holds a layer's alpha channel and pixel dimensions fixed. That is also its boundary: edits that change the garment's outline invalidate the mesh support and require re-running decomposition and animation. A related failure is that when a garment layer shares pixels with skin, the editor may alter the skin unless the instruction protects it explicitly.

\begin{figure*}[t]
  \centering
  \includegraphics[width=\linewidth]{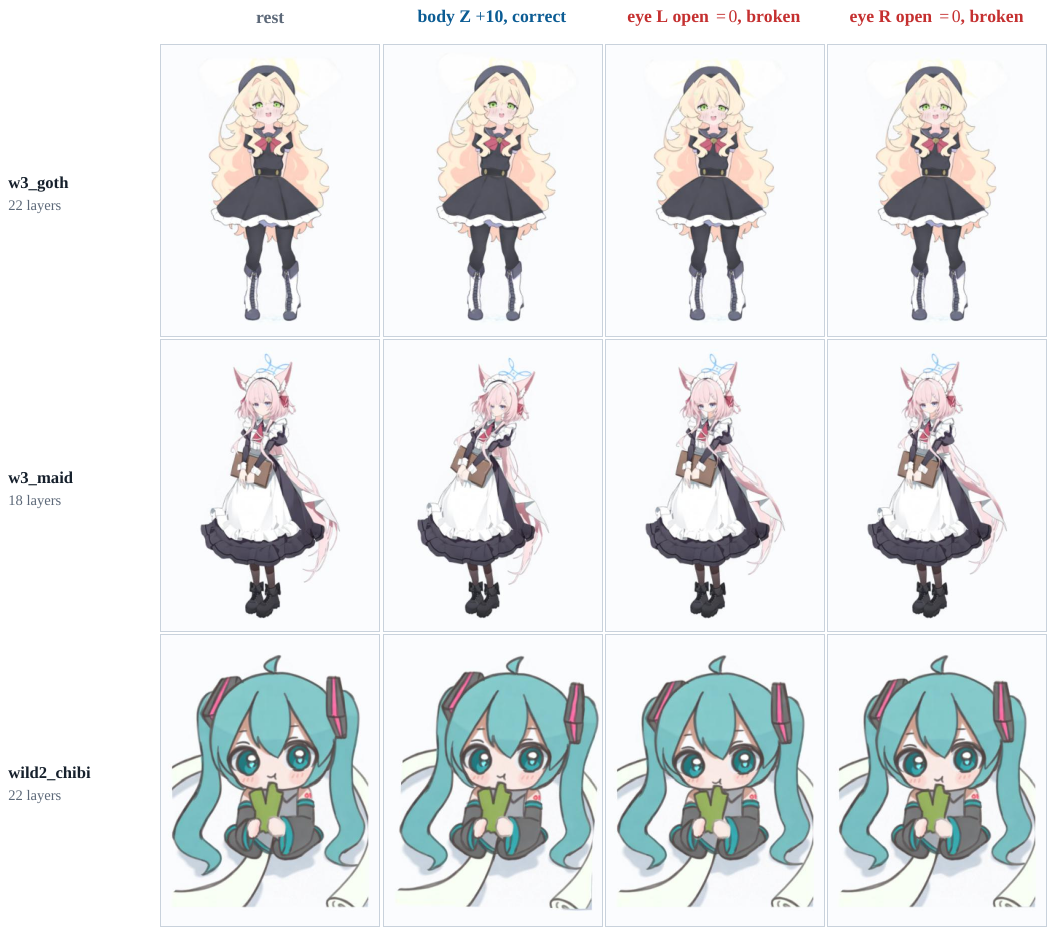}
  \caption{\textbf{Failure mode: the eye-open parameter on auto-decomposed layers.} Three in-the-wild rigs at rest (left), under $\texttt{ParamBodyAngleZ}$ at its maximum (centre, correct), and with $\texttt{ParamEyeLOpen}\!=\!\texttt{ParamEyeROpen}\!=\!0$ (right, broken). The blink displaces the whole head group rather than the eyelids: the face slides out from under the hair, the skin layer translates off the head, or the head shrinks and rotates. Because the centre column of each row is the same rig under a different parameter and is correct, the mesh and the rig are sound and the fault is specific to the eye channel. Raw renders, no per-example fix.}
  \Description{Three characters shown at rest, under a correct body-sway pose, and under a broken eye-close pose.}
  \label{fig:eye_param_failure}
\end{figure*}

\emph{The eye-open channel is unreliable on auto-decomposed layers.} Driving $\texttt{ParamEyeLOpen}\!=\!\texttt{ParamEyeROpen}\!=\!0$ on an in-the-wild rig displaces the whole head group instead of only the eyelid layers (Fig.~\ref{fig:eye_param_failure}); the same rigs are correct under the head-angle, body-sway and mouth parameters, so this is specific to that channel. Two causes compound: our Stage~1 does not reliably emit the eyelid as a separate layer, so there is no layer for the motion to attach to, and in training the two eye parameters contribute roughly $6$k deformation records each against roughly $91$k for each head angle, with the smallest displacements of any parameter, so the channel is also the least supervised. On artist-authored stacks, where the eyelid \emph{is} its own layer, the parameter behaves. Isolating eyelid layers in Stage~1 and up-sampling eye records in training are both straightforward next steps, and we exclude blink from the in-the-wild pose sweeps in the supplementary material rather than showing it silently.

\emph{Background props are decomposed and animated like body parts.} Stage~1 has no notion of what belongs to the character, so a sheet of paper behind a chibi, a held plush toy, or a floating motif becomes a layer, receives a mesh, and is deformed by head-turn parameters that should not touch it. A character/background gate before Stage~1 would address this and is not part of the pipeline evaluated here.

\emph{Dependence on Stage~1.} Stage~2 can only animate the layers it is given. A layer cut through a sleeve, or one whose occluded region was completed poorly, produces artefacts no matter how accurate the displacement field is, and our cross-decomposer experiment shows the end-to-end gap on in-the-wild art is dominated by decomposition rather than by motion error (Appendix~\ref{app:crossdecomp}).

\emph{The long tail of the parameter vocabulary is not learnable from present data.} Extending the model from $8$ to $24$ Cubism parameters (\S\ref{sec:results_p24}) produces $11$ parameters at usable quality and $8$ that are unusable, and the split is explained by training support rather than by anything about the motion: brow angle has $17$ to $18$ moving-layer samples in the whole corpus against head turn's $2{,}841$, and the two parameters with negative direction cosine are the two rarest. Warm-starting from the small-vocabulary model recovers a substantial part of the gap ($0.4792 \to 0.5588$ on the $16$ new parameters) precisely because a rare parameter can re-index a displacement field the shared trunk already represents, but it cannot manufacture the missing supervision. Brow expression and mouth form therefore remain out of reach until the corpus grows in those channels, and we release the failures visibly rather than pruning the vocabulary to what happens to work.

\emph{Our own primary metric has a confound we had to discover.} Per-vertex direction cosine is not invariant to mesh density: a denser mesh contributes interior vertices whose true displacement is small and whose direction is consequently near-degenerate, yet which still clear the movement threshold and enter the average. We measured this at $0.063$ cosine for a $24\%$ re-meshing that raised mean vertex count from $47.5$ to $71.9$, which is twice the size of the model effect we were studying at the time (\S\ref{sec:results_p24}). Any per-vertex comparison across different meshes is therefore confounded, including our own mesh ablation, and Live2D-Bench numbers should be quoted with their pack, character list and parameter list attached. A density-normalised or area-weighted variant of the metric would be a genuine contribution and we do not have one.

\emph{Scope.} Physics dampers, expression blending, and viseme parameters are runtime extensions we do not learn; the Cubism binary format is encrypted, so we ship an open asset bundle plus our own renderer rather than a \texttt{.moc3}; and our training distribution skews toward Japanese anime aesthetics, leaving photorealistic, 3D-shaded, and multi-character inputs out of scope. The released data, benchmark, and renderer provide a basis for addressing these.

\clearpage
\bibliographystyle{ACM-Reference-Format}
\bibliography{sample-base}

@article{yang2025omnisvg,
  title={Omnisvg: A unified scalable vector graphics generation model},
  author={Yang, Yiying and Cheng, Wei and Chen, Sijin and Zeng, Xianfang and Yin, Fukun and Zhang, Jiaxu and Wang, Liao and Yu, Gang and Ma, Xingjun and Jiang, Yu-Gang},
  journal={Advances in Neural Information Processing Systems},
  volume={38},
  pages={113670--113696},
  year={2026}
}

@article{jacobson2011bbw,
  title={Bounded biharmonic weights for real-time deformation.},
  author={Jacobson, Alec and Baran, Ilya and Popovic, Jovan and Sorkine, Olga},
  journal={ACM Trans. Graph.},
  volume={30},
  number={4},
  pages={78},
  year={2011}
}

@article{blattmann2023stablevideo,
  title={Stable video diffusion: Scaling latent video diffusion models to large datasets},
  author={Blattmann, Andreas and Dockhorn, Tim and Kulal, Sumith and Mendelevitch, Daniel and Kilian, Maciej and Lorenz, Dominik and Levi, Yam and English, Zion and Voleti, Vikram and Letts, Adam and others},
  journal={arXiv preprint arXiv:2311.15127},
  year={2023}
}

@inproceedings{soni2023deeplearninglipsync,
  title={Deep Learning Technique to generate lip-sync for live 2-D Animation},
  author={Soni, Ashish and Deshmukh, Janhvi and Shende, Ayush and Gawande, Ramanshu and Agatkar, Priti and Chilbule, Kshitija},
  booktitle={2023 IEEE International Students' Conference on Electrical, Electronics and Computer Science (SCEECS)},
  pages={1--7},
  year={2023},
  organization={IEEE}
}

@inproceedings{cao2025uni3c,
  title={Uni3C: Unifying Precisely 3D-Enhanced Camera and Human Motion Controls for Video Generation},
  author={Cao, Chenjie and Zhou, Jingkai and Li, Shikai and Liang, Jingyun and Yu, Chaohui and Wang, Fan and Xue, Xiangyang and Fu, Yanwei},
  booktitle={Advances in Neural Information Processing Systems (NeurIPS)},
  year={2025}
}

@article{tencent2025hunyuanvideoi2v,
  title={Hunyuanvideo: A systematic framework for large video generative models},
  author={Kong, Weijie and Tian, Qi and Zhang, Zijian and Min, Rox and Dai, Zuozhuo and Zhou, Jin and Xiong, Jiangfeng and Li, Xin and Wu, Bo and Zhang, Jianwei and others},
  journal={arXiv preprint arXiv:2412.03603},
  year={2024}
}

@misc{chen2026paircoder,
      title={PairCoder++: Pair Programming as a Universal Paradigm for Verified Code-Driven Multimodal and Structured-Artifact Generation},
      author={Junhao Chen and Xiang Li and Mingjin Chen and Boran Zhang and Henghaofan Zhang and Yibin Xu and Yuehan Cui and Fangsheng Weng and Fei Ma and Qi Tian and Ruqi Huang and Hao Zhao},
      year={2026},
      eprint={2607.01883},
      archivePrefix={arXiv},
      primaryClass={cs.CL},
      url={https://arxiv.org/abs/2607.01883},
}

@inproceedings{chen2025svgthinker,
  title={SVGThinker: Instruction-Aligned and Reasoning-Driven Text-to-SVG Generation},
  author={Chen, Hanqi and Zhao, Zhongyin and Chen, Ye and Liang, Zhujin and Ni, Bingbing},
  booktitle={Proceedings of the 33rd ACM International Conference on Multimedia},
  pages={11004--11012},
  year={2025}
}

@InProceedings{Chen_2026_CVPR_hvg3d,
    author    = {Chen, Mingjin and Chen, Junhao and Fan, Zhaoxin and Lee, Yujian and Dang, Zichen and Wang, Lili and Cui, Yawen and Chau, Lap-Pui and Wang, Yi},
    title     = {HVG-3D: Bridging Real and Simulation Domains for 3D-Conditional Hand-Object Interaction Video Synthesis},
    booktitle = {Proceedings of the IEEE/CVF Conference on Computer Vision and Pattern Recognition (CVPR)},
    month     = {June},
    year      = {2026},
    pages     = {15986-15997}
}

@InProceedings{Sun_2026_CVPR_Animator,
    author    = {Sun, Mingze and Zeng, Cheng and Pei, Jiansong and Chen, Junhao and Song, Chaoyue and Wang, Shaohui and Chang, Tianyuan and Huang, Bin and Zeng, Zijiao and Huang, Ruqi},
    title     = {Animator-Centric Skeleton Generation on Objects with Fine-Grained Details},
    booktitle = {Proceedings of the IEEE/CVF Conference on Computer Vision and Pattern Recognition (CVPR)},
    month     = {June},
    year      = {2026},
    pages     = {17336-17345}
}

@article{liu2025omnipsd,
  title={OmniPSD: Layered PSD Generation with Diffusion Transformer},
  author={Liu, Cheng and Song, Yiren and Wang, Haofan and Shou, Mike Zheng},
  journal={arXiv preprint arXiv:2512.09247},
  year={2025}
}

@inproceedings{zhu2024champ,
  title={Champ: Controllable and consistent human image animation with 3d parametric guidance},
  author={Zhu, Shenhao and Chen, Junming Leo and Dai, Zuozhuo and Dong, Zilong and Xu, Yinghui and Cao, Xun and Yao, Yao and Zhu, Hao and Zhu, Siyu},
  booktitle={European Conference on Computer Vision},
  pages={145--162},
  year={2024},
  organization={Springer}
}

@article{lin2026seethrough,
  title={See-through: Single-image Layer Decomposition for Anime Characters},
  author={Lin, Jian and Li, Chengze and Qin, Haoyun and Chan, Kwun Wang and Jin, Yanghua and Liu, Hanyuan and Choy, Stephen Chun Wang and Liu, Xueting},
  journal={arXiv preprint arXiv:2602.03749},
  year={2026}
}

@inproceedings{gao2025linrbridge,
  title={LINR Bridge: Vector Graphic Animation via Neural Implicits and Video Diffusion Priors},
  author={Gao, Wenshuo and Lan, Xicheng and Zhang, Luyao and Yang, Shuai},
  booktitle={IEEE International Conference on Image Processing Workshops (ICIPW)},
  year={2025}
}

@article{aneja2019realtimelipsync,
  title={Real-time lip sync for live 2d animation},
  author={Aneja, Deepali and Li, Wilmot},
  journal={arXiv preprint arXiv:1910.08685},
  year={2019}
}

@InProceedings{Chen_2026_CVPR_LottieGPT,
    author    = {Chen, Junhao and Gao, Kejun and Cui, Yuehan and Sun, Mingze and Chen, Mingjin and Wang, Shaohui and Long, Xiaoxiao and Ma, Fei and Tian, Qi and Zhao, Hao and Huang, Ruqi},
    title     = {LottieGPT: Tokenizing Vector Animation for Autoregressive Generation},
    booktitle = {Proceedings of the IEEE/CVF Conference on Computer Vision and Pattern Recognition (CVPR)},
    month     = {June},
    year      = {2026},
    pages     = {31639-31651}
}

@inproceedings{kirillov2023sam,
  title={Segment anything},
  author={Kirillov, Alexander and Mintun, Eric and Ravi, Nikhila and Mao, Hanzi and Rolland, Chloe and Gustafson, Laura and Xiao, Tete and Whitehead, Spencer and Berg, Alexander C and Lo, Wan-Yen and others},
  booktitle={Proceedings of the IEEE/CVF international conference on computer vision},
  pages={4015--4026},
  year={2023}
}

@article{yu2025pointtrackedit,
  title={Generative Video Motion Editing with 3D Point Tracks},
  author={Lee, Yao-Chih and Zhang, Zhoutong and Huang, Jiahui and Wang, Jui-Hsien and Lee, Joon-Young and Huang, Jia-Bin and Shechtman, Eli and Li, Zhengqi},
  journal={arXiv preprint arXiv:2512.02015},
  year={2025}
}

@misc{wang2025wan22,
  author = {aigc-apps},
  title = {VideoX-Fun: A Video Generation Pipeline for Diffusion Transformer},
  year = {2026},
  publisher = {GitHub},
  url = {https://github.com/aigc-apps/VideoX-Fun}
}

@article{deng2025spiritus,
  title={Spiritus: An AI-assisted tool for creating 2D characters and animations},
  author={Sun, Qirui and Ni, Yunyi and Yuan, Teli and Zhang, Jingjing and Yang, Fan and Yao, Zhihao and Mi, Haipeng},
  journal={arXiv preprint arXiv:2503.09127},
  year={2025}
}

@misc{levy2026livesvgzeroshotsvganimation,
      title={LiveSVG: Zero-Shot SVG Animation via Video Generation}, 
      author={Matan Levy and Ran Margolin and Bar Cavia and Dvir Samuel and Yael Pritch and Shmuel Peleg and Alex Rav Acha and Ariel Shamir and Dani Lischinski},
      year={2026},
      eprint={2605.30174},
      archivePrefix={arXiv},
      primaryClass={cs.CV},
      url={https://arxiv.org/abs/2605.30174}, 
}

@inproceedings{hu2024animateanyone,
  title={Animate anyone: Consistent and controllable image-to-video synthesis for character animation},
  author={Hu, Li},
  booktitle={Proceedings of the IEEE/CVF conference on computer vision and pattern recognition},
  pages={8153--8163},
  year={2024}
}

@misc{bai2025qwen25vl,
      title={Qwen2.5-VL Technical Report}, 
      author={Shuai Bai and Keqin Chen and Xuejing Liu and Jialin Wang and Wenbin Ge and Sibo Song and Kai Dang and Peng Wang and Shijie Wang and Jun Tang and Humen Zhong and Yuanzhi Zhu and Mingkun Yang and Zhaohai Li and Jianqiang Wan and Pengfei Wang and Wei Ding and Zheren Fu and Yiheng Xu and Jiabo Ye and Xi Zhang and Tianbao Xie and Zesen Cheng and Hang Zhang and Zhibo Yang and Haiyang Xu and Junyang Lin},
      year={2025},
      eprint={2502.13923},
      archivePrefix={arXiv},
      primaryClass={cs.CV},
      url={https://arxiv.org/abs/2502.13923}, 
}

@article{wang2024animatex,
  title={Animate-x: Universal character image animation with enhanced motion representation},
  author={Tan, Shuai and Gong, Biao and Wang, Xiang and Zhang, Shiwei and Zheng, Dandan and Zheng, Ruobing and Zheng, Kecheng and Chen, Jingdong and Yang, Ming},
  journal={arXiv preprint arXiv:2410.10306},
  year={2024}
}

@misc{ma2024rigformer,
      title={RigidFormer: Learning Rigid Dynamics using Transformers}, 
      author={Zhiyang Dou and Minghao Guo and Haixu Wu and Doug Roble and Tuur Stuyck and Wojciech Matusik},
      year={2026},
      eprint={2605.09196},
      archivePrefix={arXiv},
      primaryClass={cs.CV},
      url={https://arxiv.org/abs/2605.09196}, 
}

@inproceedings{gao2025layerdecomp,
  title={Generative image layer decomposition with visual effects},
  author={Yang, Jinrui and Liu, Qing and Li, Yijun and Kim, Soo Ye and Pakhomov, Daniil and Ren, Mengwei and Zhang, Jianming and Lin, Zhe and Xie, Cihang and Zhou, Yuyin},
  booktitle={Proceedings of the IEEE/CVF Conference on Computer Vision and Pattern Recognition},
  pages={7643--7653},
  year={2025}
}

@inproceedings{chen2024text2ac,
  title={Text2AC: a framework for game-ready 2D agent character (AC) generation from natural language},
  author={Sun, Qirui and Luo, Qiaoyang and Ni, Yunyi and Mi, Haipeng},
  booktitle={Extended Abstracts of the CHI Conference on Human Factors in Computing Systems},
  pages={1--7},
  year={2024}
}

@article{he2026liwi,
  title={LiWi: Layering in the Wild},
  author={He, Yu and Li, Fang and Tong, Haoyang and Ma, Lichen and Shan, Xinyuan and Fu, Jingling and Chen, Dong and Liu, Luohang and Huang, Junshi and Li, Yan},
  journal={arXiv preprint arXiv:2605.14552},
  year={2026}
}

@inproceedings{shewchuk1996triangle,
  title={Triangle: Engineering a 2D quality mesh generator and Delaunay triangulator},
  author={Shewchuk, Jonathan Richard},
  booktitle={Workshop on applied computational geometry},
  pages={203--222},
  year={1996},
  organization={Springer}
}

@article{gal2023livesketch,
  title={Breathing Life Into Sketches Using Text-to-Video Priors},
  author={Gal, Rinon and Vinker, Yael and Alaluf, Yuval and Bermano, Amit H. and Cohen-Or, Daniel and Shamir, Ariel and Chechik, Gal},
  journal={arXiv preprint arXiv:2311.13608},
  year={2023}
}

@article{ho2020denoising,
  title={Denoising diffusion probabilistic models},
  author={Ho, Jonathan and Jain, Ajay and Abbeel, Pieter},
  journal={Advances in neural information processing systems},
  volume={33},
  pages={6840--6851},
  year={2020}
}

@inproceedings{
chen2026dancetogether,
title={DanceTogether: Generating Interactive Multi-Person Video without Identity Drifting},
author={Junhao Chen and Mingjin Chen and Jianjin Xu and Xiang Li and Junting Dong and Mingze Sun and Puhua Jiang and Hongxiang Li and Yuhang Yang and Hao Zhao and Xiao-Xiao Long and Ruqi Huang},
booktitle={The Fourteenth International Conference on Learning Representations},
year={2026},
url={https://openreview.net/forum?id=7VEECFBzmm}
}

@article{suzuki1985topological,
  title={Topological structural analysis of digitized binary images by border following},
  author={Suzuki, Satoshi and others},
  journal={Computer vision, graphics, and image processing},
  volume={30},
  number={1},
  pages={32--46},
  year={1985},
  publisher={Elsevier}
}

@InProceedings{sun2025drive,
    author    = {Sun, Mingze and Chen, Junhao and Dong, Junting and Chen, Yurun and Jiang, Xinyu and Mao, Shiwei and Jiang, Puhua and Wang, Jingbo and Dai, Bo and Huang, Ruqi},
    title     = {DRiVE: Diffusion-based Rigging Empowers Generation of Versatile and Expressive Characters},
    booktitle = {Proceedings of the IEEE/CVF Conference on Computer Vision and Pattern Recognition (CVPR)},
    month     = {June},
    year      = {2025},
    pages     = {21170-21180}
}

@inproceedings{chen2025idea23d,
  title={Idea23d: Collaborative lmm agents enable 3d model generation from interleaved multimodal inputs},
  author={Chen, Junhao and Li, Xiang and Ye, Xiaojun and Li, Chao and Fan, Zhaoxin and Zhao, Hao},
  booktitle={Proceedings of the 31st International Conference on Computational Linguistics},
  pages={4149--4166},
  year={2025}
}

@inproceedings{siddiqui2024meshgpt,
  title={Meshgpt: Generating triangle meshes with decoder-only transformers},
  author={Siddiqui, Yawar and Alliegro, Antonio and Artemov, Alexey and Tommasi, Tatiana and Sirigatti, Daniele and Rosov, Vladislav and Dai, Angela and Nie{\ss}ner, Matthias},
  booktitle={Proceedings of the IEEE/CVF conference on computer vision and pattern recognition},
  pages={19615--19625},
  year={2024}
}

@inproceedings{wang2025layertracer,
  title={Layertracer: Cognitive-aligned layered svg synthesis via diffusion transformer},
  author={Song, Yiren and Chen, Danze and Shou, Mike Zheng},
  booktitle={Proceedings of the IEEE/CVF International Conference on Computer Vision},
  pages={19731--19741},
  year={2025}
}

@inproceedings{ke2024marigold,
  title={Repurposing diffusion-based image generators for monocular depth estimation},
  author={Ke, Bingxin and Obukhov, Anton and Huang, Shengyu and Metzger, Nando and Daudt, Rodrigo Caye and Schindler, Konrad},
  booktitle={Proceedings of the IEEE/CVF conference on computer vision and pattern recognition},
  pages={9492--9502},
  year={2024}
}

@article{yang2026omnilottie,
  title={OmniLottie: Generating Vector Animations via Parameterized Lottie Tokens},
  author={Yang, Yiying and Cheng, Wei and Chen, Sijin and Fu, Honghao and Zeng, Xianfang and Cai, Yujun and Yu, Gang and Ma, Xingjun},
  journal={arXiv preprint arXiv:2603.02138},
  year={2026}
}

@article{oquab2024dinov2,
  title={Dinov2: Learning robust visual features without supervision},
  author={Oquab, Maxime and Darcet, Timoth{\'e}e and Moutakanni, Th{\'e}o and Vo, Huy and Szafraniec, Marc and Khalidov, Vasil and Fernandez, Pierre and Haziza, Daniel and Massa, Francisco and El-Nouby, Alaaeldin and others},
  journal={arXiv preprint arXiv:2304.07193},
  year={2023}
}

@inproceedings{reddy2021im2vec,
  title={Im2vec: Synthesizing vector graphics without vector supervision},
  author={Reddy, Pradyumna and Gharbi, Michael and Lukac, Michal and Mitra, Niloy J},
  booktitle={Proceedings of the IEEE/CVF conference on computer vision and pattern recognition},
  pages={7342--7351},
  year={2021}
}

@article{chen2026ultraman,
  title={Ultraman: ultra-fast and high-resolution texture generation for 3D human reconstruction from a single image},
  author={Chen, Mingjin and Chen, Junhao and Gao, Huan-ang and Chen, Xiaoxue and Fan, Zhaoxin and Zhao, Hao},
  journal={Machine Vision and Applications},
  volume={37},
  number={2},
  pages={24},
  year={2026},
  publisher={Springer}
}

@inproceedings{li2025layerd,
  title={Layerd: Decomposing raster graphic designs into layers},
  author={Suzuki, Tomoyuki and Liu, Kang-Jun and Inoue, Naoto and Yamaguchi, Kota},
  booktitle={Proceedings of the IEEE/CVF International Conference on Computer Vision},
  pages={17783--17792},
  year={2025}
}

@misc{chen2026enginenativeeditable3dworld,
      title={Engine-Native Editable 3D World Reconstruction with Objects and Lighting}, 
      author={Junhao Chen and Xinghao Chen and Henghaofan Zhang and Zihao Qiao and Saining Zhang and Yongzhi Li and Ruqi Huang and Sisi Li and Yimin Sheng and Jianyi Zhu and Hao Zhao},
      year={2026},
      eprint={2607.20889},
      archivePrefix={arXiv},
      primaryClass={cs.CV},
      url={https://arxiv.org/abs/2607.20889}, 
}

@inproceedings{sederberg1986freeformdeformation,
  title={Free-form deformation of solid geometric models},
  author={Sederberg, Thomas W and Parry, Scott R},
  booktitle={Proceedings of the 13th annual conference on Computer graphics and interactive techniques},
  pages={151--160},
  year={1986}
}

@article{huang2025cartoonalive,
  title={CartoonAlive: Towards Expressive Live2D Modeling from Single Portraits},
  author={He, Chao and Ren, Jianqiang and Xiang, Jianjing and Shen, Xiejie},
  journal={arXiv preprint arXiv:2507.17327},
  year={2025}
}

@inproceedings{chen-etal-2026-paircoder,
    title = "{P}air{C}oder: Pair Programming-Inspired Two-Agent Collaboration for Code Generation",
    author = "Chen, Junhao  and
      Li, Xiang  and
      Xu, Yibin  and
      Cui, Yuehan  and
      Weng, Fangsheng  and
      Zhao, Hao  and
      Ma, Fei  and
      Tian, Qi",
    editor = "Liakata, Maria  and
      Moreira, Viviane P.  and
      Zhang, Jiajun  and
      Jurgens, David",
    booktitle = "Findings of the {A}ssociation for {C}omputational {L}inguistics: {ACL} 2026",
    month = jul,
    year = "2026",
    address = "San Diego, California, United States",
    publisher = "Association for Computational Linguistics",
    url = "https://aclanthology.org/2026.findings-acl.149/",
    doi = "10.18653/v1/2026.findings-acl.149",
    pages = "3043--3058",
    ISBN = "979-8-89176-395-1"
}

@misc{viga,
      title={Vision-as-Inverse-Graphics Agent via Interleaved Multimodal Reasoning},
      author={Shaofeng Yin and Jiaxin Ge and Zora Zhiruo Wang and Xiuyu Li and Michael J. Black and Trevor Darrell and Angjoo Kanazawa and Haiwen Feng},
      year={2026},
      eprint={2601.11109},
      archivePrefix={arXiv},
      primaryClass={cs.CV},
      url={https://arxiv.org/abs/2601.11109},
}

@misc{xu2024CADMLLM,
      title={CAD-MLLM: Unifying Multimodality-Conditioned CAD Generation With MLLM}, 
      author={Jingwei Xu and Chenyu Wang and Zibo Zhao and Wen Liu and Yi Ma and Shenghua Gao},
      year={2024},
      eprint={2411.04954},
      archivePrefix={arXiv},
      primaryClass={cs.CV}
}

@inproceedings{chen2024meshanything,
  title={Meshanything: Artist-created mesh generation with autoregressive transformers},
  author={Chen, Yiwen and He, Tong and Huang, Di and Ye, Weicai and Chen, Sijin and Tang, Jiaxiang and Cai, Zhongang and Yang, Lei and Yu, Gang and Lin, Guosheng and others},
  booktitle={International Conference on Learning Representations},
  volume={2025},
  pages={51369--51389},
  year={2025}
}

@article{wu2025duetsvg,
  title={DuetSVG: Unified Multimodal SVG Generation with Internal Visual Guidance},
  author={Zhang, Peiying and Zhao, Nanxuan and Fisher, Matthew and Xu, Yiran and Liao, Jing and Liu, Difan},
  journal={arXiv preprint arXiv:2512.10894},
  year={2025}
}

@article{huang2025psdiffusion,
  title={PSDiffusion: Harmonized Multi-Layer Image Generation via Layout and Appearance Alignment},
  author={Huang, Dingbang and Li, Wenbo and Zhao, Yifei and Pan, Xinyu and Wang, Chun and Zeng, Yanhong and Dai, Bo},
  journal={arXiv preprint arXiv:2505.11468},
  year={2025}
}

@inproceedings{pu2025art,
  title={Art: Anonymous region transformer for variable multi-layer transparent image generation},
  author={Pu, Yifan and Zhao, Yiming and Tang, Zhicong and Yin, Ruihong and Ye, Haoxing and Yuan, Yuhui and Chen, Dong and Bao, Jianmin and Zhang, Sirui and Wang, Yanbin and others},
  booktitle={Proceedings of the Computer Vision and Pattern Recognition Conference},
  pages={7952--7962},
  year={2025}
}

@inproceedings{meng2025anidoc,
  title={Anidoc: Animation creation made easier},
  author={Meng, Yihao and Ouyang, Hao and Wang, Hanlin and Wang, Qiuyu and Wang, Wen and Cheng, Ka Leong and Liu, Zhiheng and Shen, Yujun and Qu, Huamin},
  booktitle={Proceedings of the IEEE/CVF Conference on Computer Vision and Pattern Recognition},
  pages={18187--18197},
  year={2025}
}

@article{liu2023text2layer,
  title={Text2layer: Layered image generation using latent diffusion model},
  author={Zhang, Xinyang and Zhao, Wentian and Lu, Xin and Chien, Jeff},
  journal={arXiv preprint arXiv:2307.09781},
  year={2023}
}

@misc{weng2026feedforward3deditinglearns,
      title={Feedforward 3D Editing Learns from Semantic-Part Transformation}, 
      author={Jiawei Weng and Saining Zhang and Zhenxin Diao and Peishuo Li and Henghaofan Zhang and Junhao Chen and Hao Zhao},
      year={2026},
      eprint={2605.27351},
      archivePrefix={arXiv},
      primaryClass={cs.CV},
      url={https://arxiv.org/abs/2605.27351}, 
}

@article{morimoto2019rigiddeformation,
  title={Generating 2.5 D character animation by switching the textures of rigid deformation},
  author={Morimoto, Yuki and Makita, Atsuko and Semba, Takuya and Takahashi, Tokiichiro},
  journal={International Journal of Asia Digital Art and Design},
  volume={23},
  number={2},
  pages={16--21},
  year={2019},
  publisher={Asia Digital Art and Design Association}
}

@inproceedings{xing2024dynamicrafter_eccv,
  title={Dynamicrafter: Animating open-domain images with video diffusion priors},
  author={Xing, Jinbo and Xia, Menghan and Zhang, Yong and Chen, Haoxin and Yu, Wangbo and Liu, Hanyuan and Liu, Gongye and Wang, Xintao and Shan, Ying and Wong, Tien-Tsin},
  booktitle={European Conference on Computer Vision},
  pages={399--417},
  year={2024},
  organization={Springer}
}

@inproceedings{
weng2026garmentgpt,
title={Garment{GPT}: Compositional Garment Pattern Generation via Discrete Latent Tokenization},
author={Fangsheng Weng and Junhao Chen and Xiang Li and Jie Qin and Hanzhong Guo and ShaochunHao and Xiaoguang Han},
booktitle={The Fourteenth International Conference on Learning Representations},
year={2026},
url={https://openreview.net/forum?id=XzXKnazRBF}
}

@inproceedings{niu2024mofavideo,
  title={Mofa-video: Controllable image animation via generative motion field adaptions in frozen image-to-video diffusion model},
  author={Niu, Muyao and Cun, Xiaodong and Wang, Xintao and Zhang, Yong and Shan, Ying and Zheng, Yinqiang},
  booktitle={European conference on computer vision},
  pages={111--128},
  year={2024},
  organization={Springer}
}

@inproceedings{guo2025outlinedetail,
  title={Outline and Detail: A Semantic-Driven Framework for Layered 2D Character Generation},
  author={Sun, Qirui and Ni, Yunyi and Qiao, Haixin and Zhang, Jingjing and Yang, Fan and Yuan, Teli and Yao, Zhihao and Mi, Haipeng},
  booktitle={Proceedings of the 38th Annual ACM Symposium on User Interface Software and Technology},
  pages={1--14},
  year={2025}
}

@article{wu2024aniclipart,
  title={AniClipart: Clipart Animation with Text-to-Video Priors},
  author={Wu, Ronghuan and Su, Wanchao and Ma, Kede and Liao, Jing},
  journal={International Journal of Computer Vision},
  year={2024},
  doi={10.1007/s11263-024-02306-1}
}

@article{zhang2024transparent,
  title={Transparent image layer diffusion using latent transparency},
  author={Zhang, Lvmin and Agrawala, Maneesh},
  journal={arXiv preprint arXiv:2402.17113},
  year={2024}
}

@article{rossignac1999edgebreaker,
  title={Edgebreaker: Connectivity compression for triangle meshes},
  author={Rossignac, Jarek},
  journal={IEEE transactions on visualization and computer graphics},
  volume={5},
  number={1},
  pages={47--61},
  year={1999},
  publisher={IEEE}
}

@article{wang2024llamamesh,
  title={Llama-mesh: Unifying 3d mesh generation with language models},
  author={Wang, Zhengyi and Lorraine, Jonathan and Wang, Yikai and Su, Hang and Zhu, Jun and Fidler, Sanja and Zeng, Xiaohui},
  journal={arXiv preprint arXiv:2411.09595},
  year={2024}
}

@article{anand2024rignet,
  title={Rignet: Neural rigging for articulated characters},
  author={Xu, Zhan and Zhou, Yang and Kalogerakis, Evangelos and Landreth, Chris and Singh, Karan},
  journal={arXiv preprint arXiv:2005.00559},
  year={2020}
}

@inproceedings{tang2024edgerunner,
  title={Edgerunner: Auto-regressive auto-encoder for artistic mesh generation},
  author={Tang, Jiaxiang and Li, Max and Hao, Zekun and Liu, Xian and Zeng, Gang and Liu, Ming-Yu and Zhang, Qinsheng},
  booktitle={International Conference on Learning Representations},
  volume={2025},
  pages={35913--35934},
  year={2025}
}

@article{barber1996quickhull,
  title={The quickhull algorithm for convex hulls},
  author={Barber, C Bradford and Dobkin, David P and Huhdanpaa, Hannu},
  journal={ACM Transactions on Mathematical Software (TOMS)},
  volume={22},
  number={4},
  pages={469--483},
  year={1996},
  publisher={Acm New York, NY, USA}
}

@article{xing2024tooncrafter,
  title={Tooncrafter: Generative cartoon interpolation},
  author={Xing, Jinbo and Liu, Hanyuan and Xia, Menghan and Zhang, Yong and Wang, Xintao and Shan, Ying and Wong, Tien-Tsin},
  journal={ACM Transactions on Graphics (TOG)},
  volume={43},
  number={6},
  pages={1--11},
  year={2024},
  publisher={ACM New York, NY, USA}
}

@inproceedings{xu2024magicanimate,
  title={Magicanimate: Temporally consistent human image animation using diffusion model},
  author={Xu, Zhongcong and Zhang, Jianfeng and Liew, Jun Hao and Yan, Hanshu and Liu, Jia-Wei and Zhang, Chenxu and Feng, Jiashi and Shou, Mike Zheng},
  booktitle={Proceedings of the IEEE/CVF Conference on Computer Vision and Pattern Recognition},
  pages={1481--1490},
  year={2024}
}

@article{ravi2024sam3,
  title={Sam 3: Segment anything with concepts},
  author={Carion, Nicolas and Gustafson, Laura and Hu, Yuan-Ting and Debnath, Shoubhik and Hu, Ronghang and Suris, Didac and Ryali, Chaitanya and Alwala, Kalyan Vasudev and Khedr, Haitham and Huang, Andrew and others},
  journal={arXiv preprint arXiv:2511.16719},
  year={2025}
}

@misc{live2d2024cubism,
  title  = {Live2D {Cubism} Editor: Mesh Generation Tool Documentation},
  author = {{Live2D Inc.}},
  year   = {2024},
  note   = {\url{https://docs.live2d.com/cubism-editor-manual/mesh-generation/}}
}

@inproceedings{rombach2022stablediffusion,
  title={High-resolution image synthesis with latent diffusion models},
  author={Rombach, Robin and Blattmann, Andreas and Lorenz, Dominik and Esser, Patrick and Ommer, Bj{\"o}rn},
  booktitle={Proceedings of the IEEE/CVF conference on computer vision and pattern recognition},
  pages={10684--10695},
  year={2022}
}

@inproceedings{yang2025layeranimate,
  title={Layeranimate: Layer-level control for animation},
  author={Yang, Yuxue and Fan, Lue and Lin, Zuzeng and Wang, Feng and Zhang, Zhaoxiang},
  booktitle={Proceedings of the IEEE/CVF International Conference on Computer Vision},
  pages={10865--10874},
  year={2025}
}

@inproceedings{iwbench,
    title = "{IW}-Bench: Evaluating Large Multimodal Models for Converting Image-to-Web",
    author = "Guo, Hongcheng  and
      Zhang, Wei  and
      Chen, Junhao  and
      Gu, Yaonan  and
      Yang, Jian  and
      Du, Junjia  and
      Cao, Shaosheng  and
      Hui, Binyuan  and
      Liu, Tianyu  and
      Ma, Jianxin  and
      Zhou, Chang  and
      Li, Zhoujun",
    editor = "Che, Wanxiang  and
      Nabende, Joyce  and
      Shutova, Ekaterina  and
      Pilehvar, Mohammad Taher",
    booktitle = "Findings of the Association for Computational Linguistics: ACL 2025",
    month = jul,
    year = "2025",
    address = "Vienna, Austria",
    publisher = "Association for Computational Linguistics",
    url = "https://aclanthology.org/2025.findings-acl.334/",
    doi = "10.18653/v1/2025.findings-acl.334",
    pages = "6449--6466",
    ISBN = "979-8-89176-256-5"
}

@article{wan2025wan2,
      title={Wan: Open and Advanced Large-Scale Video Generative Models}, 
      author={Team Wan and Ang Wang and Baole Ai and Bin Wen and Chaojie Mao and Chen-Wei Xie and Di Chen and Feiwu Yu and Haiming Zhao and Jianxiao Yang and Jianyuan Zeng and Jiayu Wang and Jingfeng Zhang and Jingren Zhou and Jinkai Wang and Jixuan Chen and Kai Zhu and Kang Zhao and Keyu Yan and Lianghua Huang and Mengyang Feng and Ningyi Zhang and Pandeng Li and Pingyu Wu and Ruihang Chu and Ruili Feng and Shiwei Zhang and Siyang Sun and Tao Fang and Tianxing Wang and Tianyi Gui and Tingyu Weng and Tong Shen and Wei Lin and Wei Wang and Wei Wang and Wenmeng Zhou and Wente Wang and Wenting Shen and Wenyuan Yu and Xianzhong Shi and Xiaoming Huang and Xin Xu and Yan Kou and Yangyu Lv and Yifei Li and Yijing Liu and Yiming Wang and Yingya Zhang and Yitong Huang and Yong Li and You Wu and Yu Liu and Yulin Pan and Yun Zheng and Yuntao Hong and Yupeng Shi and Yutong Feng and Zeyinzi Jiang and Zhen Han and Zhi-Fan Wu and Ziyu Liu},
      journal = {arXiv preprint arXiv:2503.20314},
      year={2025}
}

@article{yin2024qwenimagelayered,
  title={Qwen-Image-Layered: Towards Inherent Editability via Layer Decomposition},
  author={Yin, Shengming and Zhang, Zekai and Tang, Zecheng and Gao, Kaiyuan and Xu, Xiao and Yan, Kun and Li, Jiahao and Chen, Yilei and Chen, Yuxiang and Shum, Heung-Yeung and others},
  journal={arXiv preprint arXiv:2512.15603},
  year={2025}
}

@article{liu2024textoon,
  title={Textoon: Generating vivid 2d cartoon characters from text descriptions},
  author={He, Chao and Ren, Jianqiang and Dong, Yuan and Xiang, Jianjing and Shen, Xiejie and Yuan, Weihao and Bo, Liefeng},
  journal={arXiv preprint arXiv:2501.10020},
  year={2025}
}

@misc{chen2026videoworldturningmonocular,
      title={One Video, One World: Turning Monocular Video into Physical 4D Scenes}, 
      author={Junhao Chen and Boran Zhang and Mingjin Chen and Henghaofan Zhang and Saining Zhang and Congcong Zhu and Hao Zhao and Ruqi Huang and Zhihao Li and Yufei Wang},
      year={2026},
      eprint={2606.31388},
      archivePrefix={arXiv},
      primaryClass={cs.CV},
      url={https://arxiv.org/abs/2606.31388}, 
}

@inproceedings{xing2025llm4svg,
  title={Empowering llms to understand and generate complex vector graphics},
  author={Xing, Ximing and Hu, Juncheng and Liang, Guotao and Zhang, Jing and Xu, Dong and Yu, Qian},
  booktitle={Proceedings of the Computer Vision and Pattern Recognition Conference},
  pages={19487--19497},
  year={2025}
}

@inproceedings{sorkine2007arap,
  title={As-rigid-as-possible surface modeling},
  author={Sorkine, Olga and Alexa, Marc and others},
  booktitle={Symposium on Geometry processing},
  volume={4},
  pages={109--116},
  year={2007}
}

@article{wang2025qwenimage,
  title={Qwen-image technical report},
  author={Wu, Chenfei and Li, Jiahao and Zhou, Jingren and Lin, Junyang and Gao, Kaiyuan and Yan, Kun and Yin, Sheng-ming and Bai, Shuai and Xu, Xiao and Chen, Yilei and others},
  journal={arXiv preprint arXiv:2508.02324},
  year={2025}
}

@misc{miao2026framessequencestemporallyconsistent,
      title={From Frames to Sequences: Temporally Consistent Human-Centric Dense Prediction}, 
      author={Xingyu Miao and Junting Dong and Qin Zhao and Yuhang Yang and Junhao Chen and Yang Long},
      year={2026},
      eprint={2602.01661},
      archivePrefix={arXiv},
      primaryClass={cs.CV},
      url={https://arxiv.org/abs/2602.01661}, 
}

@article{bridson2007fast,
  title={Fast Poisson disk sampling in arbitrary dimensions.},
  author={Bridson, Robert},
  journal={SIGGRAPH sketches},
  volume={10},
  number={1},
  pages={1},
  year={2007}
}

@article{kim2025layeringdiff,
  title={LayeringDiff: Layered Image Synthesis via Generation, then Disassembly with Generative Knowledge},
  author={Kang, Kyoungkook and Sim, Gyujin and Kim, Geonung and Kim, Donguk and Nam, Seungho and Cho, Sunghyun},
  journal={arXiv preprint arXiv:2501.01197},
  year={2025}
}

@article{guo2024animatediff,
  title={Animatediff: Animate your personalized text-to-image diffusion models without specific tuning},
  author={Guo, Yuwei and Yang, Ceyuan and Rao, Anyi and Liang, Zhengyang and Wang, Yaohui and Qiao, Yu and Agrawala, Maneesh and Lin, Dahua and Dai, Bo},
  journal={arXiv preprint arXiv:2307.04725},
  year={2023}
}

@article{chen2024meshxl,
  title={Meshxl: Neural coordinate field for generative 3d foundation models},
  author={Chen, Sijin and Chen, Xin and Pang, Anqi and Zeng, Xianfang and Cheng, Wei and Fu, Yijun and Yin, Fukun and Wang, Zhibin and Yu, Jingyi and Yu, Gang and others},
  journal={Advances in Neural Information Processing Systems},
  volume={37},
  pages={97141--97166},
  year={2024}
}

@inproceedings{rodriguez2025starvector,
  title={Starvector: Generating scalable vector graphics code from images and text},
  author={Rodriguez, Juan A and Puri, Abhay and Agarwal, Shubham and Laradji, Issam H and Rodriguez, Pau and Rajeswar, Sai and Vazquez, David and Pal, Christopher and Pedersoli, Marco},
  booktitle={Proceedings of the Computer Vision and Pattern Recognition Conference},
  pages={16175--16186},
  year={2025}
}

@inproceedings{lopes2019svgvae,
  title={A learned representation for scalable vector graphics},
  author={Lopes, Raphael Gontijo and Ha, David and Eck, Douglas and Shlens, Jonathon},
  booktitle={Proceedings of the IEEE/CVF international conference on computer vision},
  pages={7930--7939},
  year={2019}
}

@article{carlier2020deepsvg,
  title={Deepsvg: A hierarchical generative network for vector graphics animation},
  author={Carlier, Alexandre and Danelljan, Martin and Alahi, Alexandre and Timofte, Radu},
  journal={Advances in Neural Information Processing Systems},
  volume={33},
  pages={16351--16361},
  year={2020}
}

@inproceedings{wu2025chat2svg,
  title={Chat2svg: Vector graphics generation with large language models and image diffusion models},
  author={Wu, Ronghuan and Su, Wanchao and Liao, Jing},
  booktitle={Proceedings of the Computer Vision and Pattern Recognition Conference},
  pages={23690--23700},
  year={2025}
}

@article{weng2024pivotmesh,
  title={Pivotmesh: Generic 3d mesh generation via pivot vertices guidance},
  author={Weng, Haohan and Wang, Yikai and Zhang, Tong and Chen, CL and Zhu, Jun},
  journal={arXiv preprint arXiv:2405.16890},
  year={2024}
}

@inproceedings{polaczek2025neuralsvg,
  title={Neuralsvg: An implicit representation for text-to-vector generation},
  author={Polaczek, Sagi and Alaluf, Yuval and Richardson, Elad and Vinker, Yael and Cohen-Or, Daniel},
  booktitle={Proceedings of the IEEE/CVF International Conference on Computer Vision},
  pages={15458--15468},
  year={2025}
}

@misc{qwenimageedit2511,
  title = {Qwen-Image-Edit-2511: Instruction-Guided Image Editing},
  author = {{Qwen Team}},
  year = {2025},
  howpublished = {Model release, \url{https://huggingface.co/Qwen/Qwen-Image-Edit-2511}},
  note = {Instruction-guided image editing model built on the Qwen-Image backbone}
}

@inproceedings{yang2024cogvideox,
  title={Cogvideox: Text-to-video diffusion models with an expert transformer},
  author={Yang, Zhuoyi and Teng, Jiayan and Zheng, Wendi and Ding, Ming and Huang, Shiyu and Xu, Jiazheng and Yang, Yuanming and Hong, Wenyi and Zhang, Xiaohan and Feng, Guanyu and others},
  booktitle={International Conference on Learning Representations},
  volume={2025},
  pages={83048--83077},
  year={2025}
}

@article{lloyd1982least,
  title={Least squares quantization in PCM},
  author={Lloyd, Stuart},
  journal={IEEE transactions on information theory},
  volume={28},
  number={2},
  pages={129--137},
  year={1982},
  publisher={IEEE}
}

@article{code2worlds,
  title={Code2Worlds: Empowering Coding LLMs for 4D World Generation},
  author={Zhang, Yi and Wang, Yunshuang and Zhang, Zeyu and Tang, Hao},
  journal={arXiv preprint arXiv:2602.11757},
  year={2026}
}

@inproceedings{yang2024physanimator,
  title={Physanimator: Physics-guided generative cartoon animation},
  author={Xie, Tianyi and Zhao, Yiwei and Jiang, Ying and Jiang, Chenfanfu},
  booktitle={Proceedings of the Computer Vision and Pattern Recognition Conference},
  pages={10793--10804},
  year={2025}
}

\clearpage

\clearpage
\appendix
\section{Stage 1 Architecture}
\label{app:stage1}

Stage~1 initialises from Qwen-Image-Layered ~\cite{yin2024qwenimagelayered} and conditions on (i)~the VAE-latent image anchor of the input illustration, (ii)~Live2D-taxonomy layer-class tokens, (iii)~optional per-layer captions ($50\%$ dropout at training time), and (iv)~per-layer occlusion masks that drive the hidden-region $L_1$ loss; the four conditioning streams are fused at the layered-diffusion backbone's cross-attention layers, and the final supervised fine-tune runs for $30$ epochs on the Live2D-$10$K split (\S\ref{sec:setup}).

\section{Dataset}
\label{app:dataset}

\paragraph*{Corpus statistics.}
After deduplication, our corpus contains $8{,}884$ usable Live2D models ($7{,}773$ human/humanoid and $1{,}111$ non-human, tagged via Qwen3.6-Plus). These binary labels solely guide balanced benchmark sampling; training is supervised entirely by the extracted Live2D structural data.

\begin{figure}[htbp]
    \centering
    \includegraphics[width=0.9\linewidth]{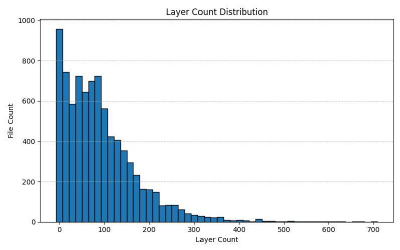}
    \vspace{1.5em} 
    \includegraphics[width=0.9\linewidth]{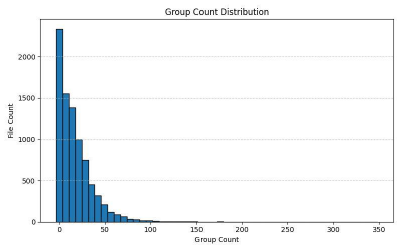}
    \vspace{-3em}
    \caption{\textbf{Dataset statistics.} The distribution of layer counts (top) and group counts (bottom) across the dataset.}
    \label{fig:dataset_statistics}
\end{figure}

Figure~\ref{fig:dataset_statistics} shows the raw layer and group count distributions. 

\begin{figure}[t]
  \centering
  \includegraphics[width=\linewidth]{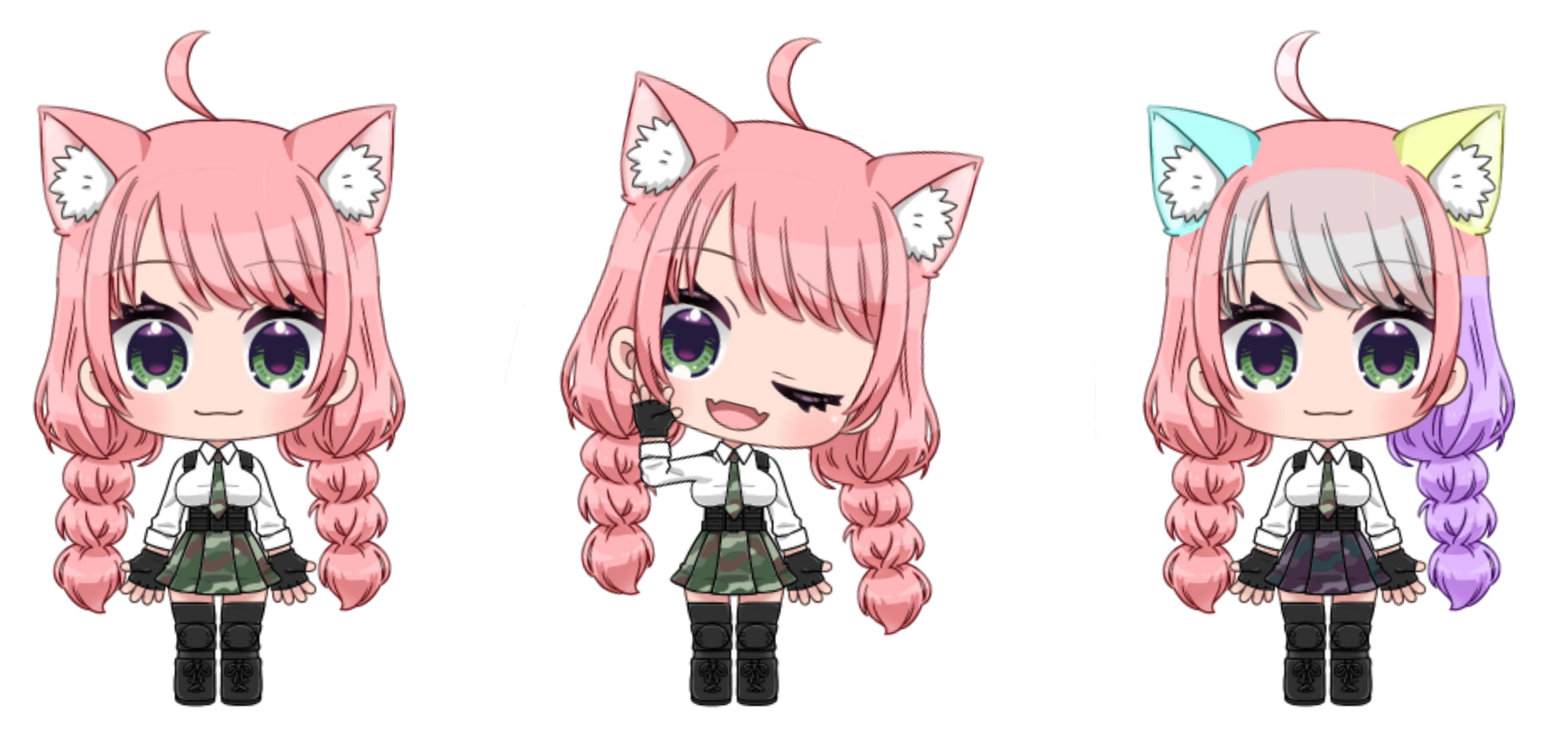}
  \caption{\textbf{Illustration of our data augmentation method.} (a) The original input data. (b) Pose and expression variations driven by underlying Live2D parameters. (c) Color and texture replacements applied to individual layers to enhance diversity.}
  \label{fig:data_aug}
\end{figure}

\paragraph*{Data augmentation.}
We apply two structure-preserving augmentations (Figure~\ref{fig:data_aug}): \emph{motion replacement}, which applies alternative parameter curves and keyposes to a character's canonical layer stack to increase animation diversity; and \emph{colour/texture replacement}, which alters RGBA appearance while retaining alpha masks, meshes, and keypose offsets. These strategies expand the Stage~1 (layer decomposition) and Stage~2 (animation) training subsets to approximately $50$K and $35$K examples, respectively.

\section{Benchmark}
\label{app:benchmark}

\paragraph*{Metric protocol.}
Live2D-Bench evaluates a predicted ordered RGBA stack
$P\!=\!\{(R_j^p,\alpha_j^p)\}$ against a ground-truth stack
$G=\{(R_i^g,\alpha_i^g)\}_{i=1}^{m}$, where the number of predicted
layers is allowed to differ from the number of ground-truth layers. RGB
values are normalized to $[0,1]$ unless stated otherwise, alpha values are
continuous in $[0,1]$, and $\epsilon=10^{-8}$ avoids zero denominators.

\paragraph*{Full-image metrics.}
For composite-level scores we first alpha-composite each stack into a
single RGB image and also collapse its layer alphas by max pooling,
\begin{equation}
  \label{eq:bench_alphafull}
\alpha_{\mathrm{full}}^g(x,y)=\max_i \alpha_i^g(x,y),\qquad
\alpha_{\mathrm{full}}^p(x,y)=\max_j \alpha_j^p(x,y).
\end{equation}
The full-image alpha score (Eq.~\ref{eq:bench_ioufull}) is the soft IoU over the continuous alpha maps,
\begin{equation}
  \label{eq:bench_ioufull}
\mathrm{IoU}_{\alpha}^{\mathrm{full}}
=
\frac{\sum_{x,y}\min(\alpha_{\mathrm{full}}^g(x,y),
                    \alpha_{\mathrm{full}}^p(x,y))}
     {\sum_{x,y}\max(\alpha_{\mathrm{full}}^g(x,y),
                    \alpha_{\mathrm{full}}^p(x,y))+\epsilon}.
\end{equation}
The reported full-image RGB-L1 (Eq.~\ref{eq:bench_rgbl1}) uses the ground-truth foreground support,
$M_g(x,y)=\mathbf{1}[\alpha_{\mathrm{full}}^g(x,y)>0]$,
\begin{equation}
  \label{eq:bench_rgbl1}
\mathrm{RGB\mbox{-}L1}^{\mathrm{full}}
=
\frac{\sum_{x,y} M_g(x,y)\,\frac{1}{3}\sum_{c\in\{R,G,B\}}
|I_c^p(x,y)-I_c^g(x,y)|}
{\sum_{x,y} M_g(x,y)+\epsilon}.
\end{equation}
PSNR and SSIM are computed on the entire composited RGB image; PSNR uses
the standard $255$-range MSE, and SSIM uses a $7{\times}7$ uniform window
(\texttt{scikit-image} defaults, i.e.\ no Gaussian weighting) and averages
the three color channels. LPIPS is computed on the full
composited RGB images with the standard pretrained AlexNet LPIPS features.

\paragraph*{Hungarian layer matching.}
For every ground-truth layer $i$ and predicted layer $j$, we compute a
pairwise alpha IoU and foreground RGB error (Eqs.~\ref{eq:bench_aij} and~\ref{eq:bench_rij}),
\begin{equation}
  \label{eq:bench_aij}
a_{ij}=
\frac{\sum_{x,y}\min(\alpha_i^g(x,y),\alpha_j^p(x,y))}
     {\sum_{x,y}\max(\alpha_i^g(x,y),\alpha_j^p(x,y))+\epsilon},
\end{equation}
\begin{equation}
  \label{eq:bench_rij}
r_{ij}=
\frac{\sum_{x,y}\mathbf{1}[\alpha_i^g(x,y)>0]\,
\frac{1}{3}\sum_{c\in\{R,G,B\}}
|R_{j,c}^p(x,y)-R_{i,c}^g(x,y)|}
{\sum_{x,y}\mathbf{1}[\alpha_i^g(x,y)>0]+\epsilon}.
\end{equation}
The assignment cost combines the two (Eq.~\ref{eq:bench_cij}),
\begin{equation}
  \label{eq:bench_cij}
c_{ij}=\frac{r_{ij}+(1-a_{ij})}{2}\in[0,1].
\end{equation}
We then solve the rectangular linear assignment problem of Eq.~\ref{eq:bench_assign},
\begin{equation}
  \label{eq:bench_assign}
M=\operatorname*{arg\,min}_{\substack{M'\subseteq\{1,\ldots,m\}\times
\{1,\ldots,n\}\\ |M'|=k,\ \mathrm{one\mbox{-}to\mbox{-}one}}}
\sum_{(i,j)\in M'} c_{ij},\qquad k=\min(m,n).
\end{equation}
The implementation uses the standard Hungarian / LAPJV solver and thus
directly supports rectangular cost matrices. If the two stacks have
different sizes, $u=\max(m,n)-k$ layers are unmatched. For matched-only
metrics we average over the $k$ assigned pairs. For penalized metrics,
unmatched layers receive the phantom values $a=0$, $r=1$, and $c=1$, and the denominator is $\max(m,n)$:
\begin{align}
  \label{eq:bench_penagg}
\bar a_{\mathrm{pen}} &= \frac{\sum_{(i,j)\in M}a_{ij}}{\max(m,n)}, \\
\bar r_{\mathrm{pen}} &= \frac{\sum_{(i,j)\in M}r_{ij}+u}{\max(m,n)}, \\
\bar c_{\mathrm{pen}} &= \frac{\sum_{(i,j)\in M}c_{ij}+u}{\max(m,n)}.
\end{align}
Matched-layer LPIPS uses the same matching $M$: each matched RGBA layer is
alpha-composited over a white background, LPIPS is computed for the pair,
and the result is averaged over matched pairs.

\paragraph*{Layer-order consistency.}
To score whether depth order is preserved, we sort the matched pairs by
ground-truth layer index and read the corresponding predicted indices
$p_1,\ldots,p_k$. The number of inversions
\begin{equation}
  \label{eq:bench_inv}
I=\#\{(s,t): s<t,\ p_s>p_t\}
\end{equation}
is exactly the number of crossing edges in the bipartite matching diagram.
The matched order score is a Kendall-style normalized agreement,
\begin{equation}
  \label{eq:bench_omatch}
O_{\mathrm{match}}=1-\frac{I}{\binom{k}{2}},\qquad k\ge 2,
\end{equation}
so a perfectly ordered stack scores $1$ and a completely reversed stack
scores $0$. For $k<2$ there is no ordering decision to make, so the score is
left undefined and the case is excluded from the aggregate rather than being
credited with a perfect score. The penalized order score also accounts for missing or extra
layers,
\begin{equation}
  \label{eq:bench_open}
O_{\mathrm{pen}}=O_{\mathrm{match}}\frac{\min(m,n)}{\max(m,n)},\qquad
\mathrm{unmatched\mbox{-}rate}=\frac{u}{\max(m,n)}.
\end{equation}
When aggregating the fair order number across benchmark cases, we use
pair-weighted averaging with weight $\binom{k}{2}$ so that dense layer
stacks contribute in proportion to the number of ordering decisions.

\paragraph*{Coverage and mask metrics.}
Cov-MAE measures whether the predicted stack has the same per-pixel layer
coverage count as the ground truth. With threshold $\tau=0.5$,
\begin{equation}
  \label{eq:bench_cov}
\mathrm{cov}(x,y)=\left|\{k:\alpha^{(k)}(x,y)>\tau\}\right|,
\end{equation}
\begin{equation}
  \label{eq:bench_covmae}
\mathrm{Cov\mbox{-}MAE}
=\frac{1}{HW}\sum_{x,y}
\left|\mathrm{cov}_g(x,y)-\mathrm{cov}_p(x,y)\right|.
\end{equation}
We additionally track the signed decomposition
\begin{align}
  \label{eq:bench_overunder}
\mathrm{over} &= \frac{1}{HW}\sum_{x,y}\max(\mathrm{cov}_p-\mathrm{cov}_g,0), \\
\mathrm{under} &= \frac{1}{HW}\sum_{x,y}\max(\mathrm{cov}_g-\mathrm{cov}_p,0),
\end{align}
whose sum equals Cov-MAE. Over counts excess predicted layer coverage,
whereas under counts missing predicted coverage.

Mask Dice is computed on the same Hungarian pairs for compatibility with
layer-decomposition baselines. For pair $(i,j)$ we use the squared soft
Dice form with smooth term $s=1$,
\begin{gather}
  \label{eq:bench_dice}
\mathrm{Dice}_{ij} = \frac{2\sum_{x,y}\alpha_i^g(x,y)\alpha_j^p(x,y)+s}{\sum_{x,y}(\alpha_i^g(x,y))^2 + \sum_{x,y}(\alpha_j^p(x,y))^2+s}, \\
\mathrm{MaskDiceLoss}_{ij} = 1 - \mathrm{Dice}_{ij}.
\end{gather}
The matched Mask Dice loss averages this loss over $M$; the penalized
version gives each unmatched layer loss $1$ and divides by $\max(m,n)$.

\section{Stage 1 Qualitative Results}
\label{app:stage1_qual}

\begin{figure*}[!tp]
  \centering
  \includegraphics[width=\linewidth]{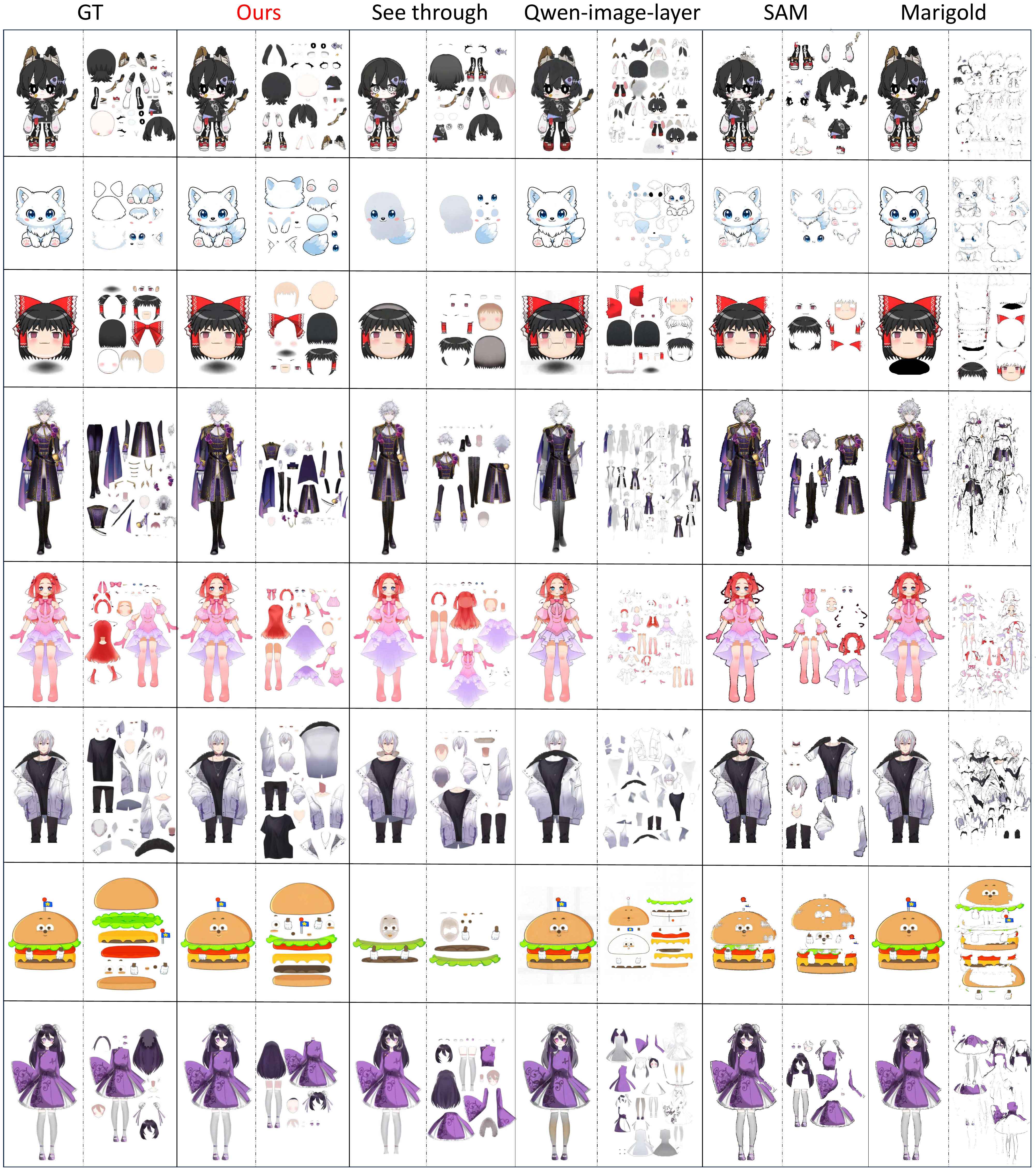}
  \caption{\textbf{Stage~1 layer decomposition against the segmentation and layering baselines}, sheet~1. Eight illustrations, one per row. Each method is given the same flattened illustration; within a method's column the left half is the re-composited result and the right half is its layer stack exploded. Left to right: ground-truth artist layers, ours, See-through, Qwen-Image-Layered, SAM, Marigold-depth. Note that Marigold, having only depth to work with, splits by depth band rather than by part, and SAM segments visible pixels without completing what a layer hides.}
  \label{fig:stage1res1}
\end{figure*}

\begin{figure*}[!tp]
  \centering
  \includegraphics[width=\linewidth]{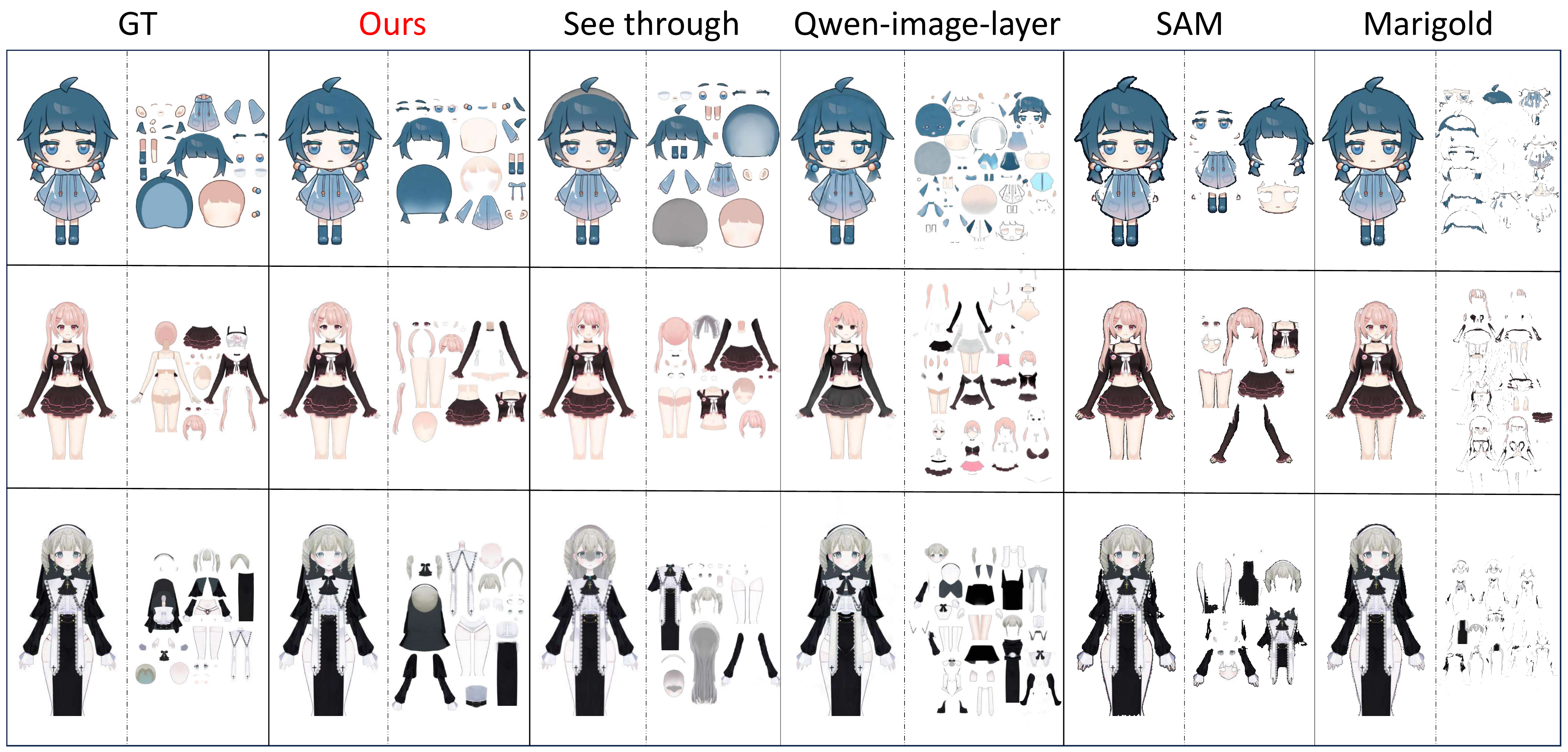}
  \caption{\textbf{Stage~1 layer decomposition against the baselines}, sheet~2. Eight further illustrations; format exactly as Fig.~\ref{fig:stage1res1}.}
  \label{fig:stage1res2}
\end{figure*}

\begin{figure*}[!tp]
  \centering
  \includegraphics[width=\linewidth]{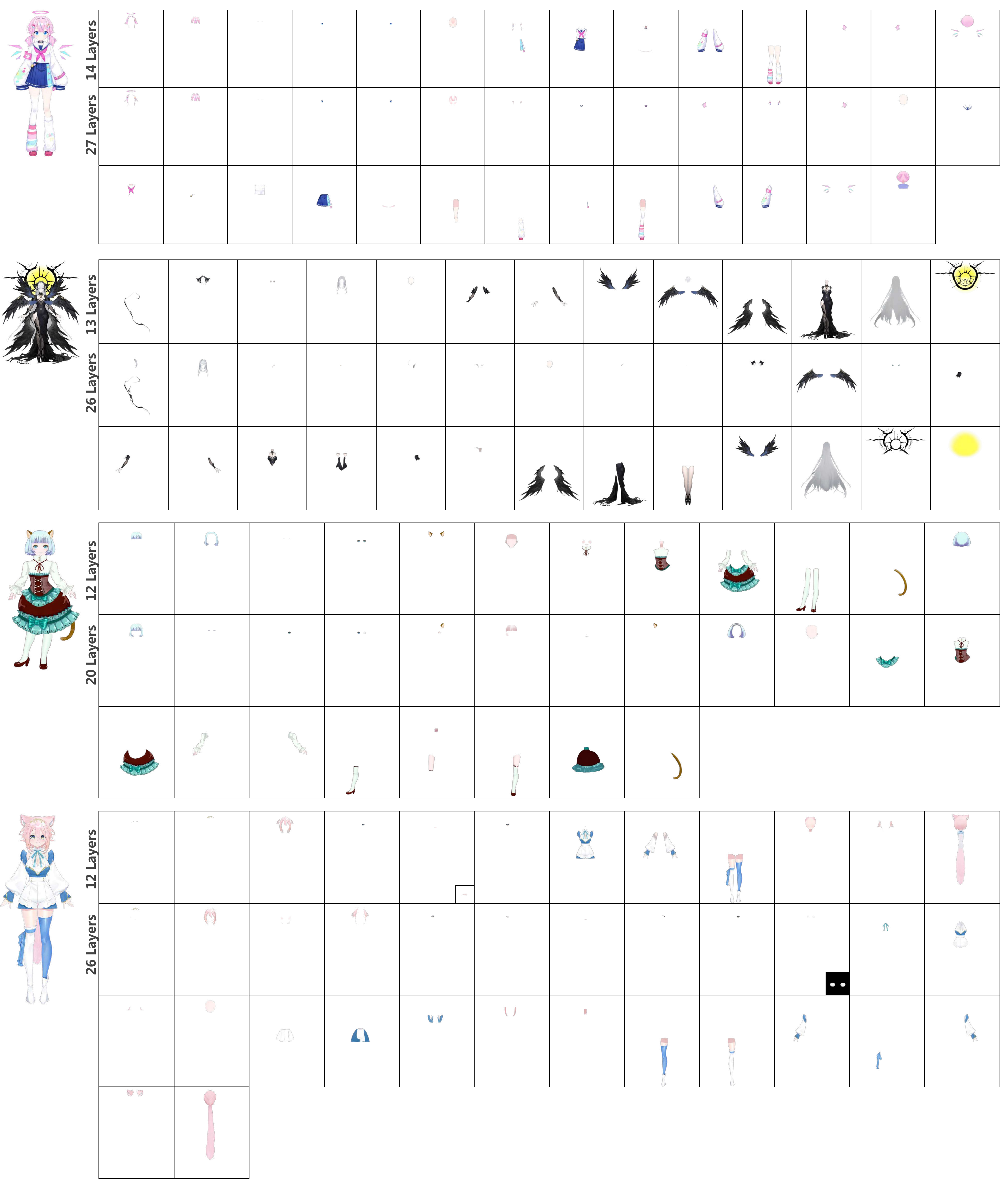}
  \caption{The reconstructed characters (left) are shown alongside their decomposed layer textures (right) under different specified layer counts. Our approach naturally adapts to free layer numbers, supporting both coarse and fine-grained separation.}
  \label{fig:stage1res3}
\end{figure*}

\begin{figure*}[!tp]
  \centering
  \includegraphics[width=\linewidth]{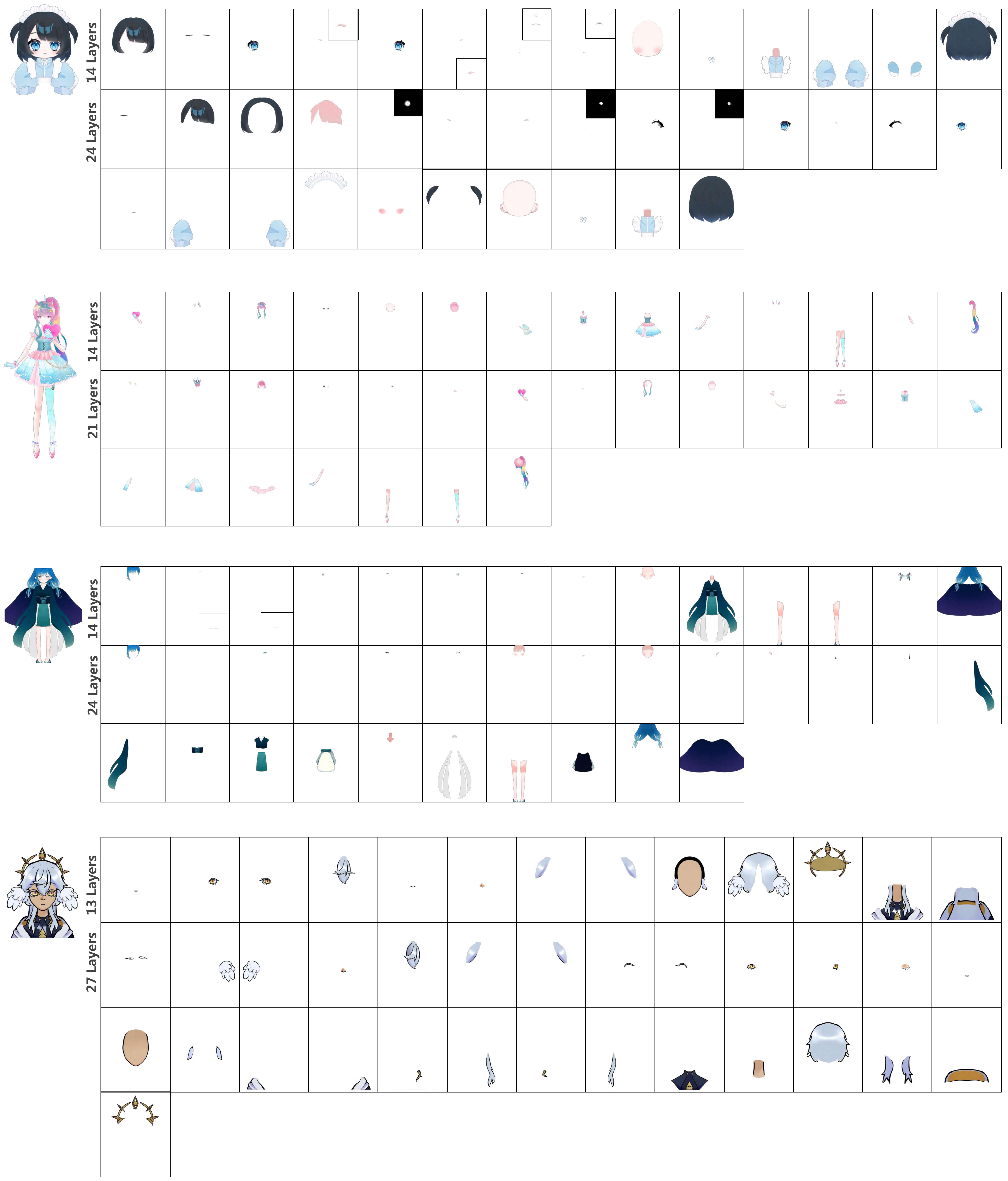}
  \caption{The reconstructed characters (left) are shown alongside their decomposed layer textures (right) under different specified layer counts. Our approach naturally adapts to free layer numbers, supporting both coarse and fine-grained separation.}
  \label{fig:stage1res4}
\end{figure*}

\begin{figure*}[!tp]
  \centering
  \includegraphics[width=\linewidth]{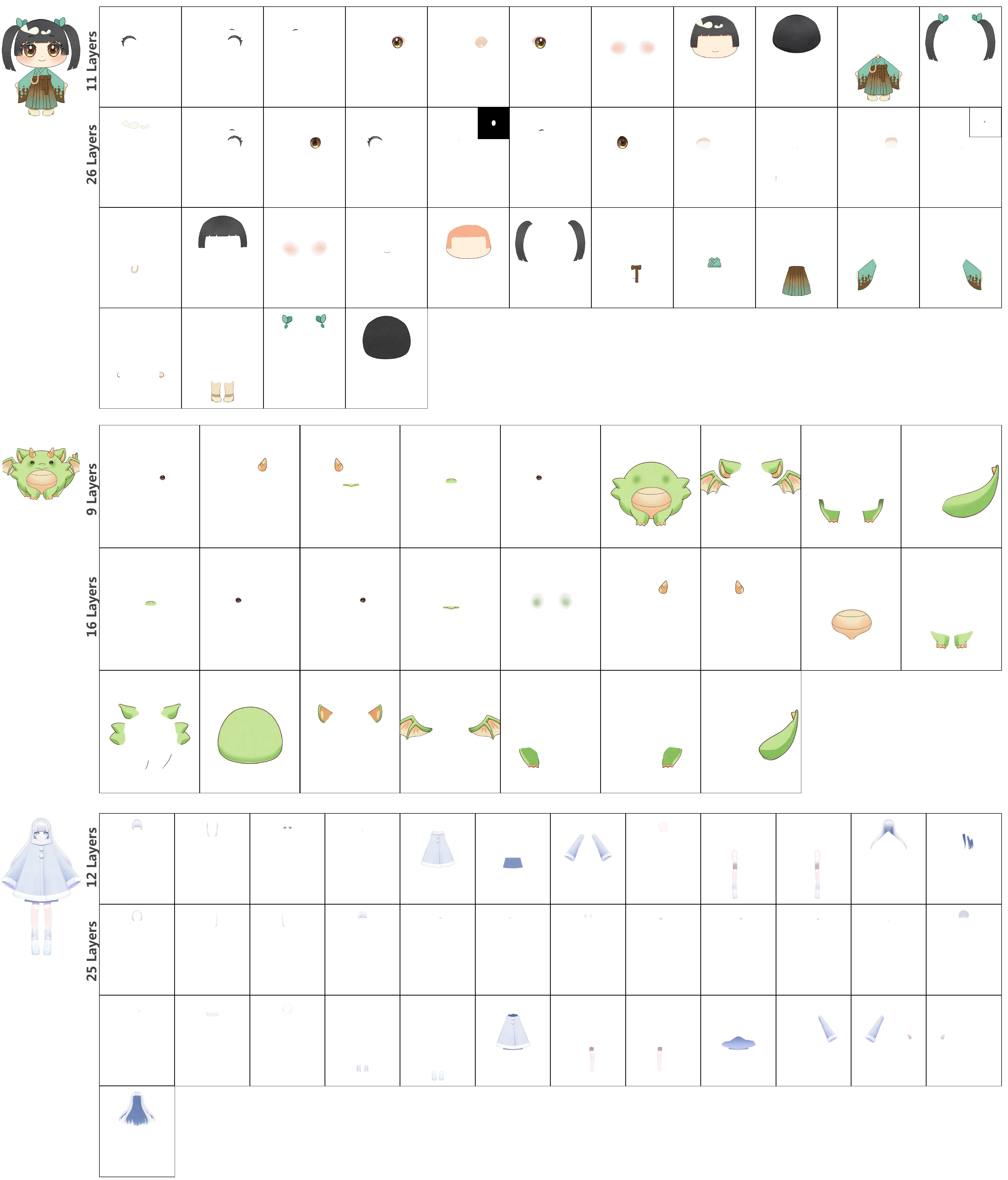}
  \caption{The reconstructed characters (left) are shown alongside their decomposed layer textures (right) under different specified layer counts. Our approach naturally adapts to free layer numbers, supporting both coarse and fine-grained separation.}
  \label{fig:stage1res5}
\end{figure*}

\begin{figure*}[!tp]
  \centering
  \includegraphics[width=\linewidth]{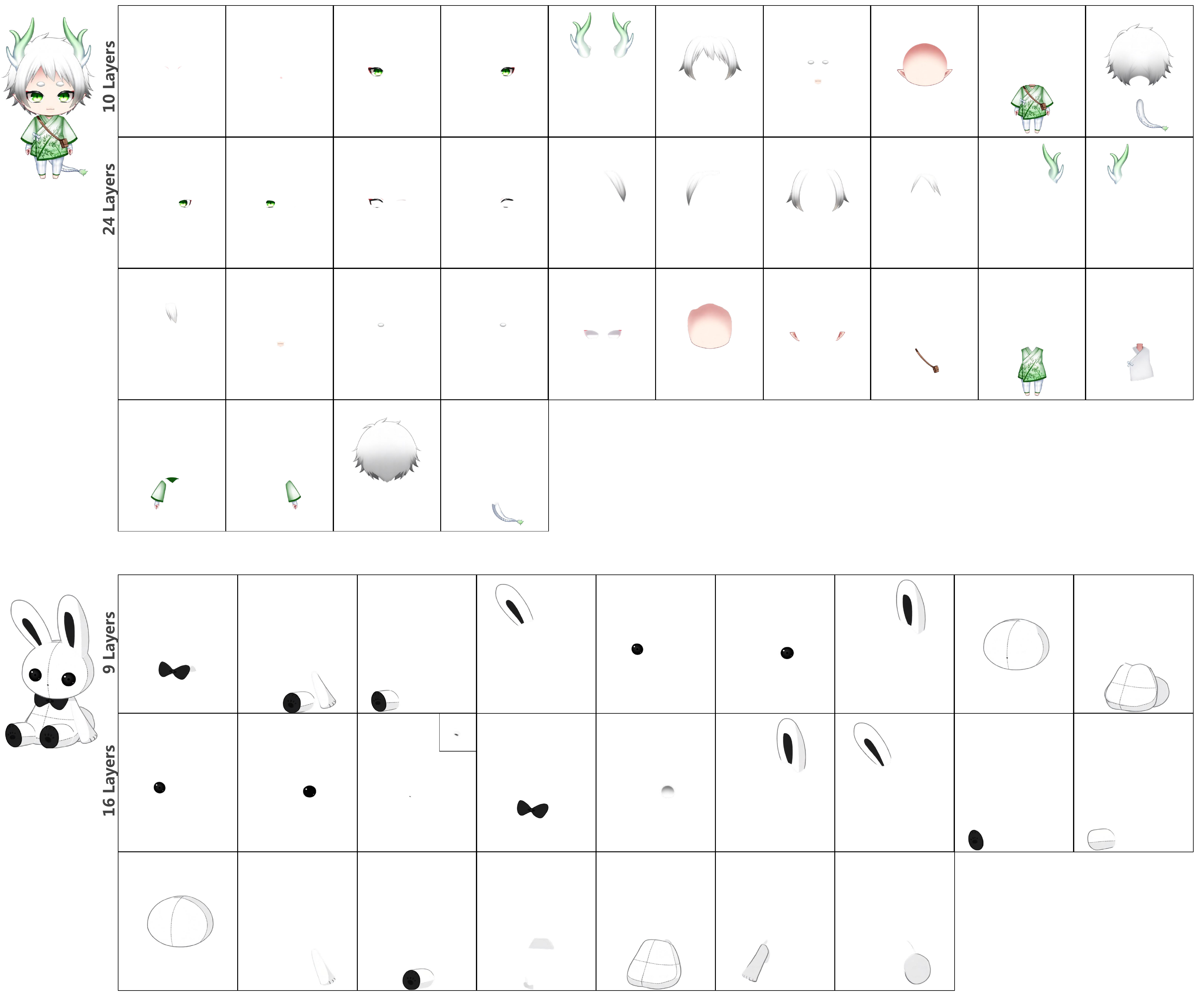}
  \caption{The reconstructed characters (left) are shown alongside their decomposed layer textures (right) under different specified layer counts. Our approach naturally adapts to free layer numbers, supporting both coarse and fine-grained separation.}
  \label{fig:stage1res6}
\end{figure*}

This section presents additional qualitative results for our Stage 1 layer decomposition. Visual comparisons against existing baselines are detailed in Figures~\ref{fig:stage1res1} and \ref{fig:stage1res2}. Additionally, Figures~\ref{fig:stage1res3} to \ref{fig:stage1res6} highlight our multi-layer generation capabilities, specifically illustrating our method's flexibility in supporting layer separation with controllable granularity.

\section{Superseded Stage~2 Design: Autoregressive Mesh-Token Stream}
\label{app:mesh_tokenization}

\paragraph*{Status of this appendix.}
The Stage~2 model of the main paper (\S\ref{sec:stage2}) is a joint continuous-regression network: it predicts per-vertex displacements for all layers of a character in one forward pass and never discretises geometry. This appendix documents an \emph{earlier} Stage~2 formulation that we no longer use, in which mesh connectivity and keypose offsets were serialised into a single autoregressive token stream over quantised coordinates. We retain it for two reasons: the layer-to-mesh construction it describes is still the basis of the content-conforming mesh used throughout the paper, and its compression statistics may be of independent interest to readers building token-based mesh models. Two caveats must be read with everything below. First, the accuracy figures reported for this design (cosines above $0.99$) were obtained under \emph{teacher forcing}, with the ground-truth prefix fed back at every decoding step; they measure next-token accuracy, not the quality of a generated rig, and they are not comparable with the true-generation numbers in the main text. Second, no number in this appendix supports any claim in the main paper.

\paragraph*{Operator alphabet.}
For a single connected component of a 2D triangle mesh with $V$ vertices and $F$ faces, the tokeniser emits a byte stream over the alphabet
$\Sigma = \{\langle$\texttt{COMP}$\rangle,\ \langle$\texttt{C}$\rangle,\ \langle$\texttt{L}$\rangle,\ \langle$\texttt{R}$\rangle,\ \langle$\texttt{E}$\rangle,\ \langle$\texttt{S}$\rangle,\ \langle$\texttt{B}$\rangle\}$
mixed with quantised vertex coordinates from the alphabet $\{0,\dots,Q-1\}$. The semantics follow EdgeBreaker ~\cite{rossignac1999edgebreaker} as adapted to neural mesh generation by EdgeRunner ~\cite{tang2024edgerunner}:
\begin{itemize}[leftmargin=1.5em,itemsep=1pt,topsep=2pt]
\item $\langle$\texttt{COMP}$\rangle$ starts a component and is followed by six integers, the three quantised $(x,y)$ pairs of the seed triangle's three vertices.
\item $\langle$\texttt{C}$\rangle$ (\emph{capture}) declares that the next face has a previously-unseen vertex as its third corner; it is followed by two integers giving its quantised coordinates.
\item $\langle$\texttt{L}$\rangle$ and $\langle$\texttt{R}$\rangle$ (\emph{previous-twin left / right}) declare that the third corner is already visited and that its respective side edge is already on the frontier; they carry no payload.
\item $\langle$\texttt{E}$\rangle$ (\emph{end}) closes a face whose two side edges are both already on the frontier; no payload.
\item $\langle$\texttt{S}$\rangle$ (\emph{split}) declares that the third corner is a visited vertex whose side edges are not on the frontier; it is followed by one integer, the compact identifier of that vertex.
\item $\langle$\texttt{B}$\rangle$ (\emph{boundary advance}) consumes a frontier edge that has no face behind it; no payload.
\end{itemize}
Encoder and decoder maintain the same frontier stack (a stack of edges with an opposite-corner annotation), so they always agree on which gate is being processed and can round-trip exactly.

\begin{figure*}[!tp]
  \centering
  \includegraphics[width=\linewidth]{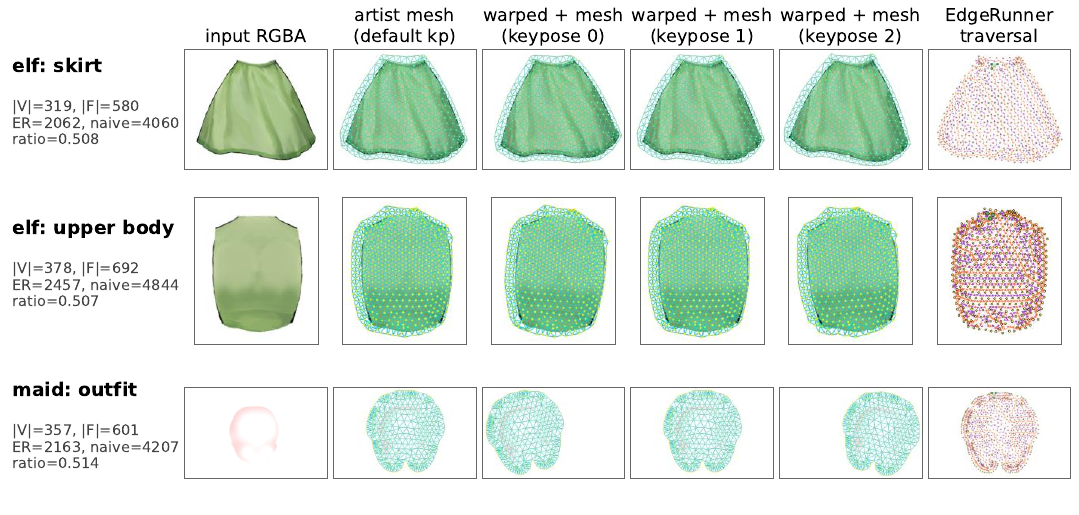}
  \caption{\textbf{Multi-character demonstration on garment layers} (superseded design). Three characters, all on
 full-body garment layers. Columns: the raw RGBA layer; the artist mesh at the default keypose; the same
 mesh re-warped to keyposes $0$/$1$/$2$ along the listed parameter axis; and the EdgeRunner per-face
 traversal emitted by the superseded Stage~2 (green seed triangle, orange capture arrows, purple split
 arrows, boundary operators trailing). Per-row token statistics are printed beside each row label.}
  \label{fig:multi_char_garments}
\end{figure*}

\begin{figure*}[!tp]
  \centering
  \includegraphics[width=\linewidth]{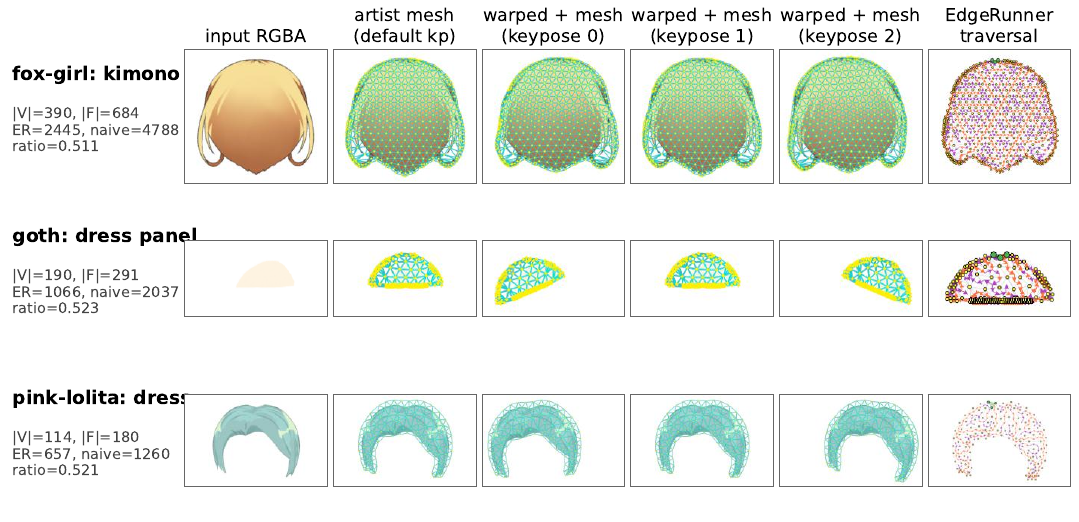}
  \caption{\textbf{Multi-character demonstration on body-component layers}, formatted identically to Fig.~\ref{fig:multi_char_garments}. Three rows show a fox-girl's kimono front panel, a goth-styled character's dress panel, and a pink-lolita's dress; together they span three new characters and three new layer shapes.}
  \label{fig:multi_char_body}
\end{figure*}

\begin{figure*}[!tp]
  \centering
  \includegraphics[width=\linewidth]{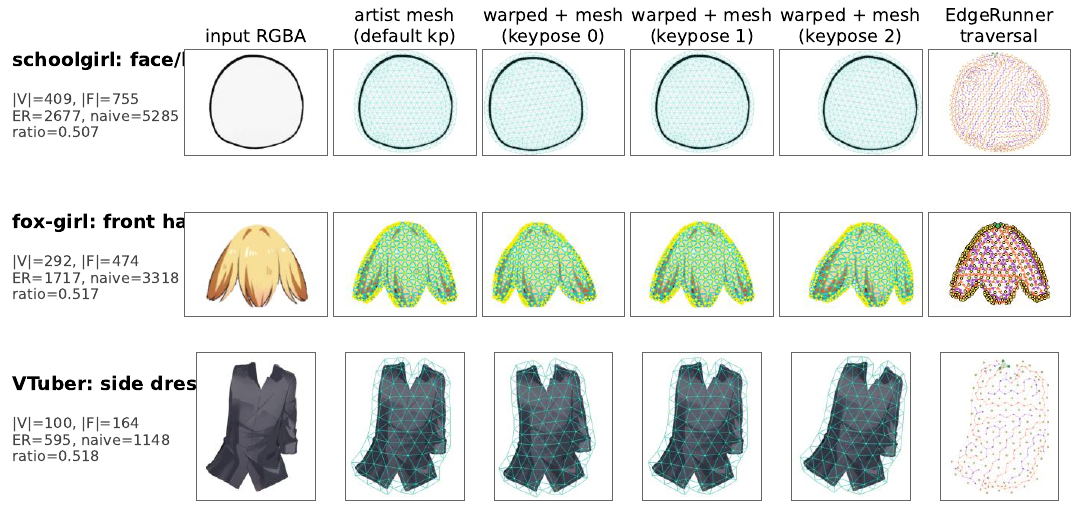}
  \caption{\textbf{Multi-character demonstration on head / hair layers} (superseded design), formatted as
 Fig.~\ref{fig:multi_char_garments}: a schoolgirl's full face/hair shape (the largest layer in the demo,
 $|V|\!=\!409$, $|F|\!=\!755$), a fox-girl's front hair, and a VTuber's side dress.}
  \label{fig:multi_char_head}
\end{figure*}

\paragraph*{Layer-to-mesh.}
At training time the input mesh is the artist-authored Live2D mesh, whose vertex count ranges from a few tens to several hundreds and whose face count grows roughly twice as fast. At inference on a novel illustration (no artist mesh), we synthesise a base mesh from the layer's alpha channel in six steps: (1)~threshold the alpha channel at $8/255$ to obtain a silhouette; (2)~morphologically dilate the silhouette by approximately $8$\,px to recover the artist's safety margin used by Live2D Cubism for mesh padding; (3)~walk the dilated boundary counter-clockwise and sample $n_b\!=\!24$ uniform points; (4)~inside the silhouette, draw $n_i\!=\!8$ Poisson-disk interior points with minimum spacing $20$\,px; (5)~run constrained Delaunay triangulation ~\cite{barber1996quickhull} on the combined point set and discard triangles whose centroid lies outside the dilated silhouette; (6)~quantise the resulting vertices to $Q\!=\!128$ bins per axis and emit them in the order required by the operator alphabet above. The full procedure together with the EdgeRunner per-face traversal that follows it is illustrated across multiple character / layer combinations in Figs.~\ref{fig:combined_elf_hair}, \ref{fig:combined_fox_kimono}, \ref{fig:combined_goth}, and~\ref{fig:combined_gumi} below; the mesh deformation across keyposes that the resulting tokens drive is shown across three multi-character composites in Figs.~\ref{fig:multi_char_garments}, \ref{fig:multi_char_body}, and~\ref{fig:multi_char_head}.

Across the nine rows of Figs.~\ref{fig:multi_char_garments} and~\ref{fig:multi_char_head} the
superseded design was exercised on seven characters (elf-girl, fox-girl, goth, pink-lolita, schoolgirl,
military maid, VTuber) and nine distinct layer types, and its EdgeRunner compression ratio stayed in the
band $0.507$ to $0.523$ throughout, so the token savings were independent of character identity, mesh size
and layer shape.

\paragraph*{Token stream size.}
The total token count of a single-component mesh is
$T = 1 + 6 + 3\,N_\texttt{C} + N_\texttt{L} + N_\texttt{R} + N_\texttt{E} + 2\,N_\texttt{S} + N_\texttt{B}$,
where $N_x$ counts the number of $\langle$\texttt{x}$\rangle$ operators emitted. The $6$ accounts for the seed-triangle vertex payload, each $\langle$\texttt{C}$\rangle$ adds $1$ operator plus $2$ coordinate integers, and each $\langle$\texttt{S}$\rangle$ adds $1$ operator plus $1$ identifier integer. Crucially, every interior vertex of the mesh is emitted by exactly one $\langle$\texttt{C}$\rangle$, so $N_\texttt{C}\!=\!V-3$ for a fully connected component, and every boundary edge contributes exactly one $\langle$\texttt{B}$\rangle$. A naive baseline that simply writes each triangle as $\langle$\texttt{F}$\rangle$ followed by $3$ $(x,y)$ pairs uses $7F$ tokens. The compression ratio between EdgeRunner and naive is thus a near-constant function of the boundary-to-face ratio and is largely independent of mesh size.

\paragraph*{Compression statistics on real Live2D data.}
We measured EdgeRunner and naive token counts on every layer of the Tsumugi character ($100$ layers, $|V|\!=\!4156$, $|F|\!=\!5941$). The aggregate compression is reported in Tab.~\ref{tab:mesh_tok_stats}. EdgeRunner produces $22{,}297$ tokens against $41{,}587$ for naive, a $0.536$ ratio ($46.4\%$ savings), with $3.75$ tokens per face on average versus $7.0$ for naive. The breakdown of the operator alphabet is dominated by $\langle$\texttt{B}$\rangle$ (boundary advances, $50.9\%$) and $\langle$\texttt{C}$\rangle$ (interior captures, $32.3\%$), with $\langle$\texttt{S}$\rangle$ (splits, $16.8\%$) accounting for the rest. The $\langle$\texttt{P}$\rangle$ operator was never triggered on this character, consistent with the under-2\% \texttt{P}-fraction we observe across the full $662$-character training set.

\begin{table}[H]
  \centering
  \caption{Mesh-tokenisation statistics on a representative Live2D character (Tsumugi, $100$ layers). EdgeRunner tokens correspond to our operator-plus-payload count; the naive baseline encodes each triangle as a fresh $\langle$\texttt{F}$\rangle$ marker plus three quantised $(x,y)$ pairs.}
  \label{tab:mesh_tok_stats}
  \small
  \setlength{\tabcolsep}{6pt}
  \renewcommand{\arraystretch}{1.05}
  \begin{tabular*}{\linewidth}{@{\extracolsep{\fill}}lr@{}}
    \toprule
    Quantity & Value \\
    \midrule
    Layers (with non-empty mesh)                & $100$ \\
    Total vertices $|V|$                        & $4{,}156$ \\
    Total faces $|F|$                           & $5{,}941$ \\
    \midrule
    EdgeRunner tokens                           & $22{,}297$ \\
    Naive XYZ tokens                            & $41{,}587$ \\
    Compression ratio (EdgeRunner / Naive)      & $0.536$ \\
    Bytes saved relative to naive               & $46.4\%$ \\
    Average EdgeRunner tokens per face          & $3.75$ \\
    Average Naive tokens per face               & $7.00$ \\
    \midrule
    Operator $\langle$\texttt{B}$\rangle$ (boundary)        & $50.9\%$ \\
    Operator $\langle$\texttt{C}$\rangle$ (capture)         & $32.3\%$ \\
    Operator $\langle$\texttt{S}$\rangle$ (split)           & $16.8\%$ \\
    Operator $\langle$\texttt{P}$\rangle$ (pop / backtrack) & $\;\,0.0\%$ \\
    \bottomrule
  \end{tabular*}
\end{table}

\paragraph*{Cross-character examples.}
We exercise the full layer-to-mesh-to-token pipeline on four character / layer combinations drawn from \emph{four different characters} spanning distinct anime aesthetics, summarised in Tab.~\ref{tab:cross_char_demos} and shown panel-by-panel in Figs.~\ref{fig:combined_elf_hair}, \ref{fig:combined_fox_kimono}, \ref{fig:combined_goth}, and~\ref{fig:combined_gumi}. Each figure is a single $2\!\times\!4$ grid in which panels~1 to 7 step through the layer-to-mesh procedure and panel~8 overlays the EdgeRunner per-face traversal that the autoregressive model emits on the same artist mesh. The seed triangle is drawn in green, $\langle$\texttt{C}$\rangle$ (capture) arrows in orange, $\langle$\texttt{S}$\rangle$ (split) arrows in purple, and $\langle$\texttt{B}$\rangle$ (boundary advance) operators are folded into the stream tail. Across all four cases the compression ratio against the naive baseline stays in the narrow band $0.512$ to $0.523$, which matches the constant savings predicted by the per-face cost analysis and confirms that the savings are independent of character identity, mesh size, and layer shape.

\begin{table}[H]
  \centering
  \caption{Per-figure token statistics for the four cross-character demos.}
  \label{tab:cross_char_demos}
  \scriptsize
  \setlength{\tabcolsep}{4pt}
  \renewcommand{\arraystretch}{1.05}
  \begin{tabular*}{\linewidth}{@{\extracolsep{\fill}}llrrrrr@{}}
    \toprule
    Character & Layer description & $|V|$ & $|F|$ & EdgeRunner & Naive & Ratio \\
    \midrule
    elf-girl                & side hair         & 249 & 432 & 1{,}548 & 3{,}024 & 0.512 \\
    fox-girl (kimono)       & front hair        & 292 & 474 & 1{,}717 & 3{,}318 & 0.517 \\
    goth (with harness)     & half-circle bust  & 190 & 291 & 1{,}066 & 2{,}037 & 0.523 \\
    pink lolita             & front dress       & 147 & 244 &   874 & 1{,}708 & 0.512 \\
    \bottomrule
  \end{tabular*}
\end{table}

\begin{figure*}[tp]
  \centering
  \includegraphics[width=\linewidth]{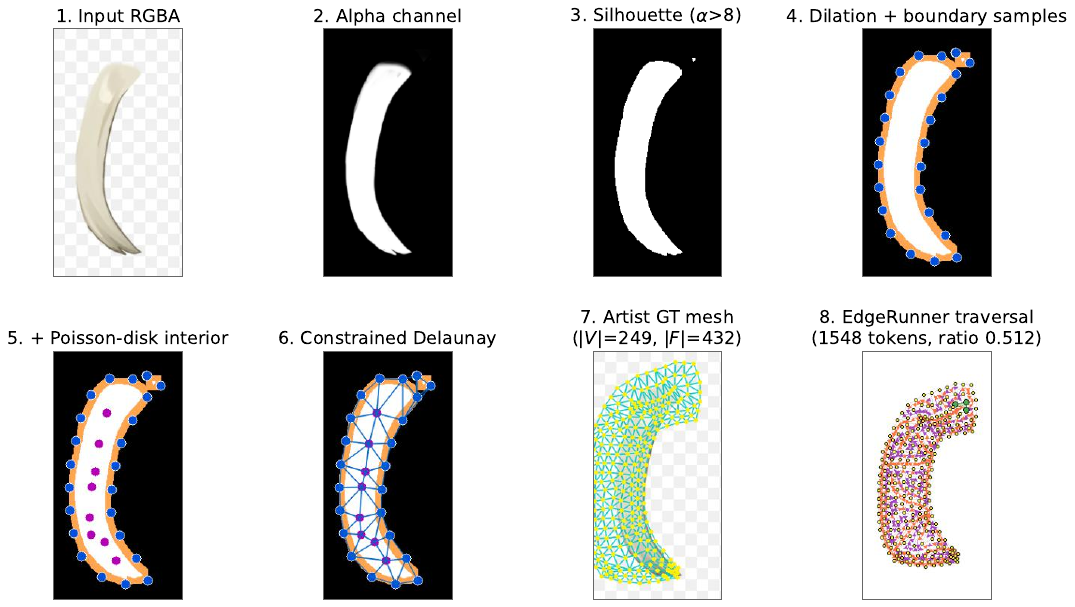}
  \caption{Layer-to-mesh pipeline and EdgeRunner traversal on a long flowing side hair layer of an elf-girl character ($|V|\!=\!249$, $|F|\!=\!432$). Panels~1 to~7 step through the auto-mesh procedure; panel~8 shows the per-face traversal that produces a $1{,}548$-token EdgeRunner stream, $0.512\times$ the $3{,}024$-token naive baseline.}
  \label{fig:combined_elf_hair}
\end{figure*}

\begin{figure*}[tp]
  \centering
  \includegraphics[width=\linewidth]{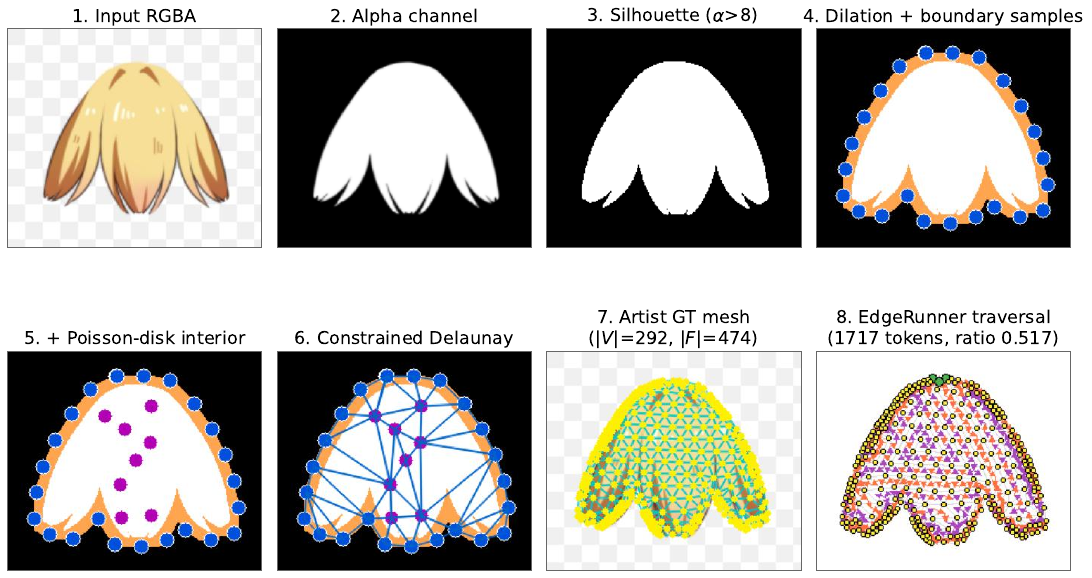}
  \caption{Layer-to-mesh pipeline and EdgeRunner traversal on a fox-girl character's front hair / fringe layer ($|V|\!=\!292$, $|F|\!=\!474$). Although the shape and aesthetic differ markedly from Fig.~\ref{fig:combined_elf_hair}, the resulting compression ratio ($0.517$) is essentially identical.}
  \label{fig:combined_fox_kimono}
\end{figure*}

\begin{figure*}[tp]
  \centering
  \includegraphics[width=\linewidth]{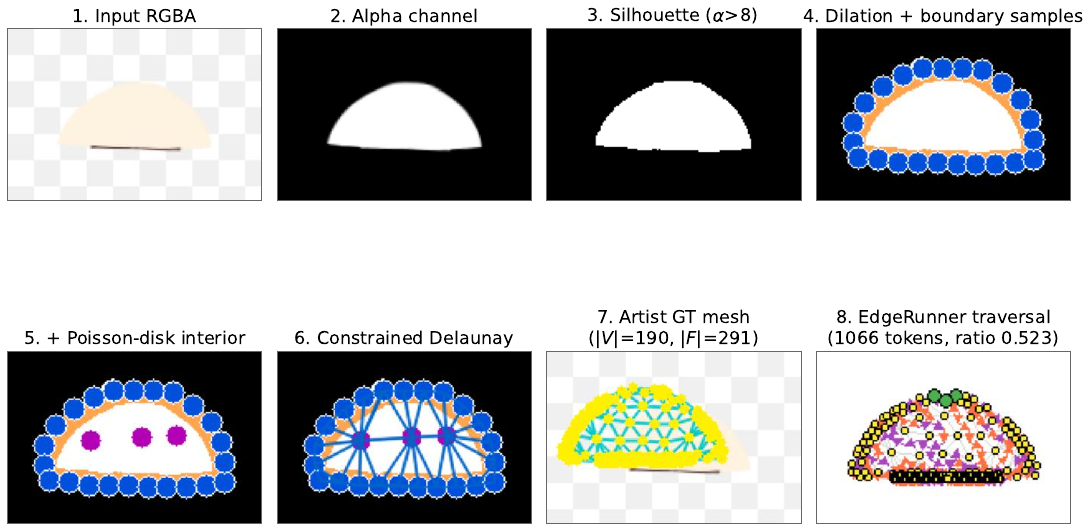}
  \caption{Layer-to-mesh pipeline and EdgeRunner traversal on a goth-styled character's half-circle bust layer ($|V|\!=\!190$, $|F|\!=\!291$). The smaller layer produces a $1{,}066$-token EdgeRunner stream, ratio $0.523$ against the $2{,}037$-token naive baseline.}
  \label{fig:combined_goth}
\end{figure*}

\begin{figure*}[tp]
  \centering
  \includegraphics[width=\linewidth]{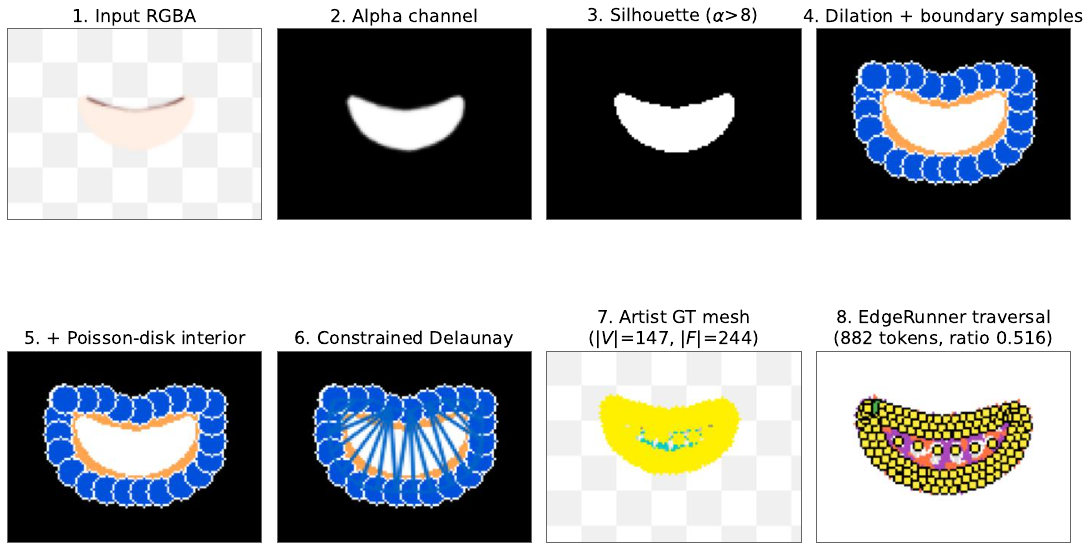}
  \caption{Layer-to-mesh pipeline and EdgeRunner traversal on a pink-lolita character's front dress layer ($|V|\!=\!147$, $|F|\!=\!244$). On a smaller layer the absolute token count drops to $874$, but the ratio against the naive baseline stays at $0.512$, demonstrating that compression savings are mesh-size invariant.}
  \label{fig:combined_gumi}
\end{figure*}

\paragraph*{Putting it all together.}
In this superseded design, the full per-layer record stitched together (i) a mesh block tokenised as above, (ii) a single $\langle$\texttt{SCALE}$\rangle$ token whose hidden state was read by a magnitude-regression head, and (iii) two keypose blocks each emitting $|V|$ quantised $(\Delta x, \Delta y)$ offsets. For a typical $60$-layer character, the token stream compressed from approximately $90$\,K JSON tokens to between $18$\,K and $22$\,K tokens, which is the regime in which a from-scratch $12$\,M-parameter causal Transformer trained stably. The joint regression model that replaced it needs no such serialisation: one vertex is one token, and a $60$-layer character is a single sequence of a few thousand tokens processed in one non-causal pass.

\section{Displacement-Magnitude Distribution}
\label{app:alpha}

Stage~2 factors every displacement into a bounded shape term and a $\log$-scale term (Eqs.~\ref{eq:stage2_heads} and~\ref{eq:stage2_recompose}) because the magnitude distribution is extremely heavy-tailed. Measuring the per-layer peak displacement $a=\max_v\|\Delta v\|_\infty$ over the full training corpus, $80\%$ of (layer, parameter, keypose) records concentrate at $a<0.05$ of canvas extent, with a long tail past $a>0.4$; the density is only readable on a log scale. Two consequences follow. First, regressing raw displacements would let the tail dominate the loss, since a single large head turn contributes two orders of magnitude more gradient than a typical eyelid motion; normalising by $a$ and predicting $\log a$ separately makes the two regimes contribute comparably, so that a $2\times$ magnitude error costs the same whether the true motion is $0.002$ or $0.2$. Second, the same statistic explains why a uniform quantisation of displacements is a poor representation for this domain: a $Q\!=\!128$-level grid over $[-1,1]$ would collapse the entire small-motion mode into roughly the central six bins, which is what motivated the per-record normalisation used by the superseded design of Appendix~\ref{app:mesh_tokenization} and what the continuous formulation avoids outright.

\section{Stage 2 Inference and Live2D Runtime Interpolation}
\label{app:inference}

Stage~2 inference is one forward pass per (parameter, keypose) pair. All layers of the character are concatenated into a single sequence of vertex tokens (Eq.~\ref{eq:stage2_joint}); the direction head emits $u_{i,j}=\tanh(\cdot)\in[-1,1]^2$ and the magnitude head emits $s_{i,j}$, and the keypose offset is their product $\widehat{\Delta v}_{i,j}=u_{i,j}\exp(s_{i,j})$ (Eq.~\ref{eq:stage2_recompose}). There is no autoregressive rollout and therefore no exposure-bias gap between training and inference: the model is evaluated in exactly the mode in which it is trained, which is why every Stage~2 number in the main text is a true-generation number. Sweeping the modelled parameters over their keyposes produces the full displacement table of Eq.~\ref{eq:l2d_model}. At runtime the parameter slider drives per-vertex linear interpolation between the two bracketing keyposes (Eq.~\ref{eq:runtime_interp}), followed by per-triangle rasterisation with premultiplied alpha and no face culling. The full deformation runs at $30$\,fps in-browser.

\section{Layer-to-Mesh Deployment Study (superseded Stage~2)}
\label{app:layer_to_mesh}

\begin{figure*}[tp]
  \centering
  \includegraphics[width=\linewidth]{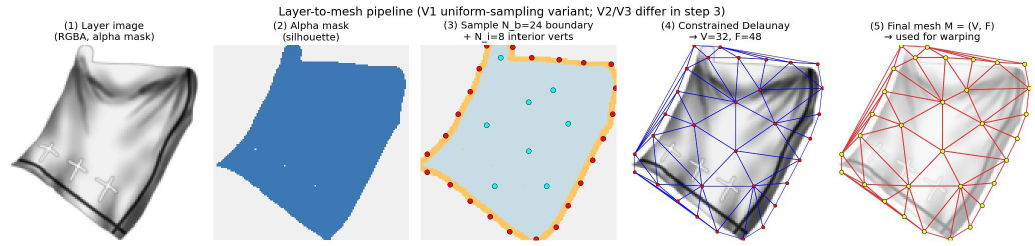}
  \caption{\textbf{Layer-to-mesh pipeline, V1 uniform-sampling variant} (superseded design), on a real square
 garment layer. Five steps: input RGBA and alpha mask; silhouette at $\alpha\!>\!8$; dilation of about
 $8$\,px (orange ring) with $n_b\!=\!24$ uniform boundary points and $n_i\!=\!8$ Poisson-disk interior
 points; constrained Delaunay giving $|V|\!=\!32$ and $|F|\!\approx\!48$; and the final mesh in
 canvas-normalised $[-1,1]^2$. Note that the shipped construction of \S\ref{sec:stage2} differs at several
 of these steps; the variants are listed in the text.}
  \label{fig:layer_to_mesh}
\end{figure*}

V1 to V5 differ only at the sampling step: V1 uses a random interior, V2 Poisson-disk
sampling~\cite{bridson2007fast}, V3 curvature-aware boundary sampling, V4 V3 plus Lloyd
CVT~\cite{lloyd1982least}, and V5 Shewchuk's Triangle~\cite{shewchuk1996triangle} at \texttt{pq30}. None
of these is the shipped construction: \S\ref{sec:stage2} masks at $\alpha\!>\!4$ rather than
$\alpha\!>\!8$, dilates by $3$\,px rather than $8$, samples the interior on a jittered lattice rather than
by Poisson disk, and uses an unconstrained Delaunay triangulation.

\paragraph{Auto-mesh + retraining.}
This study was run with the superseded autoregressive Stage~2 of Appendix~\ref{app:mesh_tokenization}, under a protocol that supplied the mesh prefix; its cosines are therefore \emph{not} comparable with the true-generation numbers of the main text and are reported only for the relative ordering of mesh-construction algorithms, which is what it was designed to establish. At training time we use artist meshes; at deployment time the input is a Stage 1 PSD layer (pixel + alpha only), so we automatically generate a base mesh: Suzuki-Abe ~\cite{suzuki1985topological} contour, $n_b$ uniform boundary points, $n_i$ Poisson-disk interior, constrained Delaunay ~\cite{barber1996quickhull}, drop center-outside triangles. We additionally study (i) dilation-px sweep $\{0,8,16,24\}$ to mimic the artist's $\sim$17\,px safety margin, and (ii) algorithm sweep V1 to V5. Tab.~\ref{tab:layer_to_mesh} reports $50$-char OOD results.

\textbf{Findings.} (i) Inference-only auto-mesh: cosine drops uniformly to $\sim\!0.91$ across all 5 algorithms (V1 to V5 within $0.001$); the gap is \emph{vertex-distribution} mismatch, not geometric quality. (ii) \emph{Retraining} on auto-meshes (paired GT offsets via inverse-distance weighting) closes it: V1/V2/V3 retrain all reach cos $\geq 0.9998$ within 5 epochs; V3 curvature-aware additionally gives the lowest mag (2.91) and best RMSE in pixel space (1.99\,px). \emph{Honest caveat:} retrained cos $0.9999$ is not directly comparable to artist-baseline $0.997$; V3 auto-mesh has $V_u{=}32$ uniform vertices vs $V{=}30$ to $100$ sharp / non-uniform artist verts, so the task is simpler. Pixel-domain RMSE is more honest: V1 retrain $2.15$\,px vs artist $0.99$\,px (still sub-3\,px); PCK@10 $0.977$ vs $0.993$.

\begin{table}[hbtp]
  \centering
  \caption{Layer-to-mesh deployment study on $50$-char OOD, run with the \emph{superseded} autoregressive Stage~2 (Appendix~\ref{app:mesh_tokenization}) under a mesh-prefix-supplied protocol. Read the relative ordering of mesh algorithms, not the absolute cosines: these are not true-generation numbers and are not comparable with Tab.~\ref{tab:stage2_mesh}.}
  \label{tab:layer_to_mesh}
  \scriptsize
  \setlength{\tabcolsep}{3pt}
  \resizebox{\columnwidth}{!}{%
  \begin{tabular}{l c c c c}
    \toprule
    Mesh source                                & vert.\,cos↑ & mag→1   & $\alpha$-err↓ & Reference \\
    \midrule
    Artist mesh (training distribution)        & \textbf{0.997} & \textbf{7.73} & \textbf{0.320} &, \\
    \midrule
    \multicolumn{5}{l}{\emph{Inference-only auto-mesh:}} \\
    V1 uniform, dilate $0$                     & 0.916          & 22.53 & 0.506 &, \\
    V1 uniform, dilate $16$                    & 0.916          & 19.39 & 0.482 &, \\
    V1 uniform (random interior)               & 0.908          & 17.81 & 0.485 & legacy \\
    V2 Poisson-disk interior                   & 0.908          & 19.27 & 0.491 & \citet{bridson2007fast} \\
    V3 curvature-aware boundary                & 0.908          & \textbf{15.31} & 0.494 & Live2D-style \\
    V4 V3 $+$ Lloyd CVT (4 iter)               & 0.908          & 18.25 & 0.510 & \citet{lloyd1982least} \\
    V5 Triangle \texttt{pq30}                  & 0.907          & 17.84 & 0.473 & \citet{shewchuk1996triangle} \\
    \midrule
    \multicolumn{5}{l}{\emph{Retrained on auto-mesh (deployment-aware, ours):}} \\
    \textbf{V1 uniform retrain}                & \textbf{0.9999} & 3.84  & 0.328 & this work \\
    \textbf{V2 Poisson retrain}                & \textbf{0.9998} & 3.19  & 0.336 & this work \\
    \textbf{V3 curvature retrain}              & \textbf{0.9999} & \textbf{2.91} & 0.324 & this work \\
    \textbf{V1+barycentric interp}             & \textbf{0.9999} & 3.34  & \textbf{0.256} & this work \\
    \bottomrule
  \end{tabular}}
\end{table}

\paragraph*{Multi-component layers.}
The single-contour formulation above (largest external contour) silently drops
disconnected components that occur routinely in in-the-wild decompositions: e.g.\ a
side-hair layer split into a main lock plus a detached tail, or a garment layer separated
by an occluding arm. We therefore extend the auto-mesh to \emph{all} external contours
above a minimum-area threshold: each component contributes its own arc-length-sampled
boundary loop to the PSLG, boundary-point budget is allocated proportional to component
area, and constrained Delaunay is run once over the union before dropping centroid-outside
triangles. This keeps every visible part meshed, while remaining identical to the single-contour
procedure when a layer has one component. We note for the record that we initially
justified this design by claiming a bounding-box grid wastes most of its vertices on
transparent pixels; measuring it on $46$ held-out layers refuted that claim (on-content
vertex fraction $0.376$ for the grid against $0.368$ for ours, silhouette IoU $0.426$ for
both), because a layer's own alpha bounding box is mostly filled. The defensible advantage
is cost at equal coverage: $83$ vertices and $112$ faces per layer against $112$ and $182$,
which is $26\%$ fewer tokens for the same opaque coverage, with the budget concentrated
along contours rather than spread uniformly.

\section{Animation Results Under Many Conditions}
\label{app:sheets}

The numbers in \S\ref{sec:results_stage2} and \S\ref{sec:results_ablations} each compress a whole
condition into one scalar. This appendix shows the same conditions as pictures, because some of them
are only convincing that way and one of them (capacity) is more convincing that way. Every cell below
is a raw render of a real rig file through the same viewer at an \emph{absolute} parameter value, so
any cell can be checked against the released rig; nothing is hand-picked or per-example fixed.

\begin{figure*}[p]
  \centering
  \includegraphics[width=\linewidth]{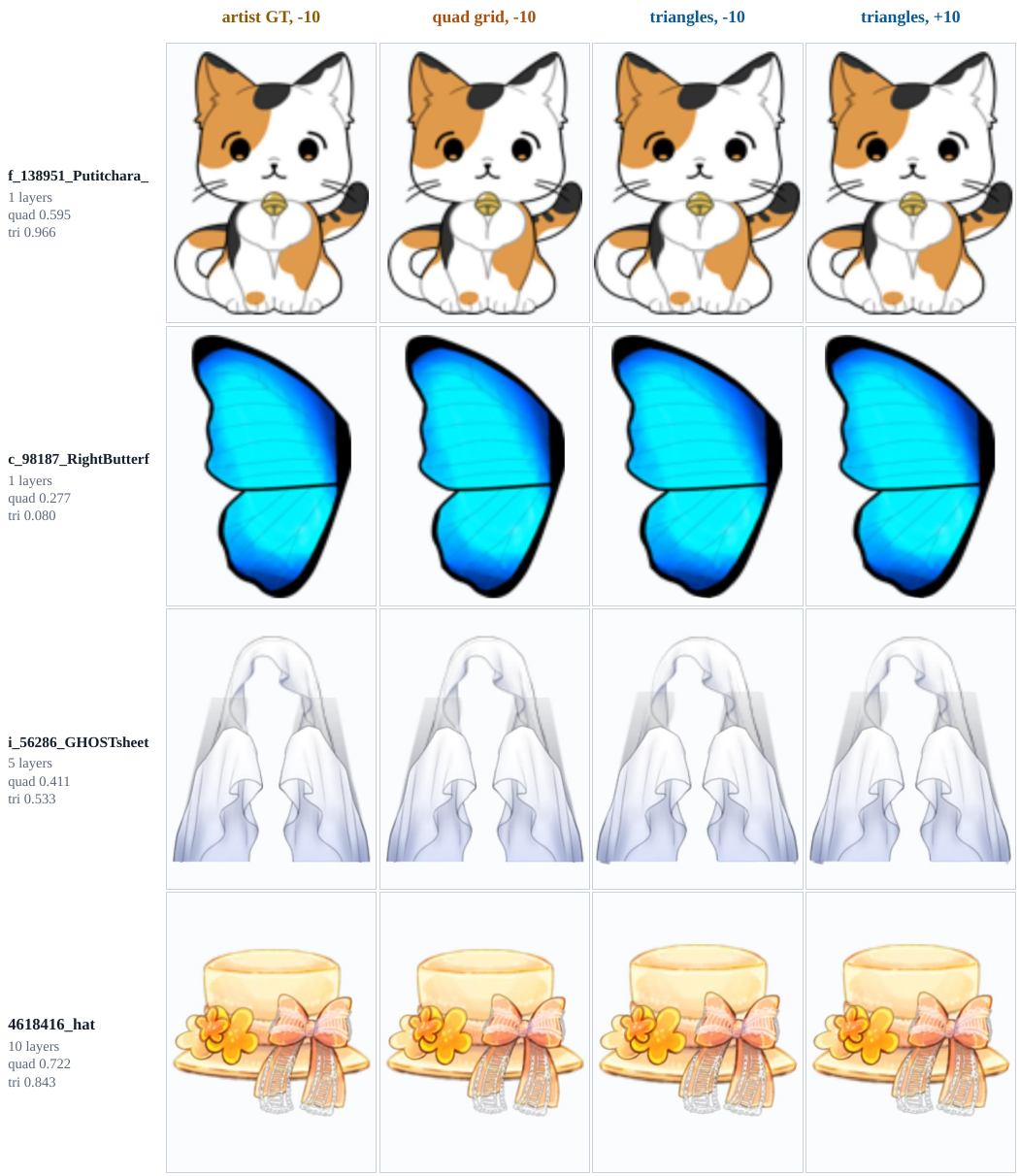}
  \caption{\textbf{Mesh representation, same character and pose.} Artist ground truth, the prediction on a quad grid, and the prediction on content-conforming triangles, at $\texttt{ParamAngleX}\!=\!+30$. Layer count and each condition's per-character direction cosine are printed beside the row. Characters are matched by name across the two rig sets.}
  \Description{Six characters shown as artist ground truth next to predictions on a quad grid and on content-conforming triangles.}
  \label{fig:app_mesh}
\end{figure*}

\paragraph*{Mesh representation (Fig.~\ref{fig:app_mesh}).}
Quad grid against content-conforming triangles, with the artist's own rig driven to the same
parameter value as a third column. Characters are matched \emph{by name} across the two rig sets
because their internal ids differ, and the per-character cosine of each condition is printed beside
the row. The visual difference between the two meshes is small at this scale, which is consistent with
the $0.7676$ against $0.7542$ scalar gap; what the figure adds is that neither mesh produces a
qualitatively different kind of error, so the choice between them is a matter of token cost and
accuracy rather than of failure mode.

\begin{figure*}[p]
  \centering
  \includegraphics[width=\linewidth]{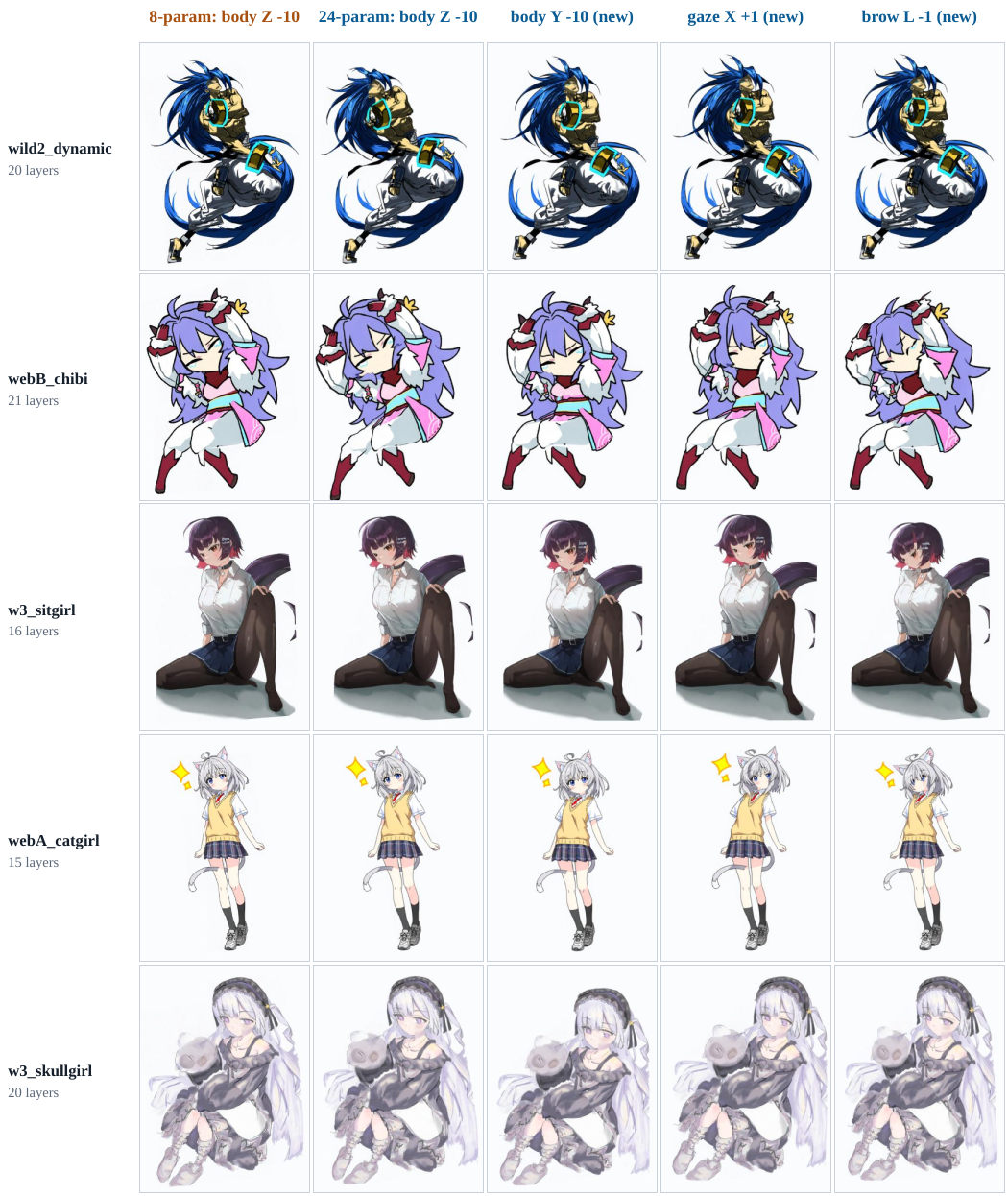}
  \caption{\textbf{$8$-parameter against $24$-parameter model.} Same illustrations, same Stage-1 layers, only the animation model differs. Columns~1--2 are the shared parameter \texttt{ParamAngleX} under each model; columns~3--5 are parameters only the $24$-parameter model has (gaze, breathing, front-hair sway).}
  \Description{Six in-the-wild characters animated by the 8-parameter and 24-parameter models, plus three parameters unique to the larger model.}
  \label{fig:app_params}
\end{figure*}

\paragraph*{Parameter vocabulary (Fig.~\ref{fig:app_params}).}
The same in-the-wild illustrations and the same Stage-1 layers, animated once by the $8$-parameter
model and once by the $24$-parameter model, followed by three parameters only the larger vocabulary
has. The first two columns are the visual form of the controlled comparison in
Tab.~\ref{tab:p24_controlled}: on the parameter both models share, they are hard to tell apart, which
is what a $0.031$ cosine difference should look like. The last three columns are the reason to prefer
the larger vocabulary at all, since gaze, breathing and hair sway have no counterpart in the smaller
one.

\begin{figure*}[p]
  \centering
  \includegraphics[width=\linewidth]{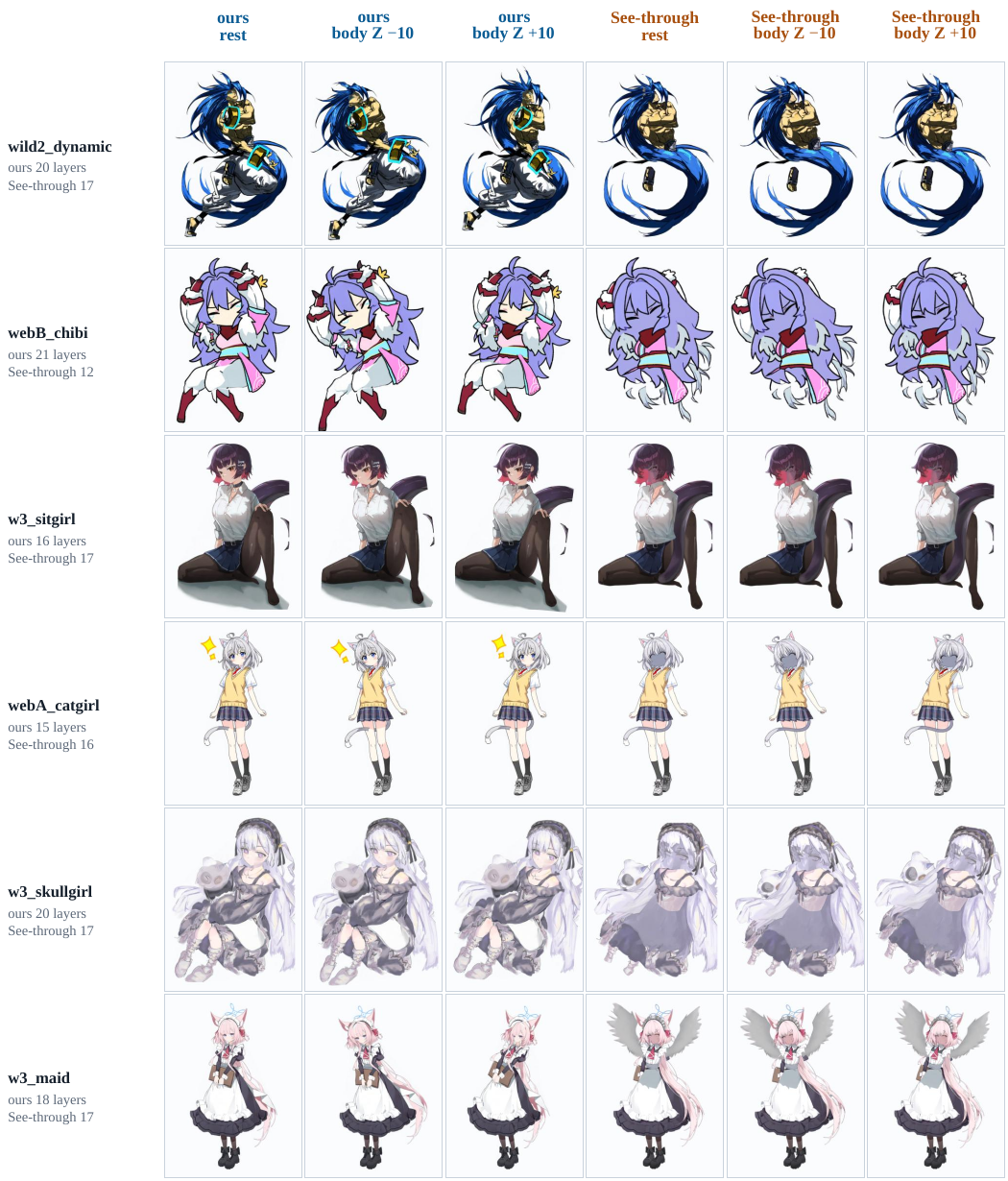}
  \caption{\textbf{Layer source is the only variable.} One frozen $24$-parameter model animates our Stage-1 decomposition and the See-through decomposition of the same illustration, each shown at rest and at $\texttt{ParamAngleX}\!=\!+30$. Layer counts are printed beside the row. Differences already visible in the rest columns are decomposition errors, not animation errors.}
  \Description{Seven illustrations decomposed two ways, each shown at rest and turned.}
  \label{fig:app_sources}
\end{figure*}

\paragraph*{Layer source (Fig.~\ref{fig:app_sources}).}
One frozen $24$-parameter animation model animating two decompositions of each illustration, at rest
and at a large turn. Showing rest as well as the pose matters: several of the differences that look
like animation errors are already present at rest, which localises them in Stage~1 rather than
Stage~2. This is the qualitative companion to the quantitative three-way comparison of
Appendix~\ref{app:crossdecomp}, where the artist's own layers form a third condition and bound what
Stage~2 can achieve.

\begin{figure*}[t]
  \centering
  \includegraphics[width=\linewidth]{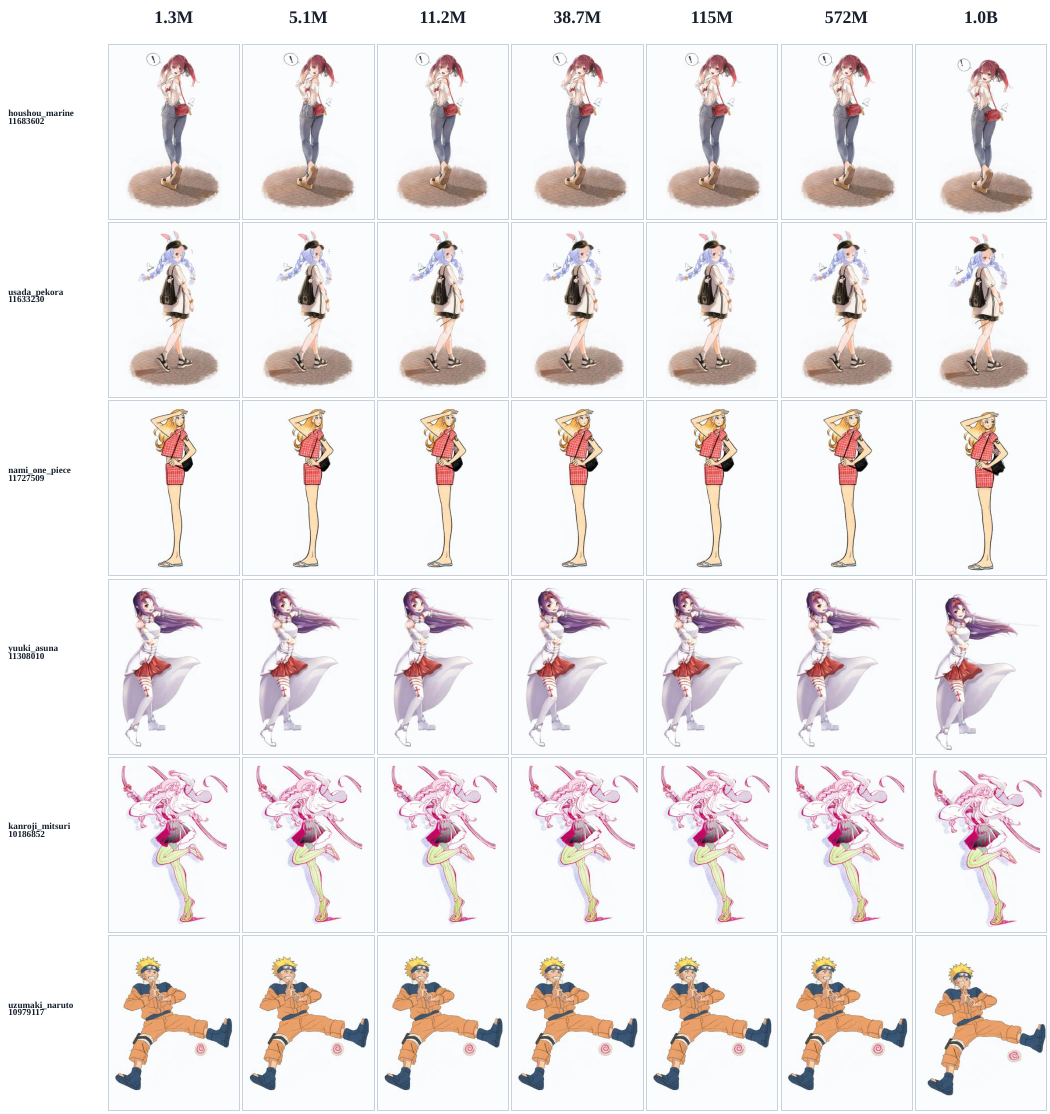}
  \caption{\textbf{Capacity is not the bottleneck.} Every value is a row of Tab.~\ref{tab:stage2_ablations}
 section~B: six models on identical data, objective and schedule across a $112\times$ parameter range.
 (a)~direction cosine never improves; the band is the measured same-config seed spread of $0.024$, and the
 $1.0$\,B run diverged so it has no score to plot. (b)~amplitude calibration is the one thing capacity
 buys. (c)~the count of rigs above $0.80$ agrees with (a), and the smallest model ties for the most.}
  \Description{Three line and bar charts of direction cosine, magnitude ratio and per-character count against model parameter count, all flat or declining.}
  \label{fig:app_scale}
\end{figure*}

\paragraph*{Capacity (Fig.~\ref{fig:app_scale}).}
Seven models from $1.3$\,M to $1.0$\,B parameters, trained on identical data with an identical
objective and schedule, all animating the same illustration at the same parameter value. The columns
are nearly indistinguishable. We consider this the most persuasive form of the
capacity result, because a reader who distrusts our metric can still see that a $770\times$ increase in
parameters does not visibly change the output. Note that the $1.0$\,B column is not empty: that run
diverged in training (validation cosine $-0.085$), and what it produces is a small, nearly uniform
displacement, which is what a collapsed model looks like rather than a crash.

\section{An Atlas of Every Parameter's Displacement Field}
\label{app:atlas}

A per-parameter table of cosines (Tab.~\ref{tab:p24_perparam}) says which parameters were learned but
not \emph{what} they learned. Reviewers of an earlier version raised exactly this: the animation looked
single-axis and the distribution of motion types was never shown. This appendix answers it directly. One
row per parameter, one column per character, and in each cell the predicted displacement field drawn as
vector arrows over the rest mesh, read straight out of the released rig files with no inference run.

The figure is meant to be read in two directions. Reading \emph{down} a column checks semantic
correctness without trusting any metric: the head parameters move the head group and leave the feet
alone, breathing moves the torso, the hair parameters move only hair, gaze moves two iris layers of a
few hundred pixels and nothing else. Reading \emph{across} a row checks consistency: the same parameter
should behave the same way on characters with different layer counts, proportions and art styles, and it
does. The parameters the model failed to learn are included and marked in red rather than omitted, and
their fields look the way a failure should look, either near-zero or pointing somewhere unrelated.

Two details make the atlas honest. Each cell uses that parameter's \emph{extreme} keypose, so no row is
flattered by a small value. And the arrow scale is fixed per cell by the $99$th percentile of that
cell's own displacement magnitudes, so a parameter with genuinely tiny motion looks tiny rather than
being renormalised into looking healthy.

\begin{figure*}[p]
  \centering
  \includegraphics[width=\linewidth]{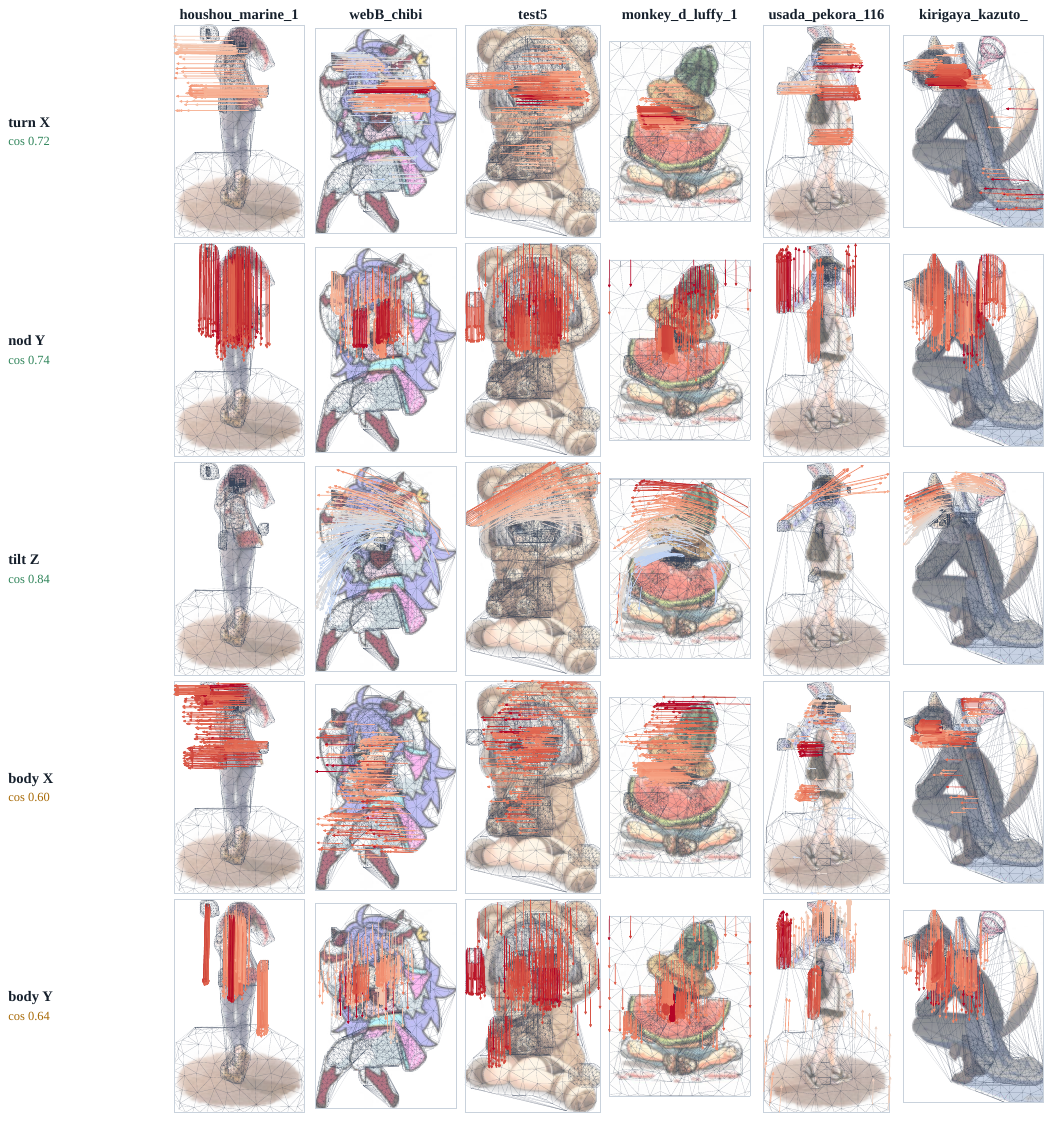}
  \caption{\textbf{Displacement field of every animation parameter, sheet 1: head and body rotation.}
  One row per parameter, one column per character, six in-the-wild characters ordered by measured motion
  magnitude. Arrows are the predicted displacement at that parameter's extreme keypose, exaggerated
  $3\times$ and coloured blue (small) to red (large), over the rest mesh in grey. The cosine beside each
  row is that parameter's measured quality, coloured green above $0.70$, amber between $0.40$ and
  $0.70$, red below. Note the qualitative signature of each axis: turn and body X are horizontal fields,
  nod and body Y vertical, and tilt Z is a rotation fan about the head.}
  \Description{Head and body rotation parameters shown as displacement fields on six characters.}
  \label{fig:atlas1}
\end{figure*}

\begin{figure*}[p]
  \centering
  \includegraphics[width=\linewidth]{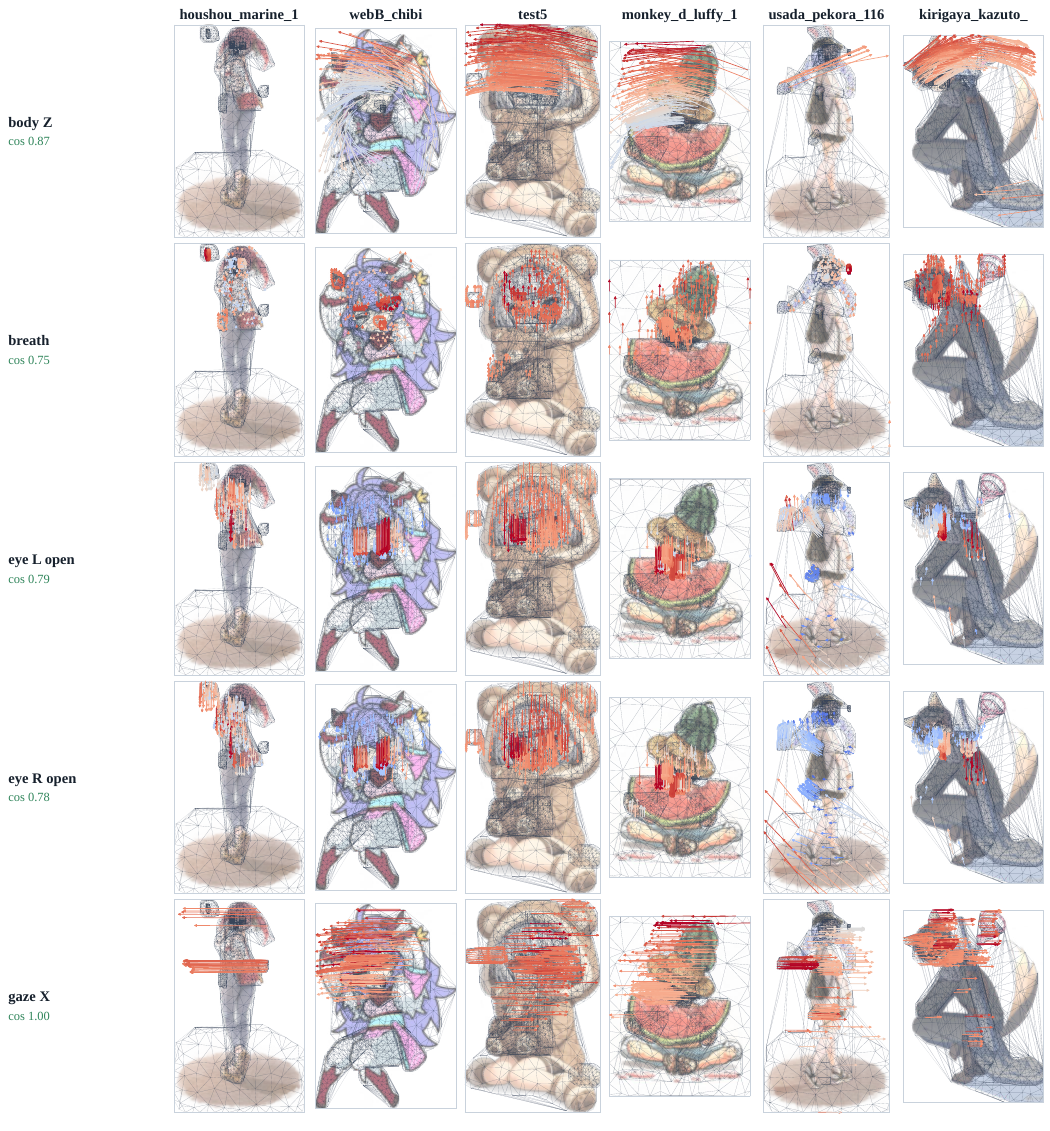}
  \caption{Sheet 2: body sway, breathing and eye opening. Format as Fig.~\ref{fig:atlas1}. Breathing is
  the clearest case of correct localisation without supervision of localisation: the field concentrates
  on the torso and decays to nothing at the head and the feet.}
  \Description{Body sway, breathing and eye-open parameters as displacement fields.}
  \label{fig:atlas2}
\end{figure*}

\begin{figure*}[p]
  \centering
  \includegraphics[width=\linewidth]{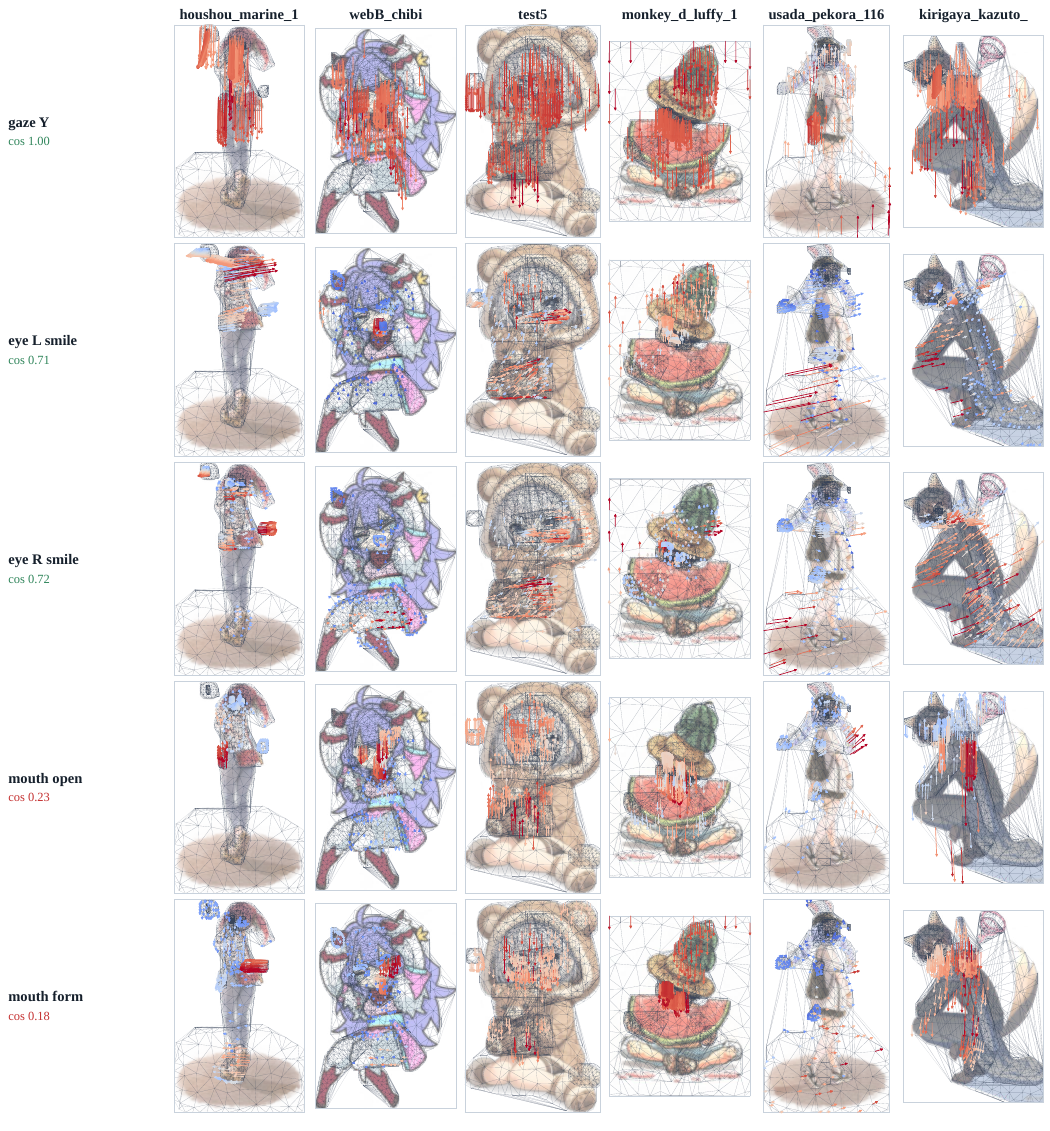}
  \caption{Sheet 3: gaze, eye smile and mouth. Format as Fig.~\ref{fig:atlas1}. Gaze is the
  highest-scoring parameter in the whole vocabulary ($0.997$ and $0.999$) and the atlas shows why the
  task is easy: the field is confined to two small iris layers and is close to a rigid translation.
  Mouth form, by contrast, is one of the failures, and its field is nearly empty.}
  \Description{Gaze, eye-smile and mouth parameters as displacement fields.}
  \label{fig:atlas3}
\end{figure*}

\begin{figure*}[p]
  \centering
  \includegraphics[width=\linewidth]{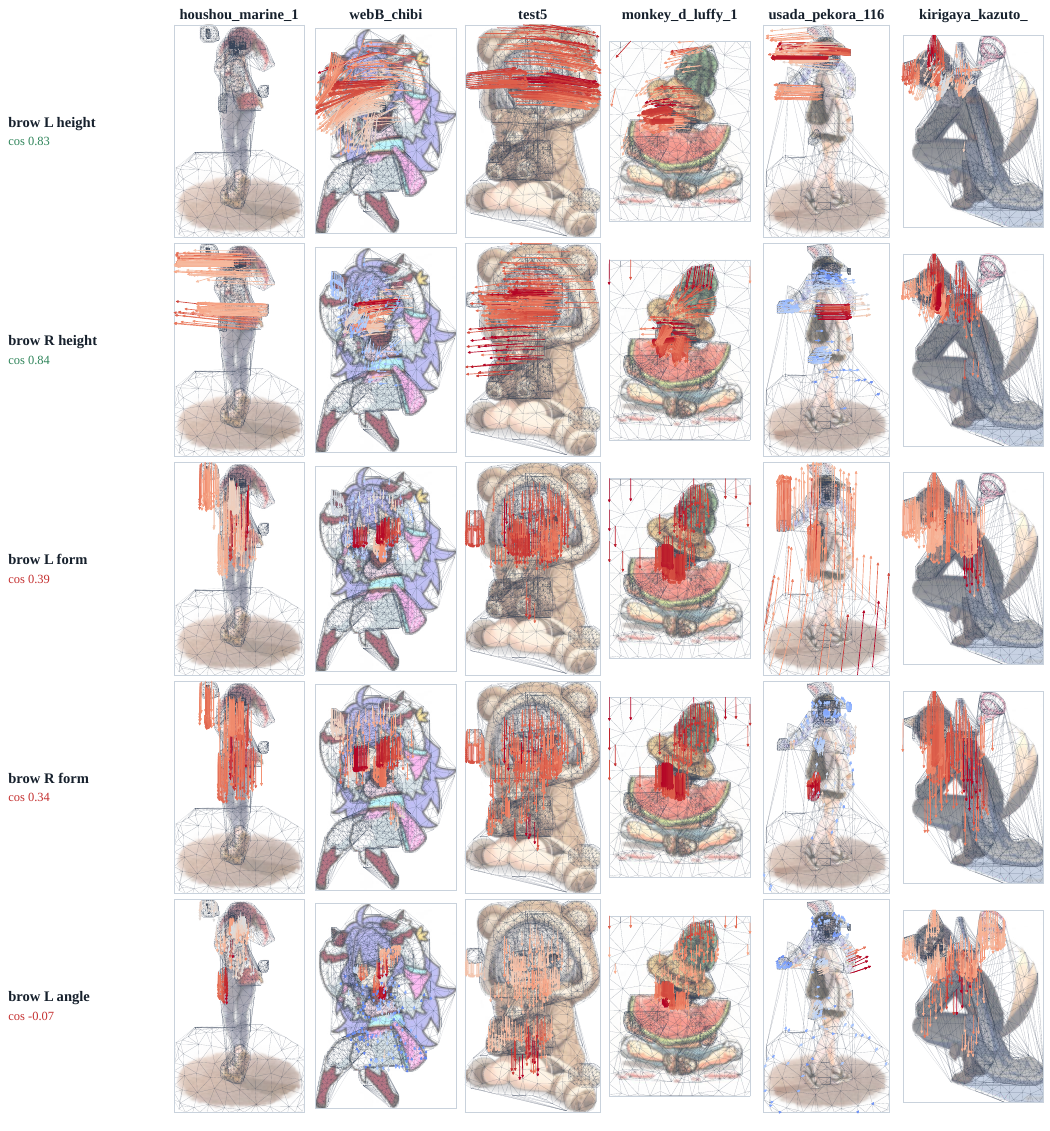}
  \caption{Sheet 4: brow controls, which are where the vocabulary extension fails. Format as
  Fig.~\ref{fig:atlas1}. Brow height is learned ($0.83$ and $0.84$); brow form and brow angle are not
  ($0.39$, $0.34$, and negative for the angles). The fields for the failed parameters are visibly
  unstructured, which is what a parameter trained on $17$ to $48$ moving-layer samples looks like.}
  \Description{Brow parameters as displacement fields, including the failed ones.}
  \label{fig:atlas4}
\end{figure*}

\begin{figure*}[t]
  \centering
  \includegraphics[width=\linewidth]{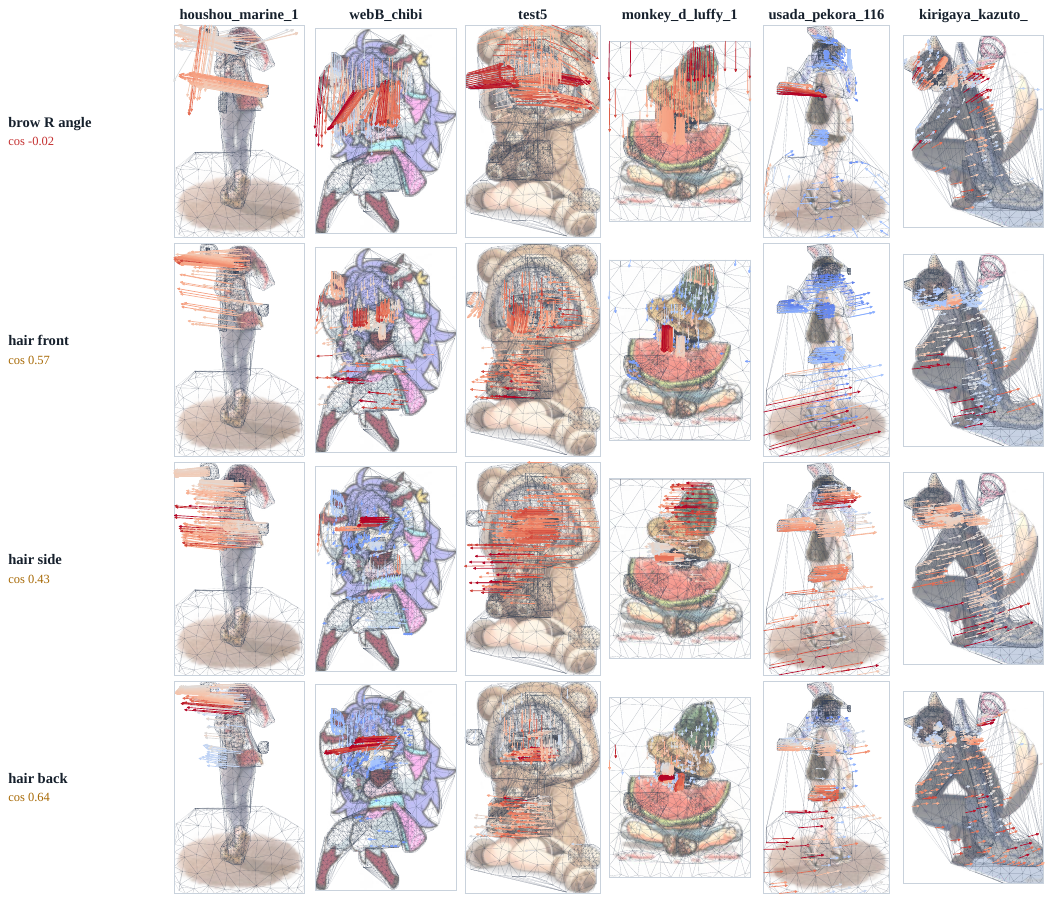}
  \caption{Sheet 5: the remaining brow angle and the three hair-sway groups. Format as
  Fig.~\ref{fig:atlas1}. Front and back hair sway are marginal ($0.57$ and $0.64$) and side hair is
  below the usable threshold ($0.43$); the atlas shows the fields are at least confined to hair layers,
  so the failure is one of amplitude and direction within the right support rather than of localisation.}
  \Description{Brow angle and hair-sway parameters as displacement fields.}
  \label{fig:atlas5}
\end{figure*}

\section{In-the-Wild Qualitative Gallery}
\label{app:itw_gallery}

The main text shows one sheet of in-the-wild rigs (Fig.~\ref{fig:itw_poses}). This appendix shows
\emph{every} in-the-wild character we processed, because the honest way to present a generative system
is to show its whole output on a fixed input set rather than a selection. The paper's appendix has no
page limit, so we use it.

Characters are ordered by \emph{measured} motion magnitude, defined as mean predicted displacement at
the extreme keypose divided by character span and printed beside each row. That ordering is worth
reading as a result in itself: it spans $0.037$ down to $0.004$, roughly a factor of nine, on inputs
that a human would describe as equally animatable. The low end is where amplitude compression
(\S\ref{sec:conclusion}) is most visible, and it correlates with characters whose layers are large and
flat, where a small displacement of many vertices is the conditional-mean answer.

\begin{figure*}[p]
  \centering
  \includegraphics[width=\linewidth]{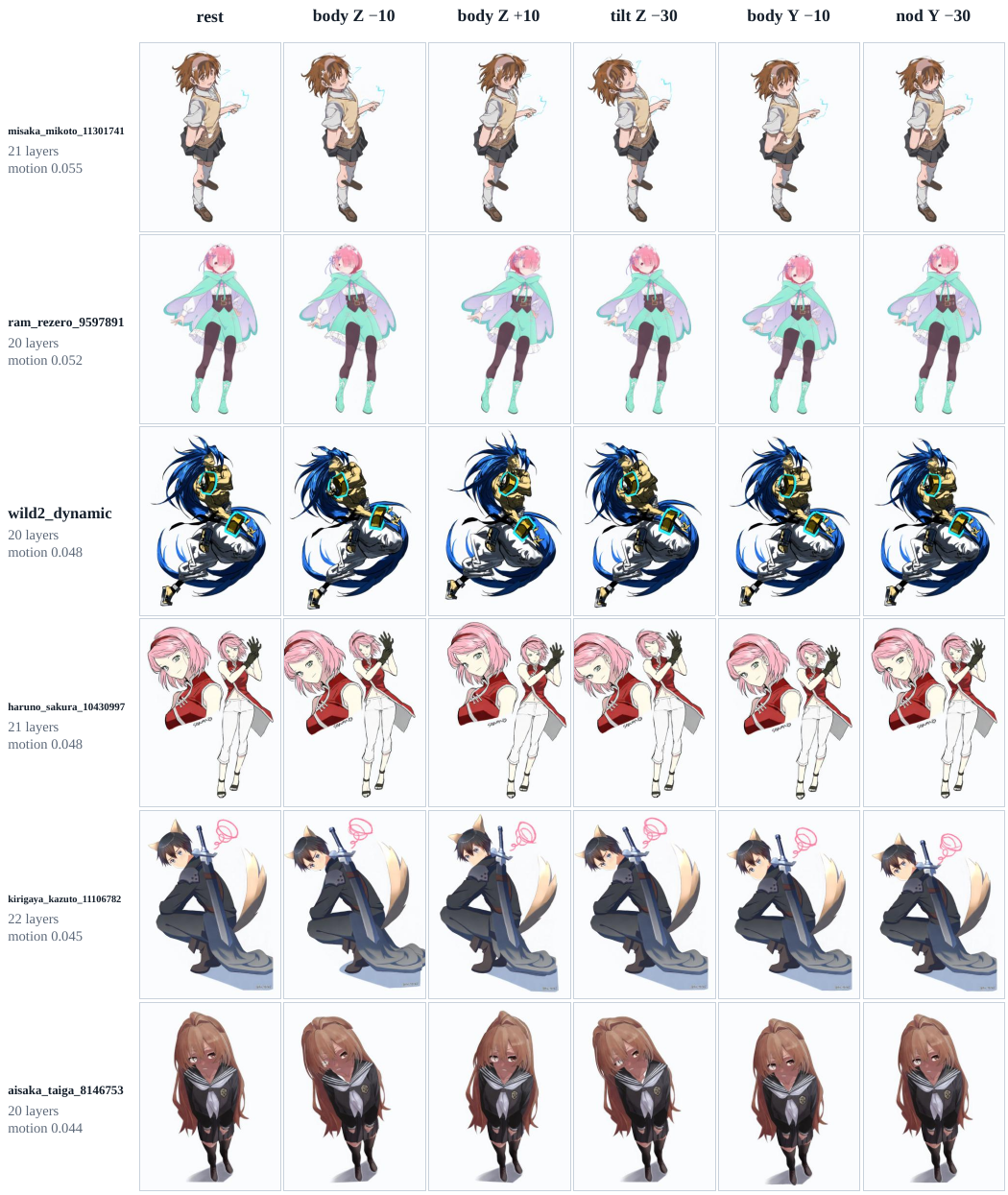}
  \caption{In-the-wild rigs, raw output, sheet~2. Format as Fig.~\ref{fig:itw_poses}.}
  \Description{Five more in-the-wild characters at six parameter values.}
  \label{fig:itw_poses_2}
\end{figure*}

\begin{figure*}[p]
  \centering
  \includegraphics[width=\linewidth]{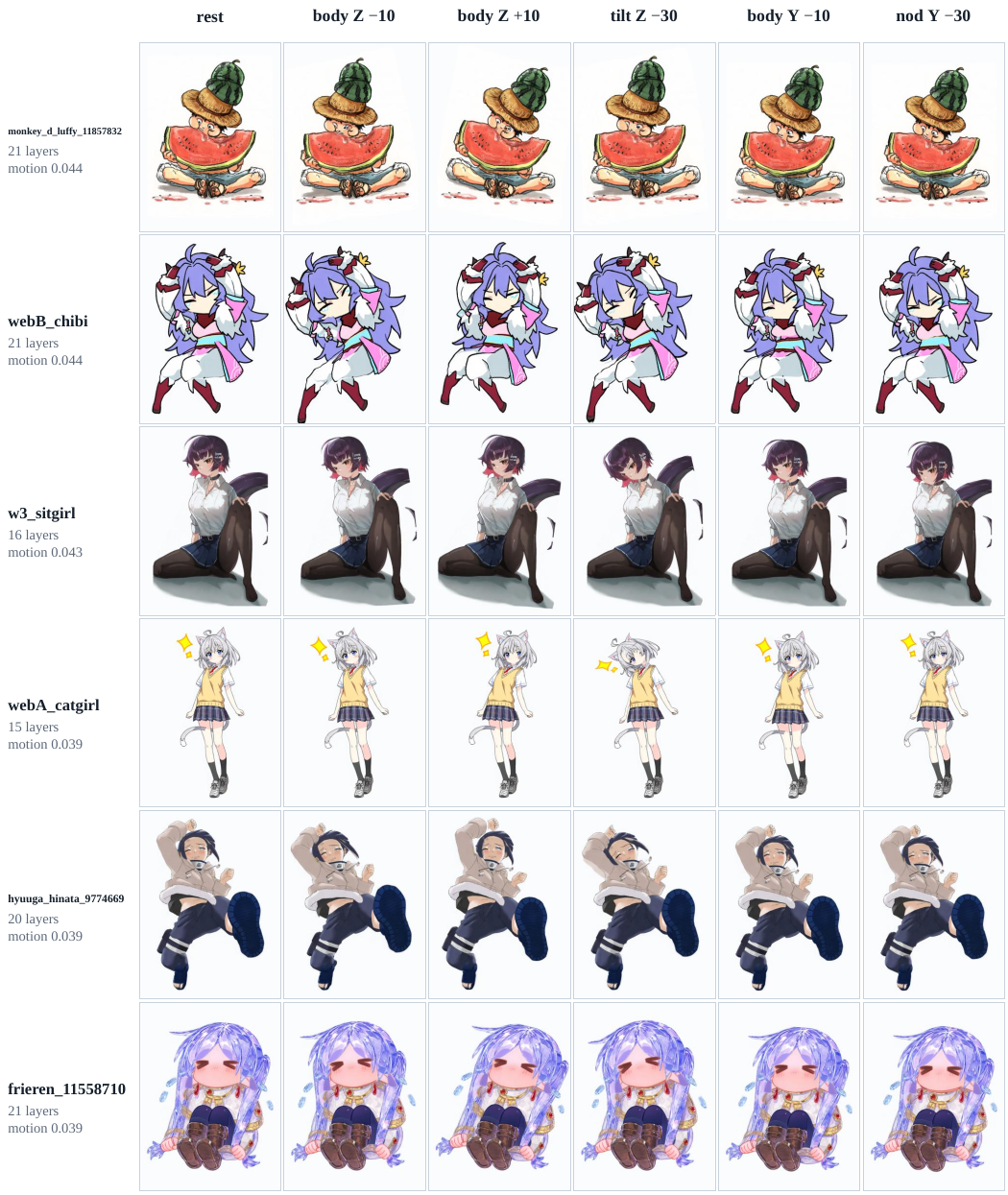}
  \caption{In-the-wild rigs, raw output, sheet~3. Format as Fig.~\ref{fig:itw_poses}.}
  \Description{Five more in-the-wild characters at six parameter values.}
  \label{fig:itw_poses_3}
\end{figure*}

\begin{figure*}[p]
  \centering
  \includegraphics[width=\linewidth]{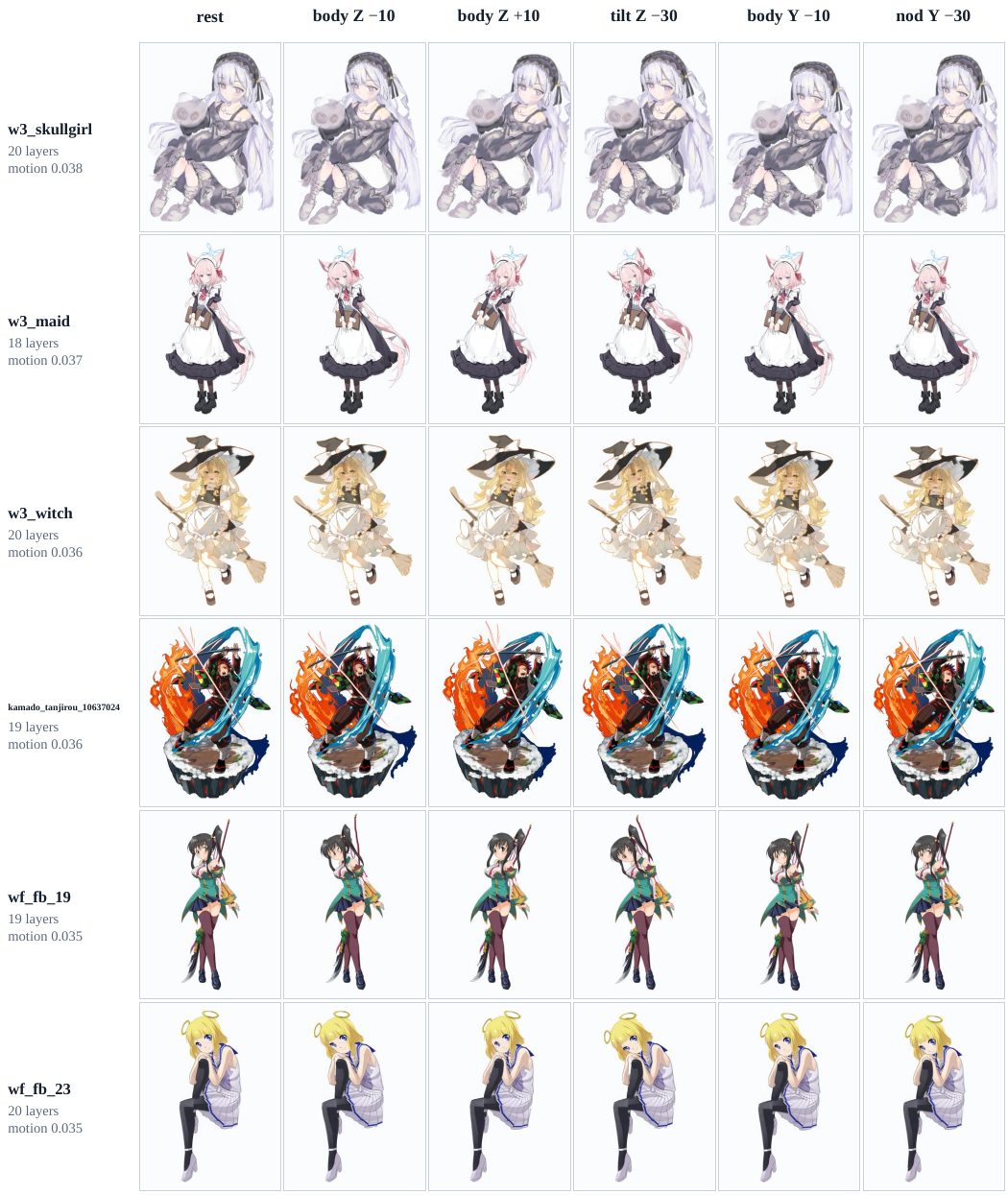}
  \caption{In-the-wild rigs, raw output, sheet~4. Format as Fig.~\ref{fig:itw_poses}.}
  \Description{Six more in-the-wild characters at six parameter values.}
  \label{fig:itw_poses_4}
\end{figure*}

\begin{figure*}[p]
  \centering
  \includegraphics[width=\linewidth]{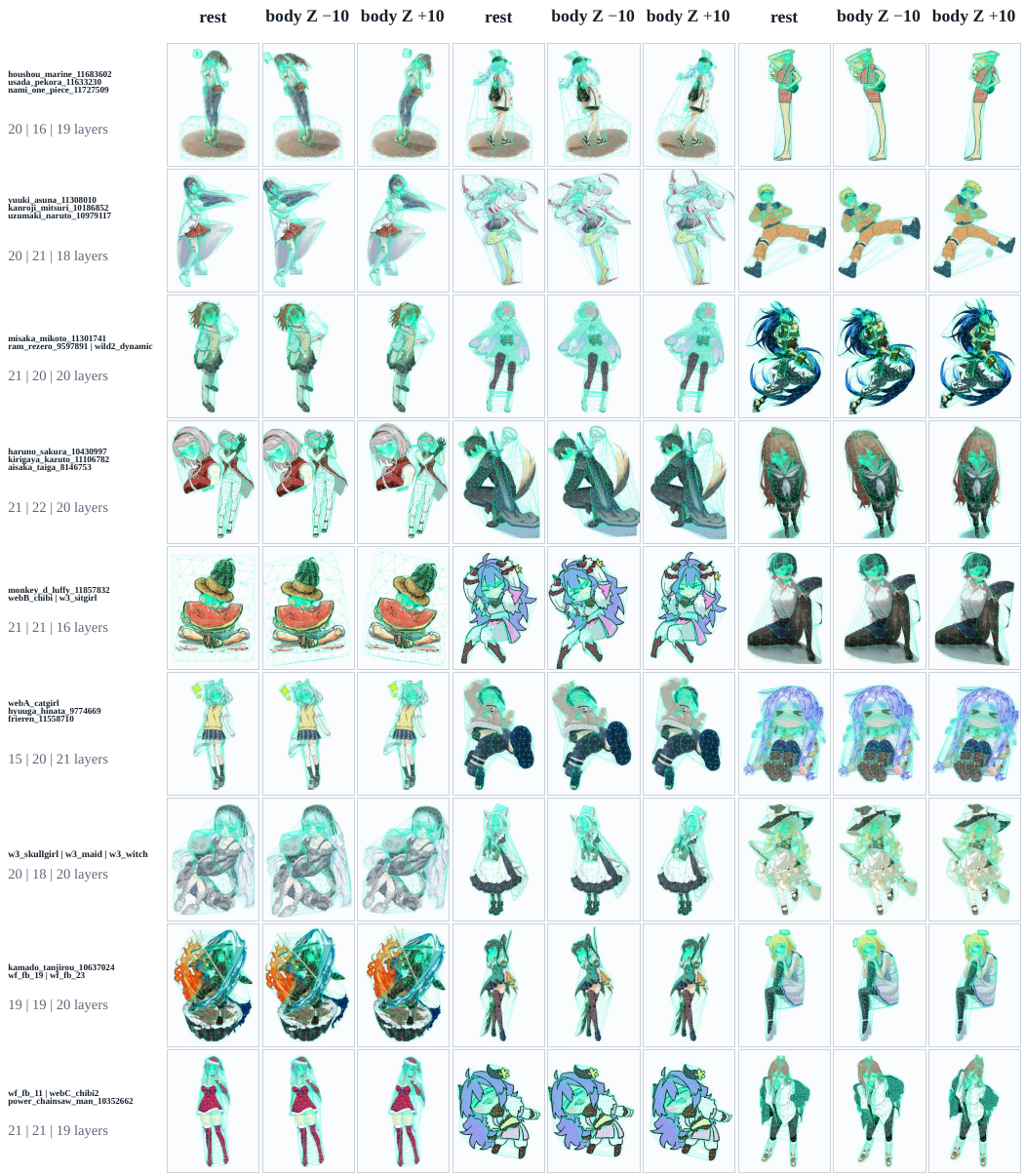}
  \caption{\textbf{The generated triangulation drawn over the artwork}, at rest and at the two poses that
  displace these characters most, three characters per row. Every visible pixel of every layer must lie
  inside a triangle or the layer tears at its silhouette during deformation. Sheet~2 is
  Fig.~\ref{fig:itw_mesh_2}.}
  \Description{In-the-wild characters with the predicted triangle mesh overlaid, at rest and animated.}
  \label{fig:itw_mesh_1}
\end{figure*}

Figures~\ref{fig:itw_poses_2}--\ref{fig:itw_poses_4} continue the pose sheets.
Fig.~\ref{fig:itw_mesh_1} shows the same rigs with the generated triangulation
drawn over the artwork, which is the view a rigger inspects: every visible pixel of every layer must
lie inside a triangle, or the layer tears at its silhouette during deformation. Over the $355$ layers of these $18$
rigs, coverage of perceptible pixels ($\alpha\!\geq\!8$) averages $0.9996$ with a median of $1.000$
and a minimum of $0.9863$; exactly one layer falls below $0.99$. The residual is anti-aliased rim
pixels at the very edge of the dilated alpha mask.

\begin{figure*}[p]
  \centering
  \includegraphics[width=\linewidth]{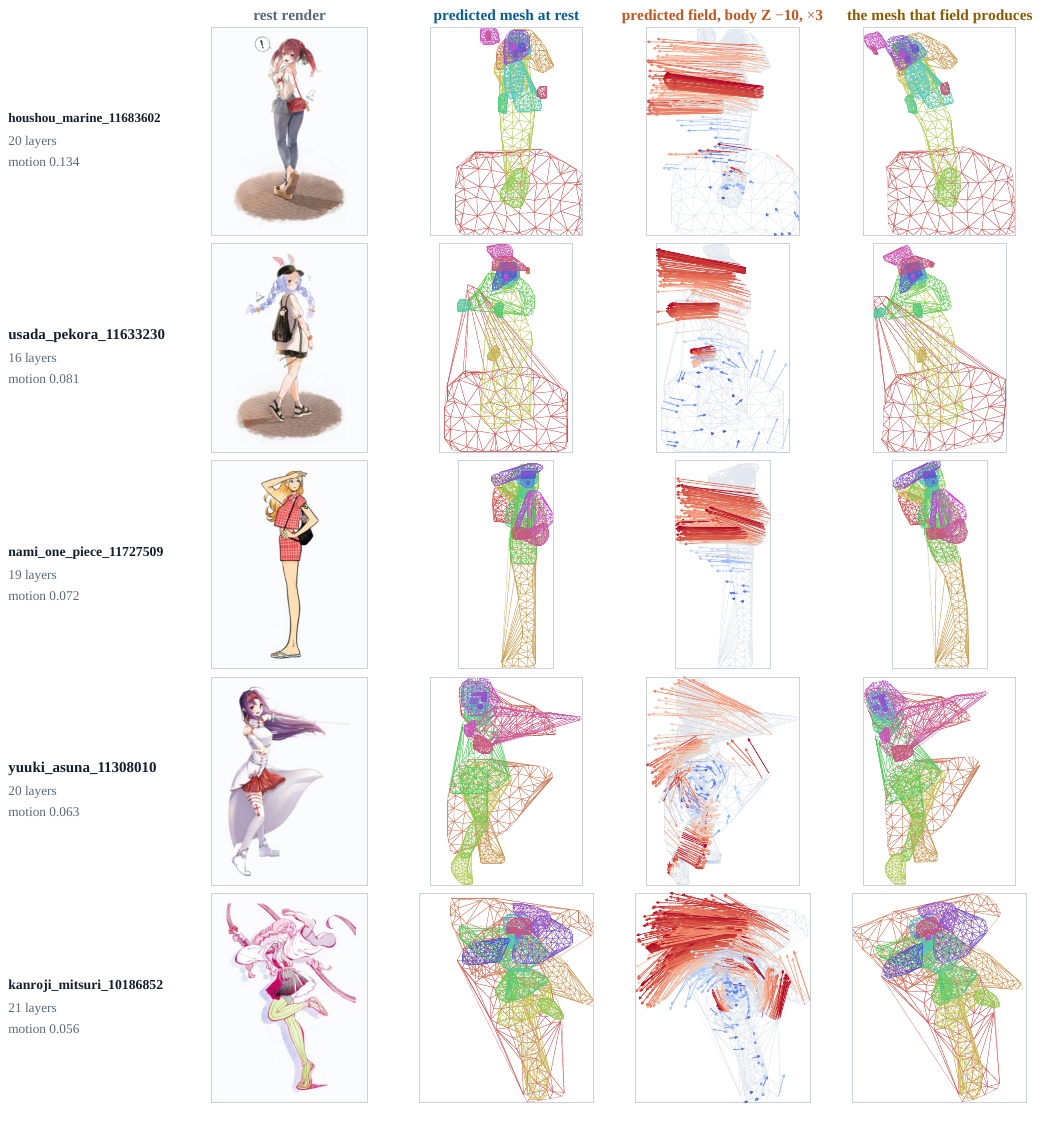}
  \caption{\textbf{What Stage~2 actually predicts, drawn from the rig rather than rendered.} Per row:
  the rig at rest; the predicted triangulation at rest with one hue per layer; the predicted
  displacement field for $\texttt{ParamAngleX}\!=\!+30$ as per-vertex arrows, exaggerated $3\times$ and
  coloured blue (small) to red (large); and the deformed triangulation. The field is continuous across
  layer boundaries although the layers share no connectivity, and its magnitude is organised by depth
  although depth is never an input.}
  \Description{In-the-wild characters with their predicted meshes and displacement-field arrows.}
  \label{fig:itw_field}
\end{figure*}

Fig.~\ref{fig:itw_field} then drops the renderer entirely and draws the prediction itself: the
triangulation at rest, the predicted per-vertex displacement field as arrows, and the deformed
triangulation, all read directly out of the released rig files. The arrows are the literal output of
Eq.~\ref{eq:stage2_recompose}, exaggerated $3\times$ for legibility and coloured by magnitude. Three
properties are visible there that a rendered frame cannot show. The field is \emph{spatially coherent
across layer boundaries} even though those layers share no connectivity, which is the joint attention
doing its job. Its magnitude is \emph{organised by depth} without depth ever being an input:
head-group layers carry the long arrows, torso layers intermediate, parts resting on the ground almost
nothing. And the triangulation is visibly \emph{denser along contours than in flat interiors}, which
is the content-conforming construction of \S\ref{sec:stage2} and the reason its vertex budget is
$26\%$ below a grid's at equal coverage.

\begin{figure*}[p]
  \centering
  \includegraphics[width=\linewidth]{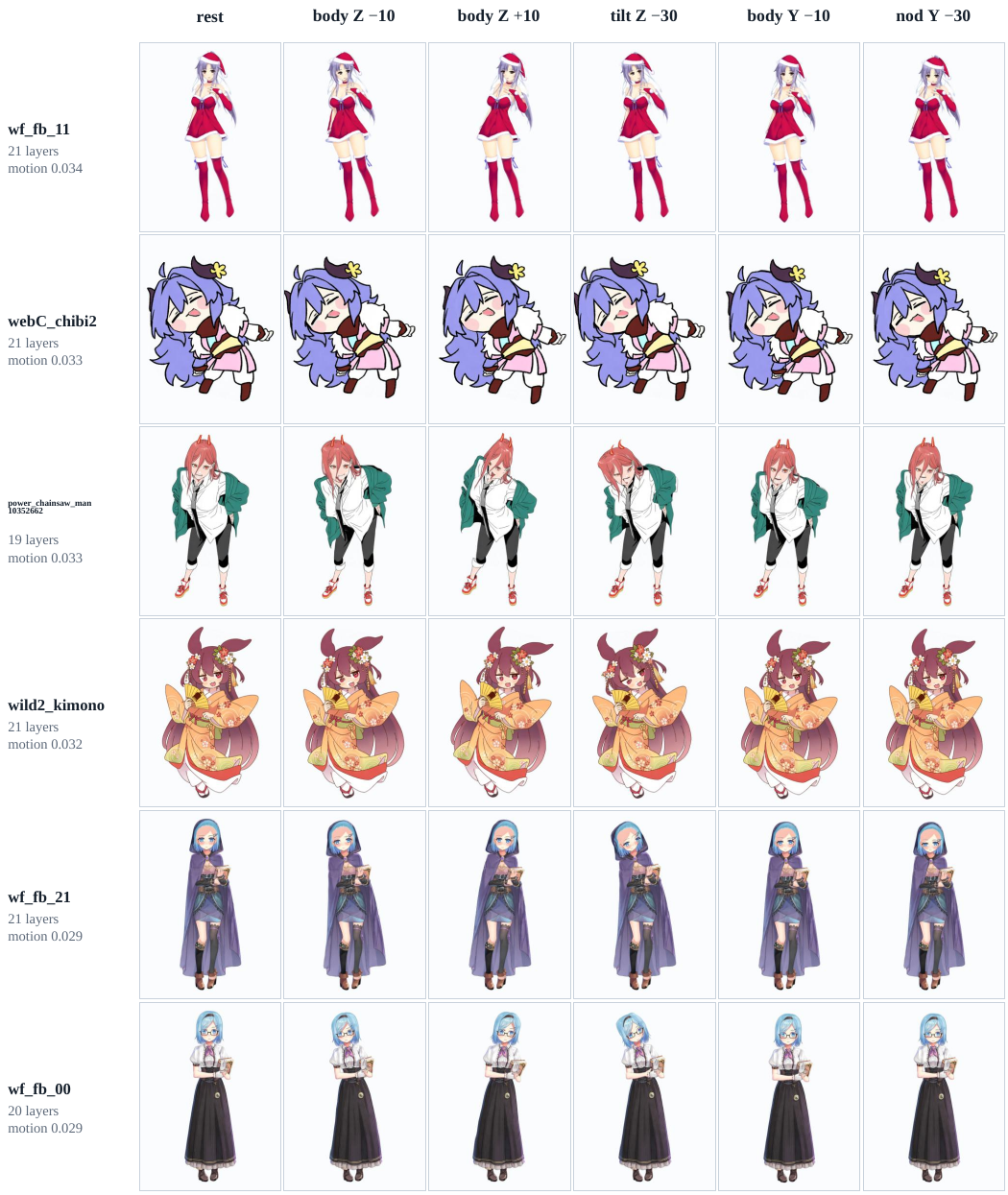}
  \caption{In-the-wild rigs, raw output, sheet~5. Format as Fig.~\ref{fig:itw_poses}.}
  \Description{Six more in-the-wild characters at six parameter values.}
  \label{fig:itw_poses_5}
\end{figure*}

\begin{figure*}[p]
  \centering
  \includegraphics[width=\linewidth]{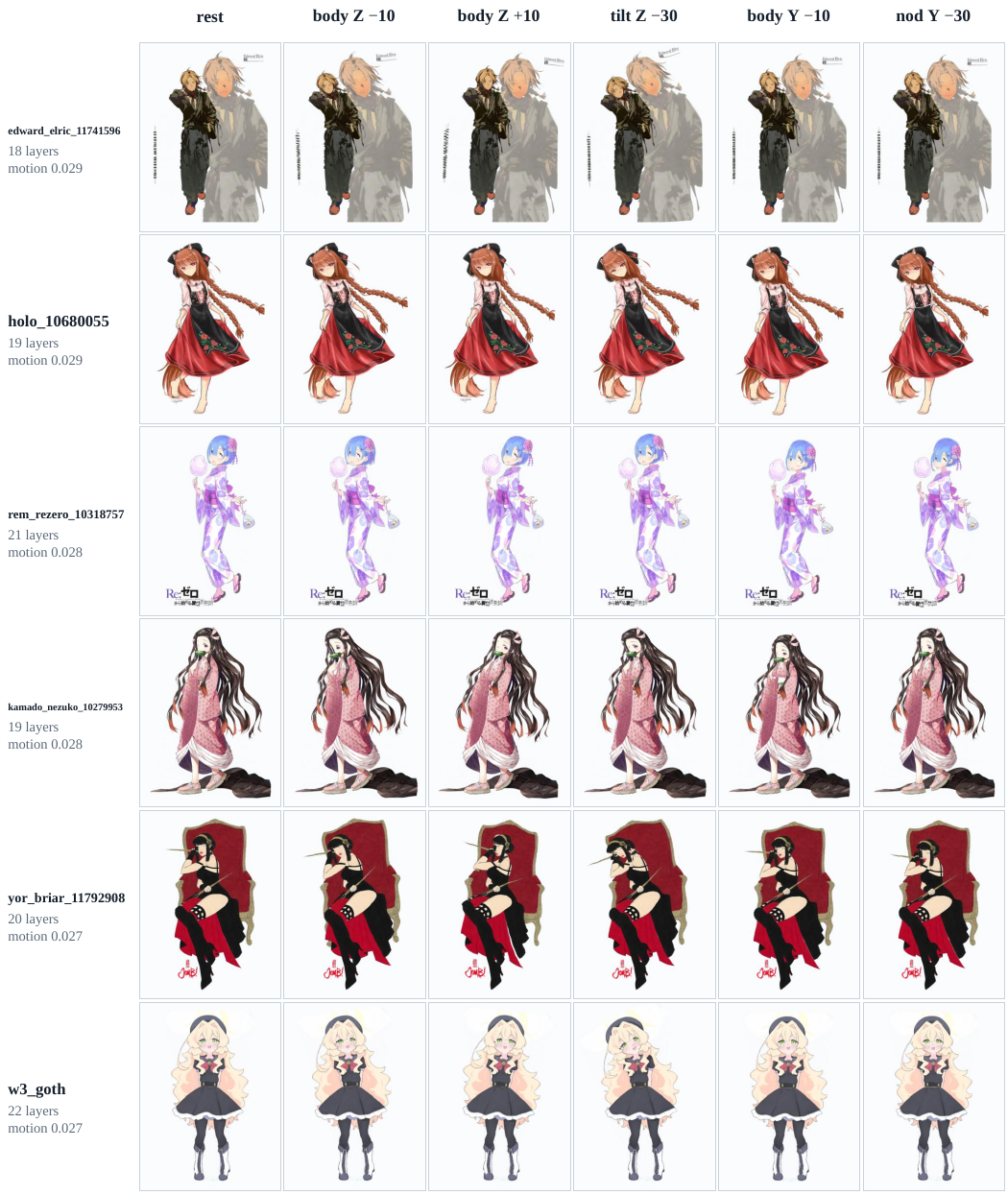}
  \caption{In-the-wild rigs, raw output, sheet~6. Format as Fig.~\ref{fig:itw_poses}.}
  \Description{Six more in-the-wild characters at six parameter values.}
  \label{fig:itw_poses_6}
\end{figure*}

\begin{figure*}[p]
  \centering
  \includegraphics[width=\linewidth]{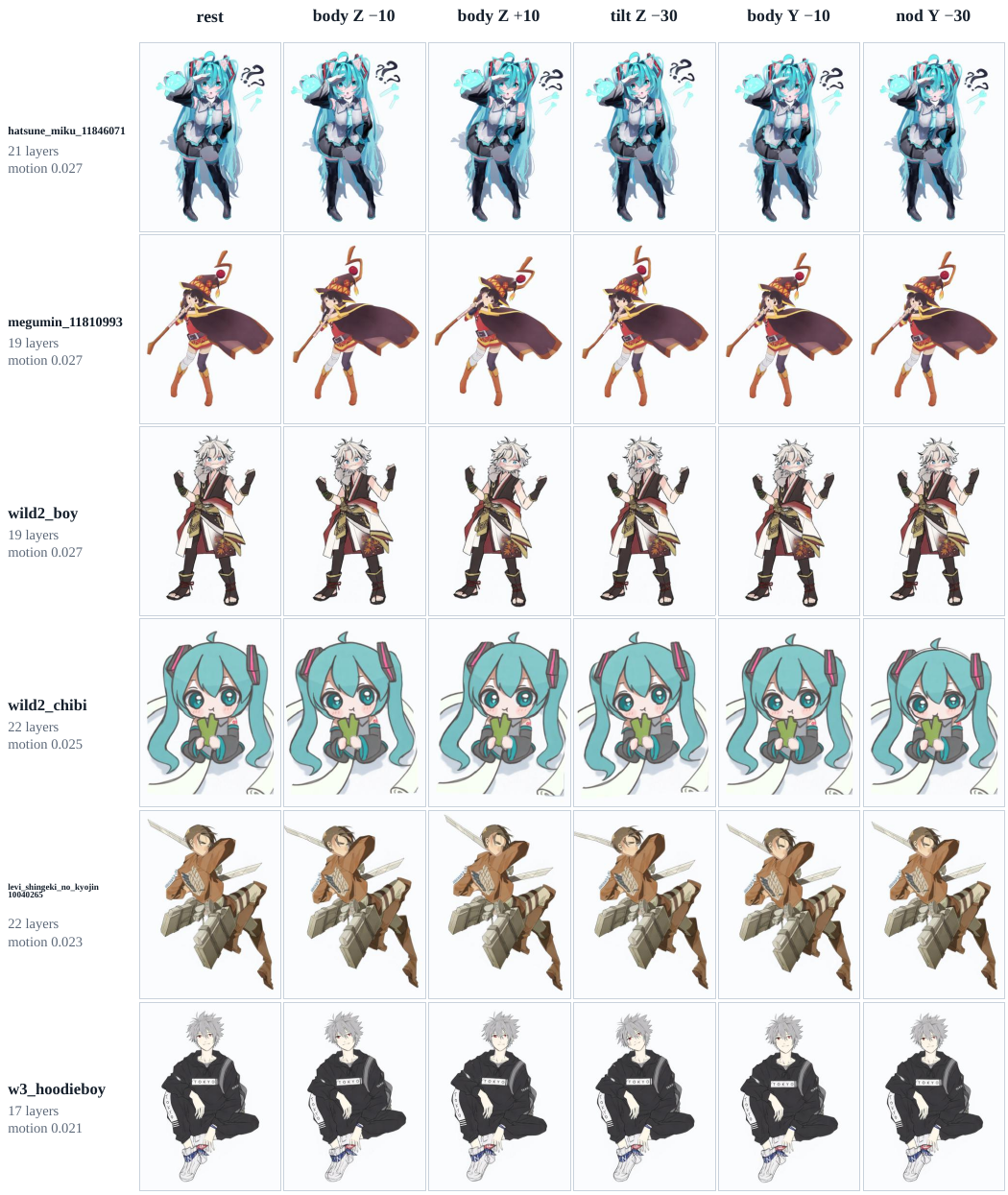}
  \caption{In-the-wild rigs, raw output, sheet~7. Format as Fig.~\ref{fig:itw_poses}.}
  \Description{Six more in-the-wild characters at six parameter values.}
  \label{fig:itw_poses_7}
\end{figure*}

\begin{figure*}[p]
  \centering
  \includegraphics[width=\linewidth]{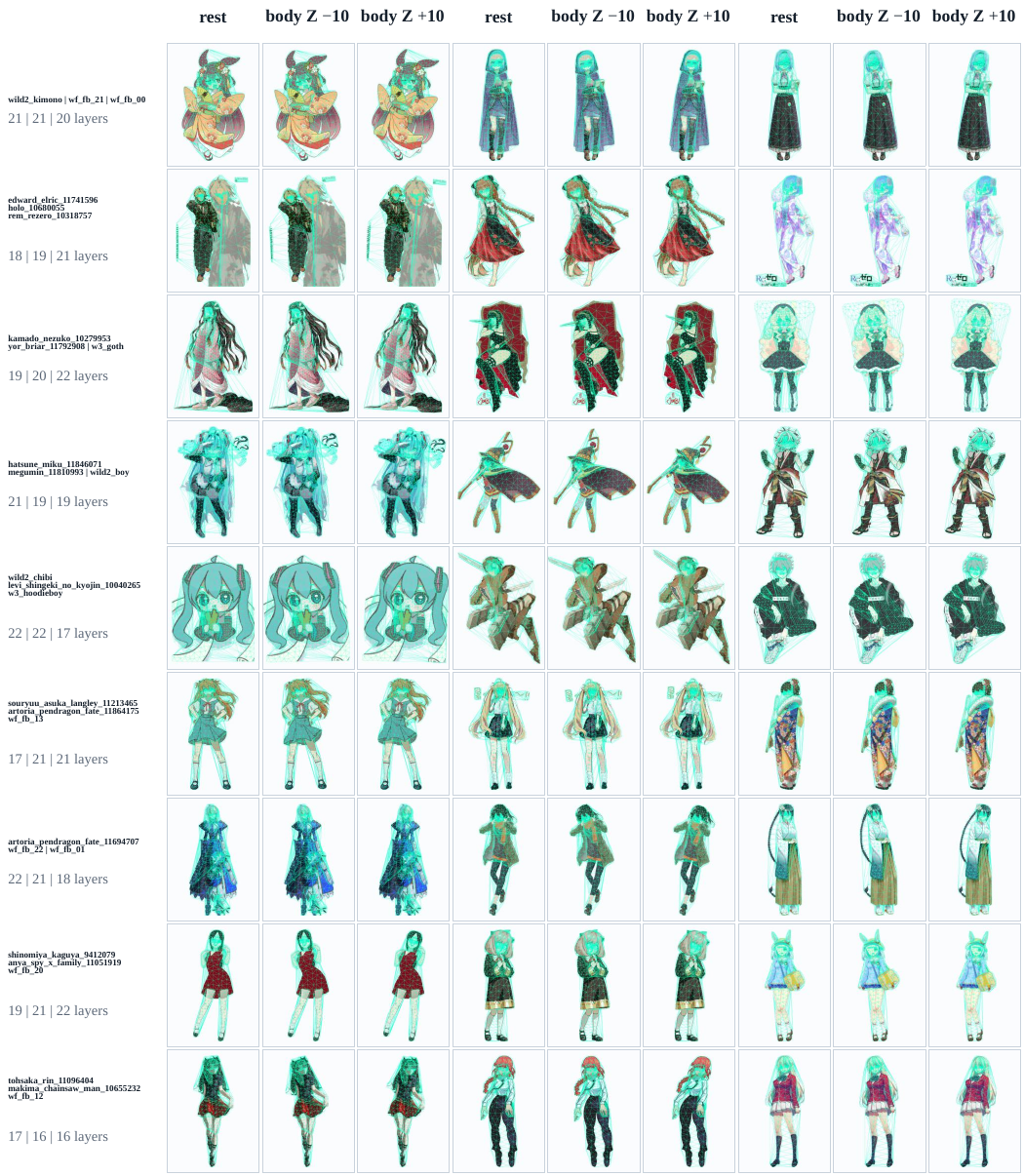}
  \caption{The generated triangulation drawn over the artwork, sheet~2. Format as
  Fig.~\ref{fig:itw_mesh_1}.}
  \Description{More in-the-wild characters with the predicted triangle mesh overlaid.}
  \label{fig:itw_mesh_2}
\end{figure*}

\section{Assets That Predate the Method Change}
\label{app:provenance}

This paper went through a change of Stage-2 method: an autoregressive discretised-token design
(Appendix~\ref{app:mesh_tokenization}) was replaced by the single-pass joint continuous regressor of
\S\ref{sec:stage2}. Most figures and tables were produced or re-measured with the current model, and
the ones that were not say so in their own caption rather than in a separate index. Concretely, the
token-traversal visualisations (Figs.~\ref{fig:combined_elf_hair}--\ref{fig:combined_gumi},
Tabs.~\ref{tab:mesh_tok_stats} and~\ref{tab:cross_char_demos}) exist precisely to document the
superseded design; the layer-to-mesh deployment study (Fig.~\ref{fig:layer_to_mesh},
Tab.~\ref{tab:layer_to_mesh}) was run under that design's mesh-prefix protocol and is retained only
for the \emph{relative ordering} of mesh algorithms, which is method-independent; and the failure
gallery (Fig.~\ref{fig:failure_gallery}) predates the change but isolates failure \emph{categories}
that persist. Everything concerning Stage~1 (decomposition, dataset statistics, mesh construction)
never involved the animation model at all.

Producing this audit caught three assets that were stale without saying so, and all three are now
fixed rather than annotated. The runtime discussion carried an unmeasured placeholder and is now
measured (Tab.~\ref{tab:runtime}). The image-to-video table silently reported a superseded-model row
as if it were current, and now separates the two protocols explicitly
(Tab.~\ref{tab:stage2_video}). And the claim that a quad grid wastes ``$40$ to $60\%$ of its
vertices on transparent pixels'', which we had used to justify the mesh design, turned out to be
false when measured ($0.376$ against $0.368$ on-content vertex fraction, indistinguishable); it has
been replaced by the argument the measurement does support, namely $26\%$ fewer tokens at equal
coverage. Two items remain genuinely outstanding and we prefer to name them: the two overview figures
need redrawing to match the current Stage~2, and the four image-to-video baselines need re-running
under the current rendering protocol.

\section{Stage 2 Runtime and Memory}
\label{app:runtime}

Measured with a script that reproduces the reference implementation phase by phase (identical
content-conforming triangulation at $110$ target vertices per layer, identical $600$\,px canvas,
identical parameter and keypose spec), with \texttt{cuda.\allowbreak synchronize()} bracketing every
timed phase and peak memory read from \texttt{max\_\allowbreak memory\_\allowbreak allocated()} after resetting
statistics at the start of the phase. One warm-up repetition is discarded and nine timed
repetitions are kept; the median is reported with $[\min,\max]$ in Tab.~\ref{tab:runtime}. The
card was an A800-80GB \emph{shared with other tenants} who held $58.5$\,GB at the time, which is
why the spread on the GPU phases is wide; the minima are the better estimate of an uncontended
run. Stage~1 is a separate $20$\,B diffusion model and is deliberately not measured here, so no
Stage-1 timing is reported or estimated.

\begin{table}[t]
  \centering
  \caption{\textbf{Stage 2 per-character wall clock}, seconds, median of nine repetitions with $[\min,\max]$. ``Stage-1 output'' is a real in-the-wild decomposition rather than a benchmark character. A complete rig is $13$ forward passes, each covering all layers of the character simultaneously.}
  \label{tab:runtime}
  \scriptsize
  \setlength{\tabcolsep}{2.5pt}
  \renewcommand{\arraystretch}{1.15}
  \begin{tabular*}{\linewidth}{@{\extracolsep{\fill}}lccc@{}}
    \toprule
    & small & large & Stage-1 output \\
    & $10$ lay., $T\!=\!1052$ & $63$ lay., $T\!=\!5476$ & $19$ lay., $T\!=\!1873$ \\
    \midrule
    PNG decode $+$ crop (CPU)        & $0.155$ & $0.528$ & $0.577$ \\
    DINOv2 features (GPU)            & $0.115$ & $0.614$ & $0.183$ \\
    mesh construction (CPU)          & $0.016$ & $0.076$ & $0.027$ \\
    joint forwards, all $13$ (GPU)   & $0.165$ & $1.061$ & $0.270$ \\
    rig JSON $+$ texture export      & $0.256$ & $0.562$ & $0.325$ \\
    \midrule
    \textbf{total per character}     & $\mathbf{0.71}$ & $\mathbf{2.84}$ & $\mathbf{1.39}$ \\
    \midrule
    peak activation memory           & $189$\,MiB & $1982$\,MiB & -- \\
    triangles produced               & $1513$ & $7930$ & $2637$ \\
    rig JSON size                    & $0.94$\,MB & $4.94$\,MB & $1.70$\,MB \\
    \bottomrule
  \end{tabular*}
\end{table}

\paragraph*{What this means for deployment.}
Resident weights are $103.9$\,MiB, of which $84.5$ is the frozen DINOv2 encoder and only $19.4$
the model we train; the checkpoint on disk is $19.0$\,MiB. Loading costs $0.42$\,s for the joint
model and $2.85$\,s for DINOv2, once per process rather than per character. The activation cost is
the only quantity that grows steeply: $5.2\times$ more tokens costs $24.7\times$ more activation
memory, exactly the $O(T^2)$ of dense attention over vertices. At the largest character we have
($125$ layers) this is still under $4$\,GB, but a character an order of magnitude denser would
need attention windowing, most naturally over layer blocks, which is a change to the mechanism
this paper argues for and therefore not one we make casually. Finally, mesh construction is
$2.7\%$ of the large character's time, which is worth stating because the triangulation is the
part of Stage~2 that looks expensive and is not: the cost is attention over vertices, and the
mesh's job is to keep the vertex count low, which is exactly what \S\ref{sec:results_stage2}
measures it doing ($83$ vertices per layer against a grid's $112$ at equal coverage).

\section{What We Tried That Did Not Work}
\label{app:negative}

The conditioning signals below are all plausible, several were suggested by reviewers of an
earlier version, and none of them beat the plain joint model. We report them because a reader
choosing what to add next is better served by knowing which additions we already paid for.
Every row of Tab.~\ref{tab:negative} uses the identical $46$-character clean benchmark, pack,
objective and schedule, and differs from the baseline only in the stated conditioning; all
numbers are true generation. The essential context is the \textbf{seed spread}: two runs of the
baseline that differ only in random seed give $0.7289$ and $0.7533$, a range of $0.024$, so a
variant must move the mean by more than that to mean anything. None of them does.

\paragraph*{Explicit per-vertex draw order (C1, C3).}
Reviewers observed that our layer ordering is a taxonomy order rather than a true character
z-order, and suggested conditioning on depth explicitly. We did: each vertex additionally
receives its layer's normalised draw-order position, front to back, Fourier-encoded and added to
the token embedding, which is a strictly larger hypothesis class than the layer-identity
embedding alone. It scores $0.7148$, below both baseline seeds. Our reading is that layer
identity plus a shared canvas already carries the ordering information the displacement field
needs: layers that occlude each other are adjacent in the stack \emph{and} overlapping in the
canvas, so the ordering is recoverable from what the model already sees, and adding it as a
separate channel only spends capacity. This does not resolve the reviewers' deeper point, which
is that a wrong ordering \emph{in the layer stack itself} is a Stage-1 error we cannot fix in
Stage-2; that remains open and is stated as a limitation.

\paragraph*{A 2.5D parallax prior.}
A head turning is not a 2D slide: near and far parts of the face should move differently, which
a per-vertex 2D displacement field must learn implicitly. We added a zero-initialised residual
that predicts a per-vertex depth $z$ and a per-pose parallax gain $g$, contributing $z\cdot g$
to the pre-tanh direction, so the variant reduces exactly to the baseline at initialisation and
can only help if the data supports it. Two seeds give $0.7225$ and $0.7170$, inside the baseline
seed range, and extending to $40$ epochs gives $0.7223$: no gain on the mean. One nuance is
worth recording rather than burying. Parallax combined with draw-order conditioning at $40$
epochs has the \emph{best median} of any variant we trained ($0.8255$ against the baseline
seeds' $0.8024$ and $0.7993$), the \emph{most} characters above $0.80$ ($25/46$), and by far the
best amplitude calibration (magnitude ratio $1.054$ against $1.238$ and $1.163$), while its mean
is lower ($0.7153$). That pattern is a left-tail effect: the variant improves the typical
character and the amplitude, and loses on a few characters badly enough to drag the mean. We do
not ship it, because we cannot show the median gain exceeds noise with two seeds, but a
2.5D prior remains the most promising of the negative results and the amplitude effect is the
one signal here that points at our known amplitude-compression failure.

\paragraph*{Per-vertex image features instead of one pooled vector per layer.}
Our model gives every vertex of a layer the \emph{same} pooled DINOv2 vector. The obvious
refinement is to sample the $16\!\times\!16$ patch-token grid at each vertex's own position, so a
vertex on a sleeve edge sees the sleeve edge. It is clearly worse: $0.6918$, and the amplitude
degrades badly (magnitude ratio $1.563$ against $1.238$). We interpret this as the conditioning
becoming too local. A layer's displacement under a head turn is a property of \emph{what the
layer is} (a fringe, an iris, a collar) far more than of what any single vertex looks like, and
per-vertex sampling lets the model latch onto local texture that does not predict global motion,
while the pooled vector forces a layer-level summary. This is also the reason we do not use
cross-attention into image patches, and it is one of the two things \S\ref{sec:stage2} states the
architecture deliberately does not do.

\paragraph*{Width and depth at fixed data.}
For completeness, two intermediate capacity points inside the small-model regime behave like the
larger ones in ablation~B: $8$ blocks at $d\,256$ ($6.7$\,M) gives $0.7325$ and $d\,384$ at
$6$ blocks ($11.2$\,M) gives $0.7168$, both inside or below the baseline seed range. The
$d\,384$ variant does reach $25/46$ characters above $0.80$ with a median of $0.8137$, the same
median-versus-mean pattern as the parallax variant.

\begin{table}[t]
  \centering
  \caption{\textbf{Negative and inconclusive results.} All on the clean $46$-character benchmark, same pack, objective and $20$-epoch schedule unless noted, true generation. The two baseline seeds bracket the noise floor: a mean difference smaller than $0.024$ is not evidence. Bold marks entries that beat \emph{both} baseline seeds on that column.}
  \label{tab:negative}
  \scriptsize
  \setlength{\tabcolsep}{3pt}
  \renewcommand{\arraystretch}{1.1}
  \begin{tabular*}{\linewidth}{@{\extracolsep{\fill}}lcccc@{}}
    \toprule
    & \multicolumn{2}{c}{dir-cos $\uparrow$} & mag & chars \\
    \cmidrule(lr){2-3}\cmidrule(lr){4-4}\cmidrule(lr){5-5}
    Variant & mean & med. & med.\,$\to\!1$ & $\geq\!0.8$ \\
    \midrule
    baseline, seed 1                  & $0.7289$ & $0.8024$ & $1.238$ & $23/46$ \\
    baseline, seed 2                  & $0.7533$ & $0.7993$ & $1.163$ & $23/46$ \\
    \midrule
    $+$ draw-order conditioning       & $0.7148$ & $0.7883$ & $1.154$ & $21/46$ \\
    $+$ parallax, seed 1              & $0.7225$ & $0.8048$ & $1.095$ & $\mathbf{24}/46$ \\
    $+$ parallax, seed 2              & $0.7170$ & $0.7828$ & $1.184$ & $22/46$ \\
    $+$ parallax, $40$ ep             & $0.7223$ & $0.8004$ & $1.193$ & $23/46$ \\
    $+$ parallax $+$ order, $40$ ep   & $0.7153$ & $\mathbf{0.8255}$ & $\mathbf{1.054}$ & $\mathbf{25}/46$ \\
    $+$ per-vertex image features     & $0.6918$ & $0.7959$ & $1.563$ & $23/46$ \\
    \midrule
    $8$ blocks, $d\,256$ ($6.7$\,M)   & $0.7325$ & $0.7939$ & $1.139$ & $23/46$ \\
    $6$ blocks, $d\,384$ ($11.2$\,M)  & $0.7168$ & $\mathbf{0.8137}$ & $1.072$ & $\mathbf{25}/46$ \\
    \bottomrule
  \end{tabular*}
\end{table}

\paragraph*{Balanced sampling does not rescue the long tail.}
The obvious response to a long-tailed parameter distribution is to rebalance the sampler, and we tried
it. Counting \emph{layer records} rather than poses (a rare parameter appears in many poses but
contributes two moving layers each, while head rotation contributes forty, so layer count is what sets
the gradient share), we physically repeat each pose $\lceil(\max/\mathrm{count})^{0.5}\rceil$ times,
capped at $24$, which grows the epoch from $26{,}133$ to $92{,}068$ poses and repeats the rarest
parameter twelve times. It does not help, and it very slightly hurts on both halves of the vocabulary:
the eight warm-started parameters fall from $0.6983$ to $0.6931$ and the sixteen new ones from $0.5588$
to $0.5562$, with validation cosine indistinguishable ($0.6957$ against $0.6958$). We read this as
informative rather than merely negative. Repetition raises a rare parameter's share of the gradient but
adds no information about it, so if the failure were a matter of optimisation pressure this intervention
should have moved it. It did not, which is consistent with the interpretation we give in
\S\ref{sec:results_p24}: with $17$ to $18$ moving-layer examples of brow angle in the whole corpus, the
constraint is the number of distinct examples, not how often the optimiser sees them. Warm-starting
helps for a different reason, namely that it transfers a displacement field the rare parameter can
re-index rather than manufacturing supervision.

\paragraph*{A methodological note on how these were measured.}
Four of these evaluations were initially wrong in a way worth documenting, because the same trap
is easy to fall into. The parallax variants store extra heads that the baseline architecture does
not have; our evaluation script built the architecture from an environment flag, we forgot to set
it, and \texttt{load\_\allowbreak state\_\allowbreak dict} silently discarded those heads. The scores that
came back were of a crippled model, and they looked plausible ($0.6307$ to $0.7096$), which is
exactly what makes the failure dangerous. We now detect optional architecture branches from the
checkpoint's own key set and \emph{assert} that no checkpoint weight goes unused, so a mismatch
raises instead of scoring. Every number in this table is from the corrected path.

\section{Rendering: Where the Aliasing Came From and What Fixed It}
\label{app:antialias}

Two reviewers reported that aliasing in our renders impeded their judgement of animation
quality, and asked whether the reported metrics reflect the method or the renderer. We
instrumented the viewer to answer both parts.

\paragraph*{Multisampling was already enabled, and was the wrong tool.}
Reading the live WebGL2 state through the browser shows the framebuffer was created with
$\texttt{SAMPLES}\!=\!4$, $\texttt{SAMPLE\_BUFFERS}\!=\!1$ and
$\texttt{MAX\_SAMPLES}\!=\!4$, so $4\times$ MSAA was active from the start and could not be
increased. MSAA anti-aliases \emph{triangle} edges, but in this renderer the visible silhouette
of hair or a skirt hem is not a triangle edge: each layer is a textured triangle fan whose
outline comes from the layer's alpha channel sampled \emph{inside} the triangles, where all
$4$ coverage samples of a pixel read the same texel and MSAA does nothing. Reporting ``MSAA is
on'' would therefore have been a non-answer.

\paragraph*{The two real causes.}
First, minification without a mip chain: \texttt{TEXTURE\_\allowbreak MIN\_\allowbreak FILTER} was
$\texttt{LINEAR}$ while a layer texture of up to $600$\,px is typically drawn into about
$300$\,px of canvas, so bilinear filtering selected roughly one texel in four and the alpha
edge crawled. Second, a backing store equal to the CSS box at
$\texttt{devicePixelRatio}\!=\!1$, i.e.\ exactly one sample per output pixel, so nothing
anti-aliased the alpha edge at all. We fixed both: the backing store is now supersampled
$2\times$ per axis (clamped so large stages cannot exceed a $4096$\,px buffer), every layer
texture gets a mip chain with trilinear minification, and anisotropic filtering at $8\times$ is
enabled where the extension exists so strongly sheared meshes do not over-blur. Mip generation
happens in premultiplied-alpha space, which is the correct space and avoids dark or bright
halos at the silhouette.

\begin{figure*}[t]
  \centering
  \includegraphics[width=\linewidth]{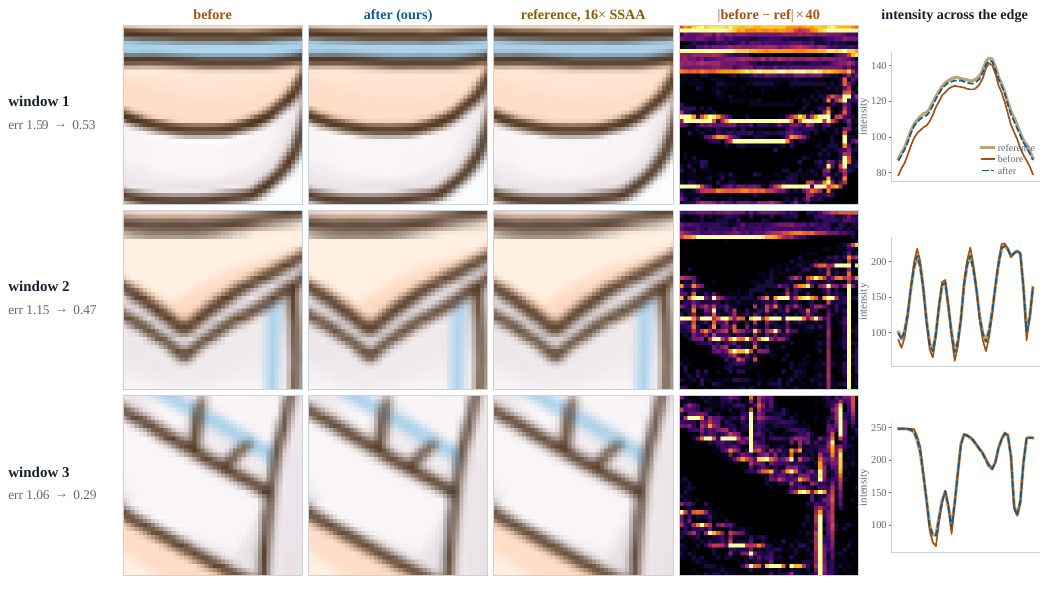}
  \caption{\textbf{Silhouette anti-aliasing, before and after the renderer fix.} Three $44\!\times\!44$\,px windows
 chosen automatically at the largest-error locations. Columns~1--3 are the rendered crops (before, after,
 and a $16\times$-supersampled reference) at nearest-neighbour magnification. Column~4 is
 $|\text{before}-\text{ref}|$ amplified $40\times$ on a perceptual colour map; column~5 is the intensity
 profile along the worst scanline of each window.}
  \Description{Three magnified crops of a character silhouette rendered before the fix, after the fix, and as a supersampled reference, with amplified error maps against the reference.}
  \label{fig:antialias}
\end{figure*}

\paragraph*{Measured, against an alias-free reference.}
We built the reference by rendering the identical view into a $3600^2$ backing store
($16\times$ area supersampling on top of MSAA) and box-filtering to $900^2$. Constructing that
reference from the old and the new code path gives images agreeing to $92.9$\,dB PSNR
(RMSE $0.006/255$), so the reference is unbiased with respect to the change and can score both
sides fairly. Scoring on the silhouette band (reference gradient above $8/255$, dilated
$2$\,px, $13.3\%$ of the frame) gives Tab.~\ref{tab:antialias}: edge-band error against the
alias-free reference drops $2.34\times$ ($+7.38$\,dB) and full-frame error $2.09\times$
($+6.39$\,dB). Excess staircase energy relative to the converged image, measured as
$\text{mean}|\nabla^2|$ on the band minus the reference's own value, falls from $0.759$ to
$0.152$, i.e.\ $5.0\times$ closer to convergence. Fig.~\ref{fig:antialias} shows two $76^2$\,px
windows on diagonal silhouette edges magnified with nearest-neighbour so no resampling can hide
the difference.

The consequence for our numbers is worth stating explicitly: the primary Stage-2 metric is
per-vertex direction cosine computed on \emph{geometry}, so it never touched the rasteriser and
is unaffected by any of this. The pixel metrics of Tab.~\ref{tab:stage2_video} and
Appendix~\ref{app:crossdecomp} were measured on the improved path, and the reviewers'
qualitative concern is addressed by the renderer fix rather than by an argument.

\begin{table}[t]
  \centering
  \caption{\textbf{Renderer anti-aliasing, before and after}, against a $16\times$-supersampled alias-free reference of the identical view (\texttt{bench\_34}, $35$ layers, rest pose, so all three images are pixel-comparable by construction). ``edge band'' is the $13.3\%$ of pixels where the reference has a silhouette gradient. Staircase is $\text{mean}|\nabla^2|$ on that band; the reference's own value is $13.172$, so the meaningful quantity is the excess over it.}
  \label{tab:antialias}
  \scriptsize
  \setlength{\tabcolsep}{3pt}
  \renewcommand{\arraystretch}{1.1}
  \begin{tabular*}{\linewidth}{@{\extracolsep{\fill}}lcccc@{}}
    \toprule
    & \multicolumn{2}{c}{full frame} & \multicolumn{2}{c}{edge band} \\
    \cmidrule(lr){2-3}\cmidrule(lr){4-5}
    Render path & PSNR & RMSE & PSNR & RMSE \\
    \midrule
    before ($\texttt{LINEAR}$, $900^2$ buffer)      & $51.99$ & $0.641$ & $43.49$ & $1.707$ \\
    after (mipmap $+$ aniso, $1800^2$ buffer)       & $\mathbf{58.38}$ & $\mathbf{0.307}$ & $\mathbf{50.87}$ & $\mathbf{0.729}$ \\
    \bottomrule
  \end{tabular*}
\end{table}

Columns~1--3 of Fig.~\ref{fig:antialias} will look nearly identical to a reader, and we say so rather
than implying otherwise: the residual is well under one intensity level per pixel, so no crop of the render
can make it dramatic. The evidence is in columns~4--5, where the staircase structure the fix removes is
visible without amplification in the intensity profile. Mean absolute error over the three windows falls
$1.27 \to 0.43$ (per window $1.59\!\to\!0.53$, $1.15\!\to\!0.47$, $1.06\!\to\!0.29$), consistent with the
$43.49\!\to\!50.87$\,dB silhouette-band PSNR of Tab.~\ref{tab:antialias}. What remains is at the level of
quantisation noise against a converged render.

\section{Parameter-Vocabulary Extension: Full Per-Parameter Results}
\label{app:p24}

This appendix backs \S\ref{sec:results_p24}. Table~\ref{tab:p24_perparam} reports every one of the
$24$ parameters individually on the $46$ held-out characters under true generation, next to the number
of moving-layer records that parameter has in training. Reading the table as a scatter of cosine
against support makes the conclusion of \S\ref{sec:results_p24} hard to avoid: the four parameters with
fewer than $30$ training samples occupy four of the five worst positions, the two with negative cosine
are the two rarest in the corpus, and the two best-learned new parameters (gaze) are the two with the
most support among the additions. Note also that the middle column, the $8$-parameter model evaluated
on the same pack and characters, shows the extension is close to free on the parameters both models
share, with \texttt{BodyAngleX} the only entry where the larger vocabulary is clearly better
($0.576\!\to\!0.599$) and \texttt{EyeROpen} the only one clearly worse ($0.863\!\to\!0.784$).

A second caveat applies to this table specifically and we flag it because it also bounds what the
benchmark can measure. A held-out character can only be scored on the parameters \emph{its artist
actually rigged}, so the effective evaluation pool shrinks for the rarer parameters: $40$ of $46$
characters carry ground truth for \texttt{AngleX}, $25$ for gaze, $15$ for brow height, and only $8$ for
\texttt{BrowLAngle}. The rarest parameters are therefore doubly disadvantaged, thin in training and
thin in evaluation, and their numbers carry correspondingly wide error bars. We state the per-parameter
character count in the table so this is visible rather than buried.

\begin{table}[t]
  \centering
  \caption{\textbf{All $24$ parameters, individually.} $46$ held-out characters, true generation, per-vertex direction cosine and magnitude ratio (ideal $1.0$). ``support'' is the number of moving-layer records in training; ``chars'' is how many of the $46$ held-out characters have artist ground truth for that parameter at all. $\ast$ marks a parameter absent from the $8$-parameter vocabulary, so its ``$8$p'' cell is empty by construction. Colour marks the capability tier of Tab.~\ref{tab:p24_tiers}: \good{good}, \marg{marginal}, \bad{unusable}.}
  \label{tab:p24_perparam}
  \scriptsize
  \setlength{\tabcolsep}{3pt}
  \renewcommand{\arraystretch}{1.05}
  \begin{tabular*}{\linewidth}{@{\extracolsep{\fill}}lrrcccc@{}}
    \toprule
    Parameter & support & chars & $8$p & $24$p cold & $24$p warm & mag \\
    \midrule
    \texttt{EyeBallY}$\ast$ & $181$ & $23$ & -- & $0.999$ & \good{$0.999$} & $1.10$ \\
    \texttt{EyeBallX}$\ast$ & $215$ & $24$ & -- & $0.997$ & \good{$0.997$} & $0.96$ \\
    \texttt{BodyAngleZ} & $1097$ & $13$ & $0.876$ & $0.875$ & \good{$0.874$} & $1.33$ \\
    \texttt{AngleZ} & $2804$ & $36$ & $0.849$ & $0.818$ & \good{$0.845$} & $0.87$ \\
    \texttt{BrowRY}$\ast$ & $25$ & $12$ & -- & $0.840$ & \good{$0.843$} & $0.81$ \\
    \texttt{BrowLY}$\ast$ & $25$ & $12$ & -- & $0.845$ & \good{$0.832$} & $0.64$ \\
    \texttt{EyeLOpen} & $143$ & $29$ & $0.797$ & $0.761$ & \good{$0.789$} & $0.83$ \\
    \texttt{EyeROpen} & $147$ & $29$ & $0.863$ & $0.762$ & \good{$0.784$} & $0.82$ \\
    \texttt{Breath}$\ast$ & $948$ & $27$ & -- & $0.739$ & \good{$0.749$} & $0.68$ \\
    \texttt{AngleY} & $2736$ & $39$ & $0.744$ & $0.742$ & \good{$0.742$} & $1.15$ \\
    \texttt{EyeRSmile}$\ast$ & $17$ & $4$ & -- & $0.372$ & \good{$0.724$} & $0.57$ \\
    \texttt{AngleX} & $2841$ & $40$ & $0.721$ & $0.709$ & \good{$0.720$} & $1.15$ \\
    \texttt{EyeLSmile}$\ast$ & $16$ & $4$ & -- & $0.561$ & \good{$0.710$} & $0.44$ \\
    \texttt{HairBack}$\ast$ & $91$ & $23$ & -- & $0.545$ & \marg{$0.644$} & $0.48$ \\
    \texttt{BodyAngleY} & $623$ & $10$ & -- & $0.643$ & \marg{$0.635$} & $1.51$ \\
    \texttt{BodyAngleX} & $683$ & $14$ & $0.576$ & $0.667$ & \marg{$0.599$} & $0.53$ \\
    \texttt{HairFront}$\ast$ & $233$ & $31$ & -- & $0.536$ & \marg{$0.574$} & $0.52$ \\
    \texttt{HairSide}$\ast$ & $244$ & $27$ & -- & $0.381$ & \marg{$0.427$} & $0.52$ \\
    \texttt{BrowLForm}$\ast$ & $29$ & $9$ & -- & $0.236$ & \bad{$0.389$} & $0.67$ \\
    \texttt{BrowRForm}$\ast$ & $48$ & $10$ & -- & $0.145$ & \bad{$0.337$} & $0.33$ \\
    \texttt{MouthOpenY} & $126$ & $31$ & $0.239$ & $0.274$ & \bad{$0.235$} & $0.38$ \\
    \texttt{MouthForm}$\ast$ & $123$ & $24$ & -- & $0.114$ & \bad{$0.176$} & $0.03$ \\
    \texttt{BrowRAngle}$\ast$ & $18$ & $9$ & -- & $-0.020$ & \bad{$-0.022$} & $0.04$ \\
    \texttt{BrowLAngle}$\ast$ & $17$ & $8$ & -- & $-0.266$ & \bad{$-0.073$} & $0.03$ \\
    \bottomrule
  \end{tabular*}
\end{table}

\paragraph*{Making density-confounded comparisons impossible by construction.}
The measurement error described in \S\ref{sec:results_p24}, where a re-meshing worth $0.063$ cosine was
briefly mistaken for a model regression worth $0.031$, was possible because our evaluation script
accepted any (checkpoint, pack) pair and averaged whatever it found. We added three switches that make
a controlled comparison expressible and an uncontrolled one obvious: restricting which parameters are
scored, mapping parameter names through a checkpoint's own vocabulary so a small-vocabulary model can be
scored on a large-vocabulary pack without shifting embedding indices, and restricting which layers are
scored while all layers still enter the forward pass as joint context. The last switch is what allows
the density control (identical-vertex-count layers only) to be measured without changing the model
input, since removing a layer from the forward pass would change every other layer's prediction. We
recommend that any future work reporting per-vertex cosine on Live2D-Bench state the pack, the character
list and the parameter list alongside the number, and we ship ours with the released models.

\section{Cross-Decomposer Generalization of the Animation Prior}
\label{app:crossdecomp}

A practical requirement for deployment is that the Stage~2 animation prior be
\emph{decomposer-agnostic}: it should produce coherent motion on any reasonable layer
stack, not only on layers emitted by our own Stage~1. This matters because a user may
already have a preferred decomposition tool, and because it isolates whether Stage~2 has
learned a transferable animation prior versus merely co-adapting to Stage~1 artifacts.

\begin{figure*}[t]
  \centering
  \includegraphics[width=\linewidth]{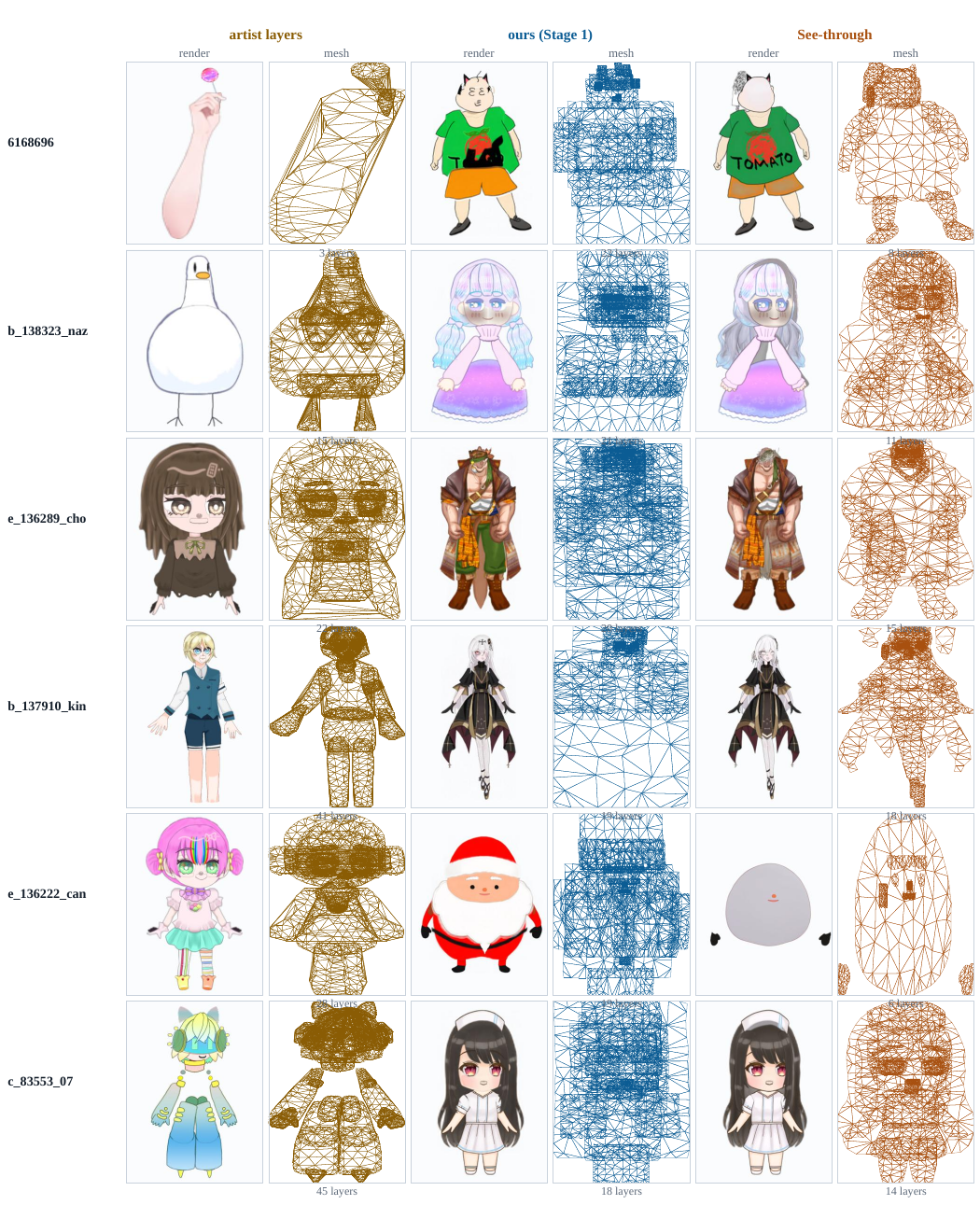}
  \caption{\textbf{Cross-decomposer generalization.} Six held-out illustrations, each given to \emph{one frozen}
 Stage~2 checkpoint as three layer stacks: the artist's own (gold), our Stage~1 (blue), and
 See-through~\cite{lin2026seethrough} (orange). Each condition appears twice, as the rendered frame and as
 the posed mesh at the same keypose, with that stack's own layer count beneath.}
  \label{fig:crossdecomp}
\end{figure*}

We test this by feeding the \emph{same} in-the-wild illustration through two independent
decomposers, our Stage~1 and the third-party See-through~\cite{lin2026seethrough}
decomposer, which produces a different layer count and split convention, and animating
both stacks with the identical, frozen Stage~2 model (auto-mesh + image-conditioned
offset prediction). As shown in Fig.~\ref{fig:crossdecomp}, the model yields coherent,
structurally consistent animation (head roll shown) on both decompositions across
characters spanning distinct art styles and body types, with no per-case tuning. The
animation prior thus transfers across decomposition sources, confirming that Stage~2
conditions on layer \emph{appearance and geometry} rather than on decomposer-specific
cues.

Motion stays coherent under all three stacks in Fig.~\ref{fig:crossdecomp} with no per-case
adjustment, so the learned animation prior is decomposer-agnostic. The artist-layer column bounds what
Stage~2 can achieve when the decomposition is perfect, and where a render shows an artefact the mesh column
beside it tells you whether Stage~2 faithfully animated a broken decomposition or predicted the wrong
motion.

\paragraph*{Quantifying cross-decomposer render fidelity.}
Beyond the qualitative evidence of Fig.~\ref{fig:crossdecomp}, we quantify how the choice of
layer-decomposition source affects end-to-end animation fidelity. On the three OOD characters
that ship an artist-authored Live2D rig: so a ground-truth \emph{animation} exists as
reference; we decompose each character three ways: (i)~its artist ground-truth layers, (ii)~our
Stage~1, and (iii)~See-through~\cite{lin2026seethrough}. Every stack is animated by the identical
frozen Stage~2 model and rendered; we report PSNR/SSIM/LPIPS of the rendered composite against
the artist's own rendered animation, averaged over all $13$ non-rest keyposes per character
(Tab.~\ref{tab:crossdecomp_render}). Using the artist's own layers is an upper bound (the
reference shares the same textures). Among the two \emph{automatic} decomposers, our Stage~1
outperforms See-through on every metric (PSNR $23.4$ vs.\ $21.1$, SSIM $0.883$ vs.\ $0.865$,
LPIPS $0.088$ vs.\ $0.120$), and the advantage holds per character on the two with appreciable
motion. This confirms the frozen Stage~2 prior transfers across decomposers while rewarding the
cleaner layer stack.

\paragraph*{The gap is decomposition, not motion.}
A rest-pose control isolates the two error sources. At rest, no animation, pure
decomposition-vs-artist mismatch, our Stage~1 already reaches PSNR $27.2$ and See-through $22.6$,
versus the identical artist layers' ceiling. Adding the predicted animation drops these by only
$\sim\!3.7$ and $\sim\!1.5$\,dB respectively, so the large gap to the GT-layer condition is
dominated by how each decomposer re-cuts boundaries and in-paints occluded regions, \emph{not} by
motion error. Motion quality proper is measured directly on the mesh (per-vertex direction cosine
and magnitude ratio, Tab.~\ref{tab:stage2_mesh}).

\paragraph*{Caveat: pixel fidelity on GT layers is floor-dominated.}
When the layer source is the artist's own layers, pixel PSNR is a \emph{weak} proxy for motion
quality: a keypose perturbs only a small fraction of pixels and both renders share identical
textures, so PSNR saturates near a high floor regardless of motion accuracy. A predict-zero-motion
control (render the rest pose against the GT-motion render) scores PSNR $41.6$ on the same
keyposes, essentially indistinguishable from the animated model's $41.8$. Pixel metrics on GT
layers should therefore be read as a \emph{decomposition-source} comparison, where the
between-condition gaps are large and meaningful; LPIPS separates the conditions more cleanly than
PSNR ($0.028$ for GT layers vs.\ $0.088$/$0.120$ for the automatic decomposers).

\begin{table}[t]
  \centering
  \small
  \setlength{\tabcolsep}{3.4pt}
  \caption{\textbf{Cross-decomposer animation render fidelity.} Rendered-composite similarity to the artist's own
 rendered Live2D animation, mean over $3$ OOD characters $\times\,13$ non-rest keyposes, with the frozen
 Stage~2 model animating each decomposition source. All higher-is-better except LPIPS. See the text for how
 to read the control columns.}
  \label{tab:crossdecomp_render}
  \resizebox{\columnwidth}{!}{%
  \begin{tabular}{lcccc}
    \toprule
    & Rest & \multicolumn{3}{c}{Animated vs.\ artist} \\
    \cmidrule(lr){2-2}\cmidrule(lr){3-5}
    Layer source & PSNR & PSNR & SSIM & LPIPS \\
    \midrule
    Artist GT layers            & $\infty$ & $41.8$ & $\mathbf{0.955}$ & $\mathbf{0.028}$ \\
    \textbf{Ours (Stage~1)}     & $\mathbf{27.2}$ & $\mathbf{23.4}$ & $\mathbf{0.883}$ & $\mathbf{0.088}$ \\
    See-through~\cite{lin2026seethrough} & $22.6$ & $21.1$ & $0.865$ & $0.120$ \\
    \bottomrule
  \end{tabular}}
\end{table}

\emph{Rest} PSNR in Tab.~\ref{tab:crossdecomp_render} is a no-motion control (candidate rest pose
against artist rest pose) that isolates decomposition-versus-artist mismatch, and the \emph{Animated}
columns add the predicted motion. The GT-layer row is an upper bound: it shares the reference textures, so
its rest PSNR is unbounded, and its \emph{Animated} PSNR doubles as a predict-nothing floor ($41.6$),
which illustrates that pixel PSNR on GT layers barely responds to motion at all.

\paragraph*{Ablating the animation prior.}
Tab.~\ref{tab:deploy_prior_ablation} ablates the Stage~2 model of the main text on the strict
$46$-character clean benchmark (zero character overlap with training) under true generation (no
teacher forcing). The decisive factor is \emph{joint cross-layer coordination}: predicting all
layers of a pose together with self-attention spanning every vertex, so the model can represent
inter-layer relative motion, lifts direction cosine from $0.693$ (per-layer independent
regression) to $\mathbf{0.736}$ and simultaneously corrects the amplitude (magnitude ratio
$0.86\!\to\!0.99$), far above a per-vertex coordinate-MLP regressor ($0.582$). Neither backbone
scale nor an alternative mesh pack helps: a $0.8$B Qwen-VL regressor converges to a muted,
order-of-magnitude-slower solution, and one re-training run on content-conforming meshes with a
$2877$-entry parameter vocabulary diverged from a data-format mismatch (the successful
content-mesh training reported in Tab.~\ref{tab:stage2_mesh_ab} of the main text used the
corrected pack). The capacity conclusion is established more thoroughly by the $5.1$\,M to
$1.0$\,B sweep of Tab.~\ref{tab:stage2_ablations}.
Two further findings. (i)~Single-model direction accuracy carries seed variance of the order of the
$0.024$ floor of \S\ref{sec:results_stage2}, which is why the joint entry here reads $0.736$ while the
committed checkpoint reads $0.7397$; a 3-seed \emph{output ensemble} with amplitude rescaling exploits
that variance to reach cos $0.743$ with magnitude ratio $1.00$, at $3\times$ inference cost. We quote no
tighter number than the $0.024$ floor: an earlier version of this appendix reported
$\text{std}\approx0.006$ here, which came from a per-record rather than a per-character metric and is
retracted in \S\ref{sec:results_stage2}. (ii)~The architectural motion priors we tried, a 2.5D
depth-parallax turn prior and explicit draw-order / composite-context conditioning, did \emph{not} beat
that floor: averaged over its three runs the parallax prior scores $0.7206$ against the base pair's
$0.7411$, i.e.\ $\Delta$cos $-0.021$, which is worse by about the width of the floor rather than better.
This indicates
the single-model regression prior is near its ceiling on direction and the remaining
end-to-end gap is dominated by the decomposition (Tab.~\ref{tab:crossdecomp_render}), not the motion model.

\begin{table}[t]
  \centering \small \setlength{\tabcolsep}{4pt}
  \caption{\textbf{Ablation of the Stage~2 animation prior} (46-char clean benchmark, true
  generation, no teacher forcing). Joint cross-layer coordination is the decisive factor; capacity
  and backbone choice are not. Comparable within this table only. $^{\ddagger}$The magnitude ratio of the
  per-vertex coordinate-MLP run was not recorded on this pool; we give the direction cosine, which was,
  and do not carry over the value measured for it on the earlier $120$-example pool, because that
  protocol supplied the mesh prefix and the two are not comparable.}
  \label{tab:deploy_prior_ablation}
  \resizebox{\columnwidth}{!}{%
  \begin{tabular}{lcc}
    \toprule
    Variant & dir-cos\,$\uparrow$ & mag-ratio\,$\to\!1$ \\
    \midrule
    \textbf{3-seed ensemble $+$ rescale}     & $\mathbf{0.743}$ & $\mathbf{1.00}$ \\
    Joint multi-layer (single, ours)         & $0.736$ & $0.99$ \\
    Per-layer independent regression         & $0.693$ & $0.86$ \\
    Per-vertex coordinate-MLP                & $0.582$ & n/a$^{\ddagger}$ \\
    $0.8$B Qwen-VL regressor                 & \multicolumn{2}{c}{muted motion, $\sim\!10\times$ slower} \\
    Content-mesh re-train (format mismatch)  & \multicolumn{2}{c}{diverged} \\
    \bottomrule
  \end{tabular}}
\end{table}

\section{Extended Ablations of the Superseded Autoregressive Stage~2}
\label{app:ablations_extended}

For completeness we record the hyperparameter sweeps and abandoned variants of the \emph{superseded} autoregressive token model (Appendix~\ref{app:mesh_tokenization}), all measured on a $50$-character OOD pool under that design's teacher-forced protocol; none of these numbers is comparable with the main-text true-generation results, and none supports a claim in the main paper. \textbf{Hyperparameter sweeps} showed $Q\!=\!128$ to be a sweet spot: lower-bin $Q{=}64$ retained $\cos\!=\!0.995$ but magnitude drifted to $21.5\times$; higher-bin $Q{=}256$ over-diluted the vocabulary, dropping $\cos$ to $0.957$ while improving the magnitude error to $\mathbf{0.181}$. A smaller $d{=}128$ ($4.4$\,M) variant reached $\cos\!=\!0.993$, within $0.5$\,pp of the $12$\,M default, i.e.\ capacity already saturated for that design too. Removing label smoothing ($\sigma_\text{LS}{=}0$) sharpened bin predictions but slightly destabilised training ($\cos\!=\!0.992$). \textbf{Abandoned variants:} adding an ARAP edge-length regulariser conflicted with the magnitude head and collapsed the cosine to $0.13$; a per-axis magnitude $(\alpha_x,\alpha_y)$ caused a $\log 0$ crash in the first epoch.

\section{Failure Gallery}
\label{app:editfail}

This gallery is generated from the model the paper ships, at each character's \emph{worst} pose. For
every held-out character we search all (parameter, keypose) pairs for the lowest per-vertex direction
cosine and show that pair; both the artist's frame and ours are read directly out of the released
benchmark rig, so no inference is run and the figure cannot drift from the data. An earlier version of
this gallery came from the superseded autoregressive design; it has been replaced.

The five rows are five \emph{different} failure modes, and we classify them by which measurable
property is wrong rather than describing them impressionistically. A character with one or two layers
gives cross-layer attention nothing to condition on, which is a degenerate support rather than a
prediction error. Amplitude collapse (magnitude ratio far below $1$) and direction error (cosine near or
below $0$) are separately measurable and appear in different rows. The worst cases reach a cosine of
$-0.97$, i.e.\ the predicted field points almost exactly opposite to the artist's, which is worth
showing plainly: our mean of $0.7676$ is an average over characters that includes cases this bad.

\begin{figure*}[p]
  \centering
  \includegraphics[width=\linewidth]{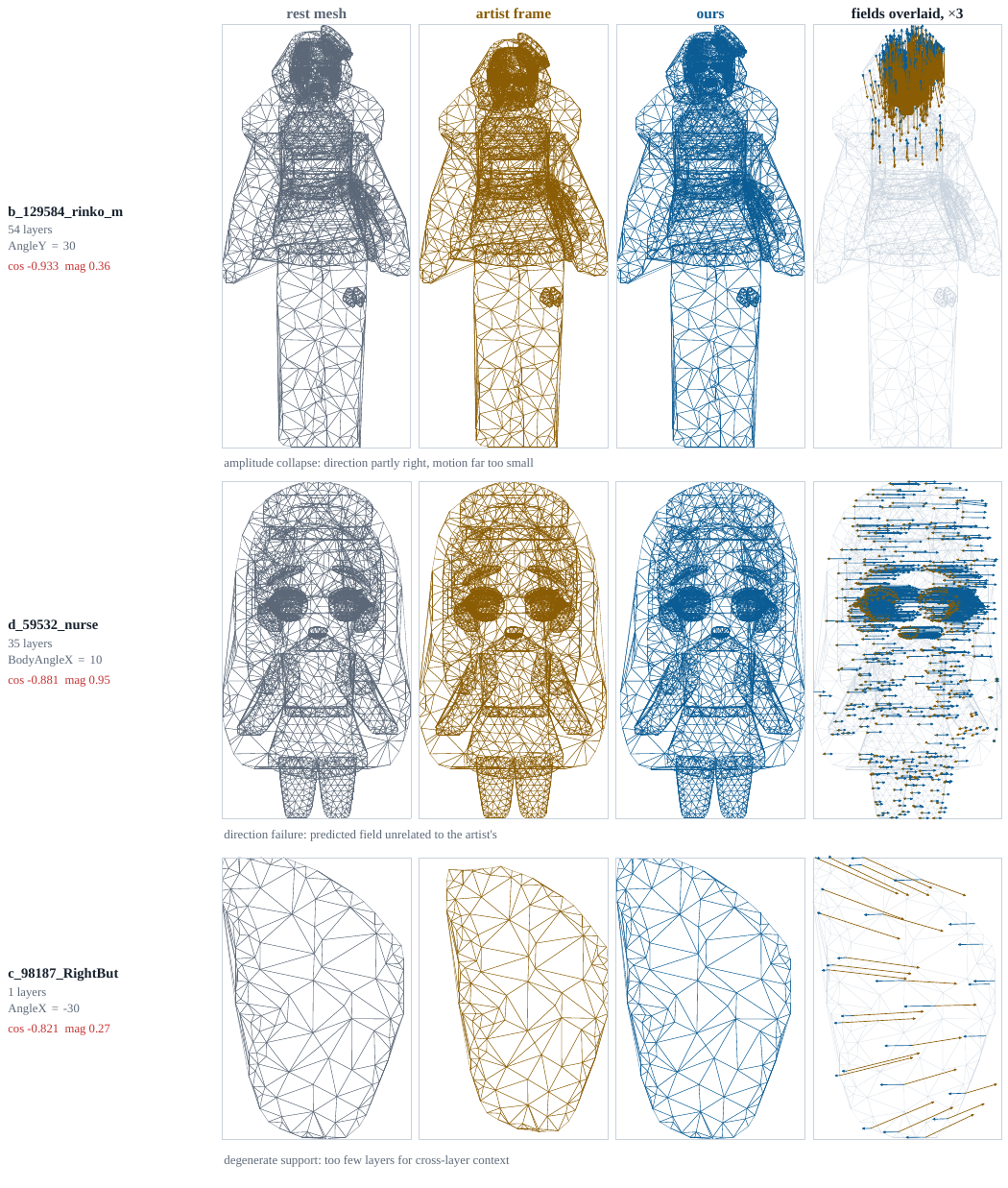}
  \caption{\textbf{Where the shipped model fails, at each character's worst pose.} Per row: the mesh at rest, the
 artist's frame, ours, and both displacement fields overlaid at $3\times$ (gold artist, blue ours). The
 parameter, keypose, cosine and magnitude ratio beside each row are that character's worst pose, computed
 from the released rig, and the line under each row is the computed failure classification. Rows are
 ordered by worst-pose cosine.}
  \Description{Five held-out characters at their worst pose, with the artist's and our displacement fields overlaid.}
  \label{fig:failure_gallery}
\end{figure*}

The rows of Fig.~\ref{fig:failure_gallery} are different failure modes, not one: a single-layer
character gives cross-layer attention nothing to condition on, whereas amplitude collapse and direction
error are distinct and separately measurable. The characters shown are those that fit the page at their
true aspect ratio.

\end{document}